\RequirePackage{fix-cm}
\documentclass[twocolumn]{svjour3}          
\smartqed  

\usepackage{times}

\usepackage{textcomp}
\usepackage{lipsum}
\usepackage{flushend}

\usepackage[utf8]{inputenc}

\usepackage{url}
\usepackage{ifpdf}

\ifpdf
\usepackage{xcolor}
\usepackage{graphics}
\usepackage[pdftex]{graphicx}
\DeclareGraphicsExtensions{.pdf,.PDF,.png,.PNG,.jpg,.JPG}
\usepackage{hyperref}
\hypersetup{
  colorlinks,%
  citecolor=blue,%
  filecolor=blue,%
  linkcolor=blue,%
  urlcolor=blue
}
\else
\usepackage{graphicx}
\DeclareGraphicsExtensions{.eps,.EPS,.ps,.PS,}
\fi

\usepackage{amsmath}
\usepackage{amssymb} 
\usepackage{stmaryrd}           
\usepackage{mathptmx}
\usepackage[mathscr]{euscript}  

\DeclareMathAlphabet{\mathcal}{OMS}{cmsy}{m}{n} 

\newcommand\imCMsym[4][\mathord]{%
  \DeclareFontFamily{U} {#2}{}
  \DeclareFontShape{U}{#2}{m}{n}{
    <-6> #25
    <6-7> #26
    <7-8> #27
    <8-9> #28
    <9-10> #29
    <10-12> #210
    <12-> #212}{}
  \DeclareSymbolFont{CM#2} {U} {#2}{m}{n}
  \DeclareMathSymbol{#4}{#1}{CM#2}{#3}
}

\imCMsym{cmmi}{124}{\CMjmath}

\usepackage{algorithm}
\usepackage[noend]{algpseudocode}

\usepackage{tabularx}
\usepackage{balance}
\usepackage{booktabs, multirow}
\usepackage[british]{babel}
\usepackage{dcolumn}
\usepackage{colortbl}

\usepackage[binary-units=true]{siunitx}

\usepackage{natbib}

\makeatletter
\if@twocolumn
  \renewcommand\normalsize{%
   \@setfontsize\normalsize\@xpt{12.5pt}%
   \abovedisplayskip=3 mm plus6pt minus 4pt
   \belowdisplayskip=3 mm plus6pt minus 4pt
   \abovedisplayshortskip=0.0 mm plus6pt
   \belowdisplayshortskip=2 mm plus4pt minus 4pt
   \let\@listi\@listI}%

  \renewcommand\small{%
   \@setfontsize\small{8.5pt}\@xpt
   \abovedisplayskip 8.5\p@ \@plus3\p@ \@minus4\p@
   \abovedisplayshortskip \z@ \@plus2\p@
   \belowdisplayshortskip 4\p@ \@plus2\p@ \@minus2\p@
   \def\@listi{\leftmargin\leftmargini
               \parsep 0\p@ \@plus1\p@ \@minus\p@
               \topsep 4\p@ \@plus2\p@ \@minus4\p@
               \itemsep0\p@}%
   \belowdisplayskip \abovedisplayskip}

\else
  \if@smallext
   \renewcommand\normalsize{%
   \@setfontsize\normalsize\@xpt\@xiipt
   \abovedisplayskip=3 mm plus6pt minus 4pt
   \belowdisplayskip=3 mm plus6pt minus 4pt
   \abovedisplayshortskip=0.0 mm plus6pt
   \belowdisplayshortskip=2 mm plus4pt minus 4pt
   \let\@listi\@listI}%

  \renewcommand\small{%
   \@setfontsize\small\@viiipt{9.5pt}%
   \abovedisplayskip 8.5\p@ \@plus3\p@ \@minus4\p@
   \abovedisplayshortskip \z@ \@plus2\p@
   \belowdisplayshortskip 4\p@ \@plus2\p@ \@minus2\p@
   \def\@listi{\leftmargin\leftmargini
               \parsep 0\p@ \@plus1\p@ \@minus\p@
               \topsep 4\p@ \@plus2\p@ \@minus4\p@
               \itemsep0\p@}%
   \belowdisplayskip \abovedisplayskip}
 \else
  \renewcommand\normalsize{%
   \@setfontsize\normalsize{9.5pt}{11.5pt}%
   \abovedisplayskip=3 mm plus6pt minus 4pt
   \belowdisplayskip=3 mm plus6pt minus 4pt
   \abovedisplayshortskip=0.0 mm plus6pt
   \belowdisplayshortskip=2 mm plus4pt minus 4pt
   \let\@listi\@listI}%

  \renewcommand\small{%
   \@setfontsize\small\@viiipt{9.25pt}%
   \abovedisplayskip 8.5\p@ \@plus3\p@ \@minus4\p@
   \abovedisplayshortskip \z@ \@plus2\p@
   \belowdisplayshortskip 4\p@ \@plus2\p@ \@minus2\p@
   \def\@listi{\leftmargin\leftmargini
               \parsep 0\p@ \@plus1\p@ \@minus\p@
               \topsep 4\p@ \@plus2\p@ \@minus4\p@
               \itemsep0\p@}%
   \belowdisplayskip \abovedisplayskip}
  \fi
\fi
\let\footnotesize\small
\makeatother

\usepackage{microtype}

\usepackage{scrtime}
\usepackage[dont-mess-around]{fnpct}

\usepackage{etoolbox}

\newcommand{\secref}[1]{Section~\ref{#1}}

\renewcommand{\eqref}[1]{Equation~(\ref{#1})}
\newcommand{\figref}[1]{Figure~\ref{#1}}
\newcommand{\tabref}[1]{Table~\ref{#1}}

\newcommand{\rot}[1]{\rotatebox[origin=c]{90}{#1}}

\DeclareSIUnit{\million}{\text{M}}
\DeclareSIUnit{\billion}{\text{B}}
\DeclareMathAlphabet\mathbfcal{OMS}{cmsy}{b}{n}

\newcolumntype{d}[1]{D{.}{.}{#1}}
\newcolumntype{Y}{D{.}{.}{1.2}}
\newcolumntype{P}[1]{>{\centering\arraybackslash}p{#1}}
\journalname{International Journal of Computer Vision}
\begin{document}

\title{EfficientPS: Efficient Panoptic Segmentation
}


\author{Rohit Mohan \and Abhinav Valada}


\institute{Rohit Mohan \at
          {mohan@cs.uni-freiburg.de}
          \and
          Abhinav Valada \at
          {valada@cs.uni-freiburg.de} \\
          {University of Freiburg, Germany}
}

\date{ }

\maketitle
\begin{abstract}
Understanding the scene in which an autonomous robot operates is critical for its competent functioning. Such scene comprehension necessitates recognizing instances of traffic participants along with general scene semantics which can be effectively addressed by the panoptic segmentation task. In this paper, we introduce the Efficient Panoptic Segmentation (EfficientPS) architecture that consists of a shared backbone which efficiently encodes and fuses semantically rich multi-scale features. We incorporate a new semantic head that aggregates fine and contextual features coherently and a new variant of Mask R-CNN as the instance head. We also propose a novel panoptic fusion module that congruously integrates the output logits from both the heads of our EfficientPS architecture to yield the final panoptic segmentation output. Additionally, we introduce the KITTI panoptic segmentation dataset that contains panoptic annotations for the popularly challenging KITTI benchmark. Extensive evaluations on Cityscapes, KITTI, Mapillary Vistas and Indian Driving Dataset demonstrate that our proposed architecture consistently sets the new state-of-the-art on all these four benchmarks while being the most efficient and fast panoptic segmentation architecture to date.
    
\end{abstract}

\section{Introduction}
\label{sec:intro}

\begin{figure}
\centering
\includegraphics[width=\linewidth]{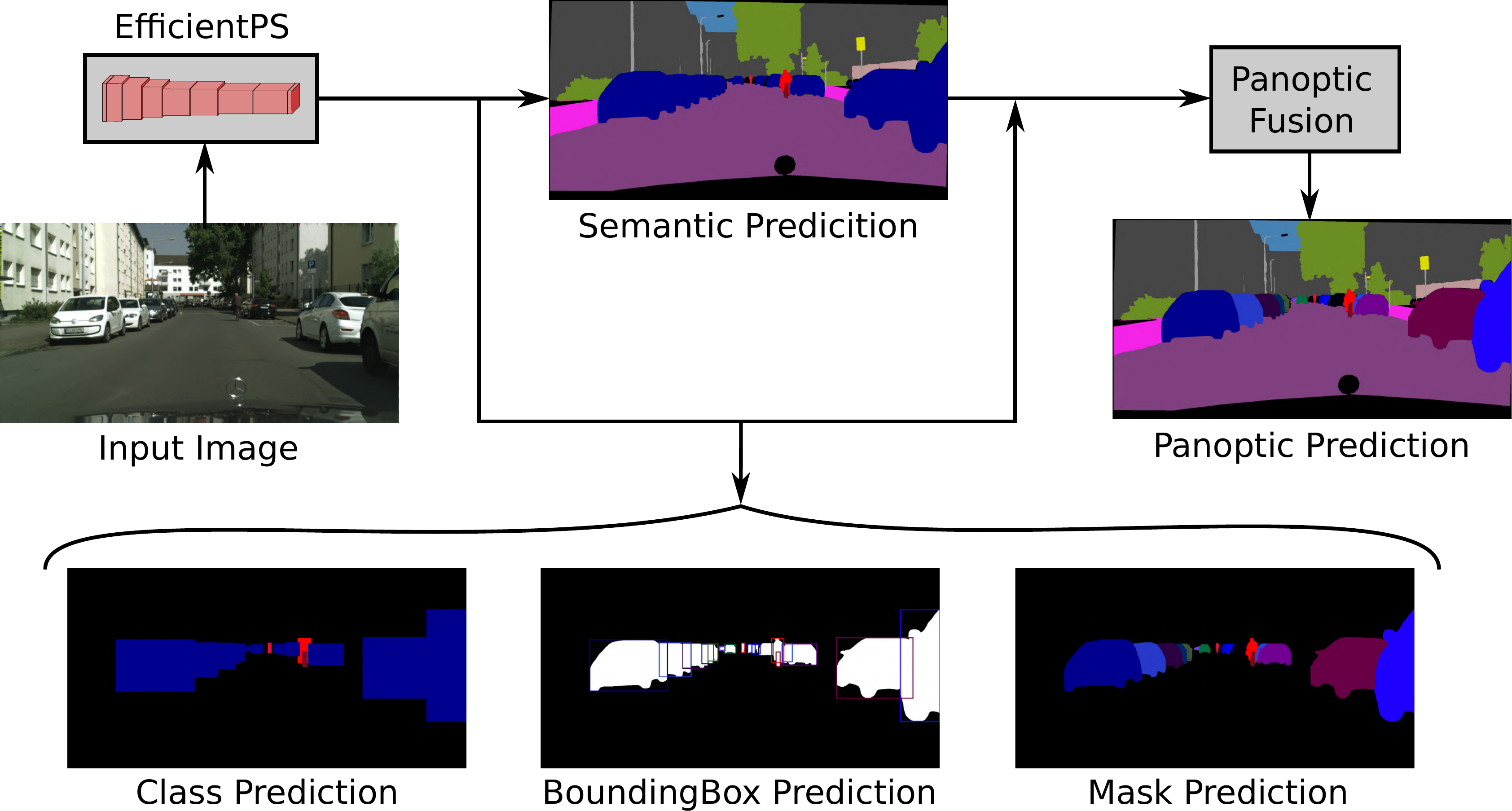}
\caption{Overview of our proposed EfficientPS architecture for panoptic segmentation. Our model predicts four outputs: semantics prediction from the semantic head, and class, bounding box and mask prediction from the instance head. All the aforementioned predictions are then fused in the panoptic fusion module to yield the final panoptic segmentation output.}
\label{fig:cover}
\end{figure}

Holistic scene understanding plays a pivotal role in enabling intelligent behavior. Humans from an early age are able to effortlessly comprehend complex visual scenes which forms the bases for learning more advanced capabilities~\citep{bremner2008theories}. Similarly, intelligent systems such as robots should have the ability to coherently understand visual scenes at both the fundamental pixel-level as well as at the distinctive object instance level. This enables them to perceive and reason about the environment holistically which facilitates interaction. Such modeling ability is a crucial enabler that can revolutionize several diverse applications including autonomous driving, surveillance, and augmented reality.

The components of a scene can generally be categorized into ‘stuff’ and ‘thing’ objects. ‘Stuff’ can be defined as uncountable and amorphous regions such as sky, road and sidewalk, while ‘thing’ are countable objects for example pedestrians, cars and riders. Segmentation of ‘stuff’ classes is primarily addressed using the semantic segmentation task, whereas segmentation of ‘thing’ classes is addressed by the instance segmentation task. Both tasks have garnered a substantial amount of attention in recent recent years~\citep{shotton2008semantic, krahenbuhl2011efficient, silberman2014instance, He_2014_CVPR}. Moreover, advances in deep learning~\citep{chen2018encoder, zhao2017pyramid, valada2016convoluted, he2017mask, liu2018path, zurn2019self} have further boosted the performance of these tasks to new heights. However, state-of-the-art deep learning methods still predominantly address theses tasks independently although their objective of understanding the scene at the pixel level establishes an inherent connection between them. More surprisingly, they have also fundamentally branched out into different directions of proposal based methods~\citep{he2017mask} for instance segmentation and fully convolutional networks~\citep{Long_2015_CVPR} for semantic segmentation, even though some earlier approaches~\citep{tighe2014scene, tu2005image, yao2012describing} have demonstrated the potential benefits in combining them.\nocite{radwan2018multimodal, valada2016towards, valada2018incorporating}

Recently, \cite{kirillov2019panoptic} revived the need to tackle these tasks jointly by coining the term panoptic segmentation and introducing the panoptic quality metric for combined evaluation. The goal of this task is to jointly predict ‘stuff’ and ‘thing’ classes, essentially unifying the separate tasks of semantic and instance segmentation. More specifically, if a pixel belongs to the ‘stuff’ class, the panoptic segmentation network assigns a class label from the ‘stuff’ classes, whereas if the pixel belongs to the ‘thing’ class, the network predicts both which ‘thing’ class it corresponds to as well as the instance of the object class. \cite{kirillov2019panoptic} also present a baseline approach for panoptic segmentation that heuristically combines predictions from individual state-of-the-art instance and semantic segmentation networks in a post-processing step. However, this disjoint approach has several drawbacks including large computational overhead, redundancy in learning and discrepancy between the predictions of each network. Although recent methods have made significant strides to address this task in top-down manner with shared components or in a bottom-up manner sequentially, these approaches still face several challenges in terms of computational efficiency, slow runtimes and subpar results compared to task-specific individual networks.

In this paper, we propose the novel EfficientPS architecture that provides effective solutions to the aforementioned problems for urban road scene understanding. The architecture consists of our new shared backbone with mobile inverted bottleneck units and our proposed 2-way Feature Pyramid Network (FPN), followed by task-specific instance and semantic segmentation heads with seperable convolutions, whose outputs are combined in our parameter-free panoptic fusion module. The entire network is jointly optimized in an end-to-end manner to yield the final panoptic segmentation output. \figref{fig:cover} shows an overview of the information flow in our network along with the intermediate predictions and the final output. The design of our proposed EfficientPS is influenced by the goal of achieving superior performance compared to existing methods while simultaneously being fast and computationally more efficient.

Currently, the best performing top-down panoptic segmentation models~\citep{porzi2019seamless,xiong2019upsnet,li2018learning} primarily employ the ResNet-101~\citep{he2016deep} or ResNeXt-101~\citep{xie2017aggregated} architecture with Feature Pyramid Networks~\citep{lin2017feature} as the backbone. Although these backbones have a high representational capacity, they consume a significant amount of parameters. In order to achieve a better trade-off, we propose a new backbone network consisting of a modified EfficientNet~\citep{tan2019efficientnet} architecture that employs compound scaling to uniformly scale all the dimensions of the network, coupled with our novel 2-way FPN. Our proposed backbone is substantially more efficient as well as effective than its popular counterparts~\citep{he2016deep, kaiser2017depthwise, xie2017aggregated}. Moreover, we identify that the standard FPN architecture has its limitations to aggregate multi-scale features due to the unidirectional flow of information. While there are other extensions that aim to mitigate this problem by adding bottom-up path augmentation~\citep{liu2018path} to the outputs of the FPN. We propose our novel 2-way FPN as an alternate that facilities bidirectional flow of information which substantially improves the panoptic quality of ‘thing’ classes while remaining comparable in runtime.

Now the outputs of our 2-way FPN are of multiple scales which we refer to as large-scale features when they have a downsampling factor of $\times4$ or $\times8$ with respect to the input image, and small-scale features when they have a downsampling factor of $\times16$ or $\times32$. The large-scale outputs comprise of fine or characteristic features, whereas the small-scale outputs contain features rich in semantic information. The presence of these distinct characteristics necessitates processing features at each scale uniquely. Therefore, we propose a new semantic head with depthwise separable convolutions, which aggregates small-scale and large-scale features independently before correlating and fusing contextual features with fine features. We demonstrate that this semantically reinforces fine features resulting in better object boundary refinement. For our instance head, we build upon Mask-R-CNN and augment it with depthwise separable convolutions and iABN sync~\citep{rota2018place} layers.

One of the critical challenges in panoptic segmentation deals with resolving the conflict of overlapping predictions from the semantic and instance heads. Most architectures~\citep{kirillov2019bpanoptic, porzi2019seamless, li2019attention, de2018panoptic} employ a standard post-processing step~\citep{kirillov2019panoptic} that adopts instance-specific ‘thing’ segmentation from the instance head and ‘stuff’ segmentation from the semantic head. This fusion technique completely ignores the logits of the semantic head while segmenting ‘thing’ regions in the panoptic segmentation output which is sub-optimal as the ‘thing’ logits of the semantic head can aid in resolving the conflict more effectively. In order to thoroughly exploit the logits from both heads, we propose a parameter-free panoptic fusion module that adaptively fuses logits by selectively attenuating or amplifying fused logit scores based on how agreeable or disagreeable the predictions of individual heads are for each pixel in a given instance. We demonstrate that our panoptic fusion mechanism is more effective and efficient than other widely used methods in existing architectures.

Furthermore, we also introduce the KITTI panoptic segmentation dataset that contains panoptic annotations for images in the challenging KITTI benchmark~\citep{Geiger2013IJRR}. As KITTI provides groundtruth for a whole suite of perception and localization tasks, these new panoptic annotations further complement the widely popularly benchmark. We hope that these panoptic annotations that we make publicly available encourages future research in multi-task learning for holistic scene understanding. Furthermore, in order to facilitate comparison, we benchmark previous state-of-the-art models on our newly introduced KITTI panoptic segmentation dataset and the IDD dataset. We perform exhaustive experimental evaluations and benchmarking of our proposed EfficientPS architecture on four standard urban scene understanding datasets including Cityscapes~\citep{cordts2016cityscapes}, Mapillary Vistas~\citep{neuhold2017mapillary}, KITTI~\citep{Geiger2013IJRR} and Indian Driving Dataset (IDD)~\citep{varma2019idd}.

Our proposed EfficientPS with a PQ score of $66.4\%$ is ranked first for panoptic segmentation on the Cityscapes benchmark leaderboard without training on \textit{coarse} annotations or using model ensembles. Additionally, EfficientPS is also ranked second for the semantic segmentation task as well as the instance segmentation task on the Cityscapes benchmark with a mIoU score of $84.2\%$ and an AP of $39.1\%$ respectively. On the Mapillary Vistas dataset, our single EfficientPS model achieves a PQ score of $40.5\%$ on the validation set, thereby outperforming all the existing methods. Similarly, EfficientPS consistently outperforms existing panoptic segmentation models on both the KITTI and IDD datasets by a large margin. More importantly, our EfficientPS architecture not only sets the new state-of-the-art on all the four panoptic segmentation benchmarks, but it is also the most computationally efficient by consuming the least amount of parameters and having the fastest inference time compared to previous state-of-the-art methods. Finally, we present detailed ablation studies that demonstrate the improvement in performance due to each of the architectural contributions that we make in this work. Moreover, we also make implementations of our proposed EfficientPS architecture, training code and pre-trained models publicly available.

In summary, the following are the main contributions of this work:
\begin{enumerate}
\item The novel EfficientPS architecture for panoptic segmentation that incorporates our proposed efficient shared backbone with our new feature aligning semantic head, a new variant of Mask R-CNN as the instance head, and our novel adaptive panoptic fusion module.
\item A new panoptic backbone consisting of an augmented EfficientNet architecture, and our proposed 2-way FPN that both encodes and aggregates semantically rich multi-scale features in a bidirectional manner.
\item A novel semantic head that captures fine features and long-range context efficiently as well as correlates them before fusion for better object boundary refinement.
\item A new panoptic fusion module that dynamically adapts the fusion of logits from the semantic and instance heads based on their mask confidences and congruously integrates instance-specific ‘thing’ classes with ‘stuff’ classes to compute the panoptic prediction.
\item The KITTI panoptic segmentation dataset that provides panoptic groundtruth annotations for images from the challenging KITTI benchmark dataset.
\item Benchmarking of existing state-of-the-art panoptic segmentation architectures on the newly introduced KITTI panoptic segmentation dataset and IDD dataset.
\item \sloppy Comprehensive benchmarking of our proposed EfficientPS architecture on Cityscapes, Mapilliary Vistas, KITTI and IDD datasets.
\item Extensive ablation studies that compare the performance of various architectural components that we propose in this work with their counterparts from state-of-the-art architectures. 
\item \sloppy Implementation of our proposed architecture and a live demo on all the four datasets is publicly available at \url{http://rl.uni-freiburg.de/research/panoptic}.
\end{enumerate}



\section{Related Works}
\label{sec:relatedworks}

Panoptic segmentation is a recently introduced scene understanding problem~\citep{kirillov2019panoptic} that unifies the tasks of semantic segmentation and instance segmentation. There are numerous methods that have been proposed for each of these sub-tasks, however only a handful of approaches have been introduced to tackle this coherent scene understanding problem of panoptic segmentation. Most works in this domain are largely built upon advances made in semantic segmentation and instance segmentation, therefore we first review recent methods that have been proposed for these closely related tasks, followed by state-of-the-art approaches that have been introduced for panoptic segmentation.

{\parskip=5pt
\noindent\textbf{Semantic Segmentation:} There has been significant advanc- es in semantic segmentation approaches in recent years. In this section, we briefly review methods that use a single monocular image to tackle this task. Approaches from the past decade, typically employ random decision forests to address this task. \cite{shotton2008semantic} use randomized decision forests on local patches for classification, whereas~\cite{plath2009multi} fuse local and global features along with Conditional Random Fields(CRFs) for segmentation. As opposed to leveraging appearance-based features, \cite{brostow2008segmentation} use cues from motion with random forests. \cite{sturgess2009combining} further combine appearance-based features with structure-from-motion features in addition to CRFs to improve the performance. However, 3D features extracted from dense depth maps~\citep{zhang2010semantic} have been demonstrated to be more effective than the combined features. \cite{kontschieder2011structured} exploit the inherent topological distribution of object classes to improve the performance, whereas ~\cite{krahenbuhl2011efficient} improve segmentation by pairing CRFs with Gaussian edge potentials. Nevertheless, all these methods employ handcrafted features that do not encapsulate all the high-level and low-level relations thereby limiting their representational ability.} 

The significant improvement in performance of classification tasks brought about by Convolutional Neural Network (CNN) based approaches motivated researchers to explore such methods for semantic segmentation. Initially, these approaches relied on patch-wise training that severely limited their ability to accurately segment object boundaries. However, they still perform substantially better than previous handcrafted methods. The advent of end-to-end learning approaches for semantic segmentation lead by the introduction of Fully Convolutional Networks (FCNs)~\citep{Long_2015_CVPR} revolutionized this field and FCNs still form the base upon which state-of-the-art architecture are built upon today. FCN is an encoded-decoder architecture where the encoder is based on the VGG-16~\citep{simonyan2014very} architecture with inner-product layers replaced with convolutions, and the decoder consists of convolution and transposed convolution layers. The subsequently proposed SegNet~\citep{badrinarayanan2017segnet} architecture introduced unpooling layers for upsampling as a replacement for transposed convolutions, whereas ParseNet~\citep{liu2015} models global context directly as opposed to only relying on the largest receptive field of the network. 
 
The PSPNet~\citep{zhao2017pyramid} architecture emphasizes on the importance of multi-scale features and propose pyramid pooling to learn feature representations at different scales. \cite{yu2015multi} introduce atrous convolutions to further exploit multi-scale features in semantic segmentation networks. Subsequently, \cite{valada2017adapnet} propose multi-scale residual units with parallel atrous convolutions with different dilation rates to efficiently learn multiscale features throughout the network without increasing the number of parameters. \cite{chen2017rethinking} propose the Atrous Spatial Pyramid Pooling (ASPP) module that concatenates feature maps from multiple parallel atrous convolutions with different dilation rates and a global pooling layer. ASPP substantially improves the performance of semantic segmentation networks by aggregating multi-scale features and capturing long-range context, however it significantly increases the computational complexity. Therefore, \cite{chen2018searching} propose Dense Prediction Cells (DPC) and \cite{valada19ijcv} propose Efficient Atrous Spatial Pyramid Pooling (eASPP) that yield better semantic segmentation performance than ASPP while being 10-times more efficient. \cite{li2019global} suggest that global feature aggregation often leads to large pattern features and also over-smooth regions of small patterns which results in sub-optimal performance. In order to alleviate this problem, the authors propose the use of a global aggregation module coupled with a local distribution module which results in features that are balanced in small and large pattern regions. There are also several works that have been proposed to improve the upsampling in decoders of encoder-decoder architectures. In \citep{chen2018encoder}, the authors introduce a novel decoder module for object boundary refinement. \cite{tian2019decoders} propose data-dependent upsampling which accounts for the redundancy in the label space as opposed to simple bilinear upsampling.
 
{\parskip=5pt
\noindent\textbf{Instance Segmentation:} Some of the initial approaches employ CRFs~\citep{he2014exemplar} and minimize integer quadratic relations~\citep{tighe2014scene}. Methods that exploit CNNs with Markov random fields \citep{zhang2016instance} and recurrent neural networks~\citep{romera2016recurrent, ren2017end} have also been explored. In this section, we primarily discuss CNN-based approaches for instance segmentation. These methods can be categorized into proposal free and proposal based methods.}

Methods in the proposal free category often obtain instance masks from a resulting transformation. \cite{bai2017deep} uses CNNs to produce an energy map of the image and then perform a cut at a single energy level to obtain the corresponding object instances. \cite{liu2017sgn} employ a sequence of CNNs to solve sub-grouping problems in order to compose object instances. Some approaches exploit FCNs which either use local coherence for estimating instances~\citep{dai2016instance} or encode the direction of each pixel to its corresponding instance centre~\citep{uhrig2016pixel}. The recent approach, SSAP~\citep{gao2019ssap} uses pixel-pair affinity pyramids for computing the probability that two pixels hierarchically belong to the same instance. However, they achieve a lower than proposal based methods which has led to a decline in their popularity.

In proposal based methods, \cite{hariharan2014simultaneous} propose a method that uses Multiscale Combinatorial Grouping~\citep{arbelaez2014multiscale} proposals as input to CNNs for feature extraction and then employ an SVM classifier for region classification. Subsequently, \cite{hariharan2015hypercolumns} propose hypercolumn pixel descriptors for simultaneous detection and segmentation. In recent works, DeepMask~\citep{pinheiro2015learning} uses a patch of an image as input to a CNN which yields a class-agnostic segmentation mask and the likelihood of the patch containing an object. FCIS~\citep{li2017fully} employs position-sensitive score maps obtained from classification of pixels based on their relative positions to perform segmentation and detection jointly. \cite{dai2016instance} propose an approach for instance segmentation that uses three networks for distinguishing instances, estimating masks and categorizing objects. Mask R-CNN~\citep{he2017mask} is one of the most popular and widely used approaches in the present time. It extends Faster R-CNN for instance segmentation by adding an object segmentation branch parallel to an branch that performs bounding box regression and classification. More recently, \cite{liu2018path} propose an approach to improve Mask R-CNN by adding bottom-up path augmentation that enhances object localization ability in earlier layers of the network. Subsequently, BshapeNet~\citep{kang2018bshapenet} extends Faster R-CNN by adding a bounding box mask branch that provides additional information of object positions and coordinates to improve the performance of object detection and instance segmentation.

{\parskip=5pt 
\noindent\textbf{Panoptic Segmentation:} In an earlier attempt of unifying semantic and instance segmentation task,~\citep{tu2005image} uses a Bayesian framework to output scene representation as a parsing graph. Further, some approaches employ auxiliary variables to reason at the segment level~\citep{yao2012describing} and combination of region-level features with per-exemplar sliding window detectors~\citep{tighe2013finding} to address the task. Methods such as minimization of an integer quadratic program~\citep{tighe2014scene} and maximization of a posteriori inference~\citep{sun2013relating} have also been explored. Nevertheless, the aforementioned methods due to their complexity and sub-par performance couldn't garner much attention to the task. But later \cite{kirillov2019panoptic} revived the unification of semantic segmentation and instance segmentation tasks by introducing panoptic segmentation. They propose a baseline model that combines the output of PSPNet \citep{zhao2017pyramid} and Mask R-CNN~\citep{he2017mask} with a simple post-processing step in which each model processes the inputs independently. The methods that address this task of panoptic segmentation can be broadly classified into two categories: top-down or proposal based methods and bottom-up or proposal free methods. Most of the current state-of-the-art methods adopt the top-down approach. \cite{de2018panoptic} propose joint training with a shared backbone that branches into Mask R-CNN for instance segmentation and augmented Pyramid Pooling module for semantic segmentation. Subsequently, \cite{li2019attention} introduce Attention-guided Unified Network that uses proposal attention module and mask attention module for better segmentation of ‘stuff’ classes. All the aforementioned methods use a similar fusion technique to~\cite{kirillov2019panoptic} for the fusion of ‘stuff’ and ‘thing’ predictions.}

In top-down panoptic segmentation architectures, predictions of both heads have an inherent overlap between them resulting in the mask overlapping problem. In order to mitigate this problem, \cite{li2018weakly} propose a weakly supervised model where ‘thing’ classes are weakly supervised by bounding boxes and ‘stuff’ classes are supervised with image-level tags. Whereas, \cite{liu2019end} address the problem by introducing the spatial ranking module and \cite{li2018learning} propose a method that learns a binary mask to constrain output distributions of ‘stuff’ and ‘thing’ explicitly. Subsequently, UPSNet~\citep{xiong2019upsnet} introduces a parameter-free panoptic head to address the problem of overlapping of instances and also predicts an extra unknown class. More recently, AdaptIS~\citep{sofiiuk2019adaptis} uses point proposals to produce instance masks and jointly trains with a standard semantic segmentation pipeline to perform panoptic segmentation. In contrast, \cite{porzi2019seamless} propose an architecture for panoptic segmentation that effectively integrates contextual information from a lightweight DeepLab-inspired module with multi-scale features from a FPN.

Compared to the popular proposal based methods, there are only a handful of proposal free methods that have been proposed. Deeper-Lab~\citep{yang2019deeperlab} was the first bottom-up approach that was introduced and it employs an encoder-decoder topology to pair object centres for class-agnostic instance segmentation with DeepLab semantic segmentation. \cite{cheng2019panoptic} further builds on Deeper-Lab by introducing a dual-ASPP and dual-decoder structure for each sub-task branch. SSAP~\citep{gao2019ssap} proposes to group pixels based on a pixel-pair affinity pyramid and incorporate an efficient graph method to generate instances while jointly learning semantic labeling.

In this work, we adopt a top-down approach due to its exceptional ability to handle large scale variation of instances which is a critical requirement for segmenting ‘thing’ classes. We present the novel EfficientPS architecture that incorporates our proposed efficient backbone with our 2-way FPN for learning rich multi-scale features in a bidirectional manner, coupled with a new semantic head that captures fine-features and long-range context effectively, and a variant of Mask R-CNN augmented with depthwise separable convolutions as the instance head. We propose a novel panoptic fusion module to dynamically adapt the fusion of logits from the semantic and instance heads to yield the panoptic segmentation output. Our architecture achieves state-of-the-art results on benchmark datasets while being the most efficient and fast panoptic segmentation architecture.

\section{EfficientPS Architecture}
\label{sec:technicalApproach}

\begin{figure*}
\centering
\includegraphics[width=\linewidth]{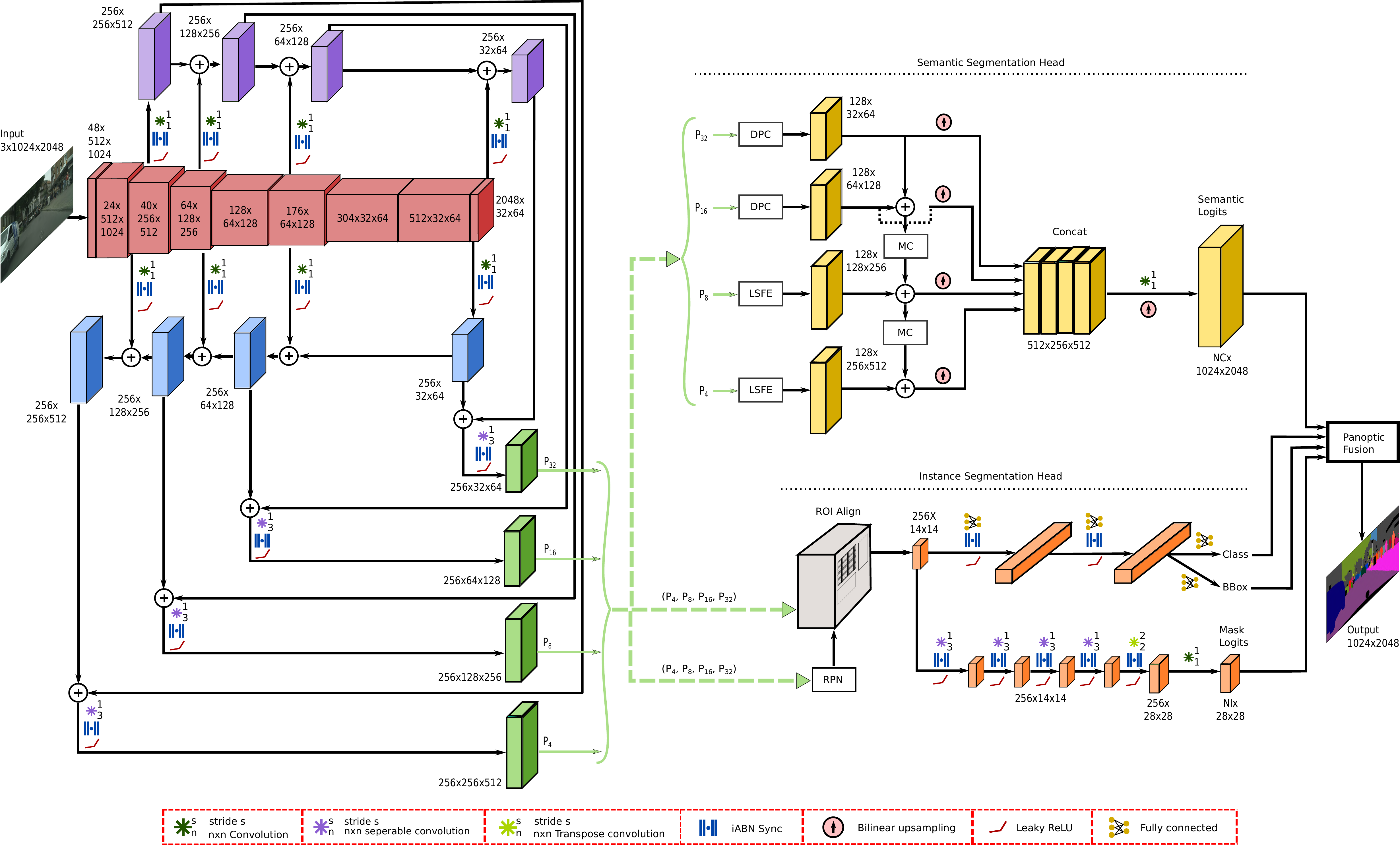}
\caption{Illustration of our proposed EfficientPS architecture consisting of a shared backbone with our 2-way FPN and parallel semantic and instance segmentation heads followed by our panoptic fusion module. The shared backbone is built upon on the EfficientNet architecture and our new 2-way FPN that enables bidirectional flow of information. The instance segmentation head is based on a modified Mask R-CNN topology and we incorporate our proposed semantic segmentation head. Finally, the outputs of both heads are fused in our panoptic fusion module to yield the panoptic segmentation output.}
\label{fig:epsnet_arch}
\end{figure*}

In this section, we first give a brief overview of our proposed EfficientPS architecture and then detail each of its constituting components. Our network follows the top-down layout as shown in \figref{fig:epsnet_arch}. It consists of a shared backbone with a 2-way Feature Pyramid Network (FPN), followed by task-specific semantic segmentation and instance segmentation heads. We build upon the EfficientNet~\citep{tan2019efficientnet} architecture for the encoder of our shared backbone (depicted in red). It consists of mobile inverted bottleneck~\citep{xie2017aggregated} units and employs compound scaling to uniformly scale all the dimensions of the encoder network. This enables our encoder to have a rich representational capacity with fewer parameters in comparison to other encoders or backbones of similar discriminative capability.

As opposed to employing the conventional FPN~\citep{lin2017feature} that is commonly used in other panoptic segmentation architectures~\citep{kirillov2019bpanoptic, li2018learning, porzi2019seamless}, we incorporate our proposed 2-way FPN that fuses multi-scale features more effectively than its counterparts. This can be attributed to the fact that the information flow in our 2-way FPN is not bounded to only one direction as depicted by the purple, blue and green blocks in \figref{fig:epsnet_arch}. Subsequently after the 2-way FPN, we employ two heads in parallel which are semantic segmentation (depicted in yellow) and instance segmentation (depicted in gray and orange) respectively. We use a variant of the Mask R-CNN~\citep{he2017mask} architecture as the instance head and we incorporate our novel semantic segmentation head consisting of dense prediction cells~\citep{chen2018searching} and residual pyramids. The semantic head consists of three different modules for capturing fine features, long-range contextual features and correlating the distinctly captured features for improving object boundary refinement. Finally, we employ our proposed panoptic fusion module to fuse the outputs of the semantic and instance heads to yield the panoptic segmentation output.

\subsection{Network Backbone}
\label{sec:backbone}

The backbone of our network consists of an encoder with our proposed 2-way FPN. The encoder is the basic building block of any segmentation network and a strong encoder is essential to have high representational capacity. In this work, we seek to find a good trade-off between the number of parameters and computational complexity to the representational capacity of the network. EfficientNets~\citep{tan2019efficientnet} which are a recent family of architectures have been shown to significantly outperform other networks in classification tasks while having fewer parameters and FLOPs. It employs compound scaling to uniformly scale the width, depth and resolution of the network efficiently. Therefore, we choose to build upon this scaled architecture with 1.6, 2.2 and 456 coefficients, commonly known as the EfficientNet-B5 model. This can be easily replaced with any of the EfficientNet models based on the capacity of the resources that are available and the computational budget.

In order to adapt EfficientNet to our task, we first remove the classification head as well as the Squeeze-and-Excitation (SE)~\citep{hu2018squeeze} connections in the network. We find that the explicit modelling of interdependencies between channels of the convolutional feature maps that are enabled by the SE connections tend to suppress localization of features in favour of contextual elements. This property is a desired in classification networks, however both are equally important for segmentation tasks, therefore we do not add any SE connections in our backbone. Second, we replace all the batch normalization~\citep{ioffe2015batch} layers with synchronized Inplace Activated Batch Normalization (iABN sync)~\citep{rota2018place}. This enables synchronization across different GPUs, which in turn yields a better estimate of gradients while performing multi-GPU training and the in-place operations frees up additional GPU memory. We analyze the performance of our modified EfficientNet in comparison to other encoders commonly used in state-of-the-art architectures in the ablation study presented in \secref{sec:encoderAblation}.

Our EfficientNet encoder comprises of nine blocks as shown in \figref{fig:epsnet_arch} (in red). We refer to each block in the figure as block 1 to block 9 in the left to right manner. The output of block 2, 3, 5, and 9 corresponds to downsampling factors $\times4, \times8, \times16$ and $\times32$ respectively. The outputs from these blocks with downsampling are also inputs to our 2-way FPN. The conventional FPN used in other panoptic segmentation networks aims to address the problem of multi-scale feature fusion by aggregating features of different resolutions in a top-down manner. This is performed by first employing a $1\times1$ convolution to reduce or increase the number of channels of different encoder output resolutions to a predefined number, typically 256. Then, the lower resolution features are upsampled to a higher resolution and are subsequently added together. For example, $\times32$ resolution encoder output features will be resized to the $\times16$ resolution and added to the $\times16$ resolution encoder output features. Finally, a $3\times3$ convolution is used at each scale to further learn fused features which yields the P\textsubscript{4}, P\textsubscript{8}, P\textsubscript{16} and P\textsubscript{32} outputs. This FPN topology has a limited unidirectional flow of information resulting in an ineffective fusion of multi-scale features. Therefore, we propose to mitigate this problem by adding a second branch that aggregates multi-scale features in a bottom-up manner to enable bidirectional flow of information. 

Our proposed 2-way FPN shown in \figref{fig:epsnet_arch} consists of two parallel branches. Each branch consists of a $1\times1$ convolution with 256 output filters at each scale for channel reduction. The top-down branch shown in blue follows the aggregation scheme of a conventional FPN from right to left. Whereas, the bottom-up branch shown in purple, downsamples the higher resolution features to the next lower resolution from left to right and subsequently adds them with the next lower resolution encoder output features. For example, $\times4$ resolution features will be resized to the $\times8$ resolution and added to the $\times8$ resolution encoder output features. Then in the next stage, the outputs from the bottom-up and top-down branches at each resolution are correspondingly summed together and passed through a $3\times3$ depthwise separable convolution with 256 output channels to obtain the P\textsubscript{4}, P\textsubscript{8}, P\textsubscript{16}, and P\textsubscript{32} outputs respectively. We employ depthwise separable convolutions as opposed to standard convolutions in an effort to keep the parameter consumption low. We evaluate the performance of our proposed 2-way FPN in comparison to the conventional FPN in the ablation study presented in \secref{sec:fpnAblation}.

\begin{figure}
\centering
\footnotesize
{\renewcommand{\arraystretch}{1.2}
\begin{tabular}{P{2.5cm} P{2.5cm} P{2.5cm}}
\includegraphics[width=0.85\linewidth]{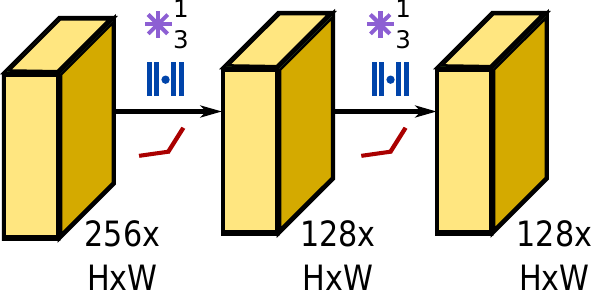} \hfill & 
\hfill \includegraphics[width=0.85\linewidth]{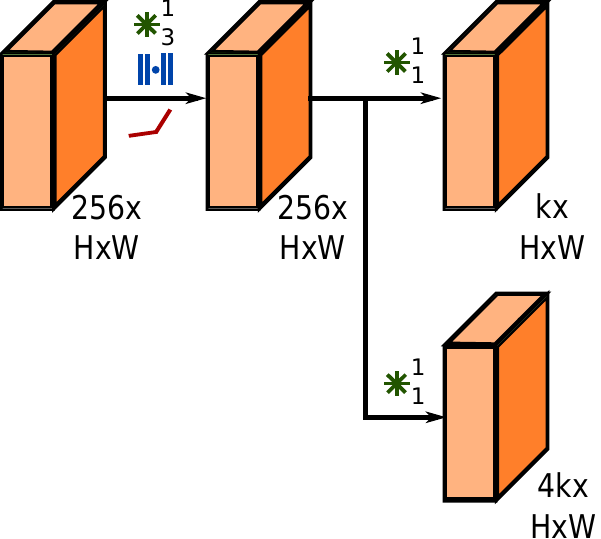} \hfill & \includegraphics[width=0.85\linewidth]{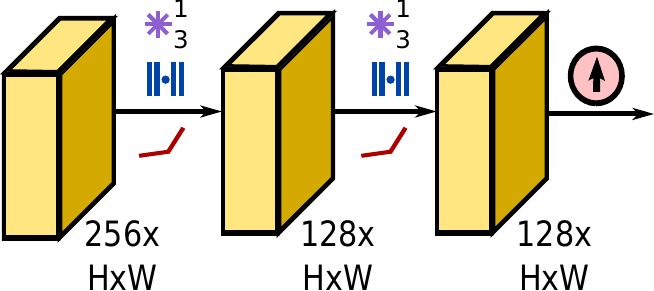} \\
(a) LSFE Module & (b) RPN Module & (c) MC Module \\ 
\\
\multicolumn{3}{c}{\includegraphics[width=0.975\linewidth]{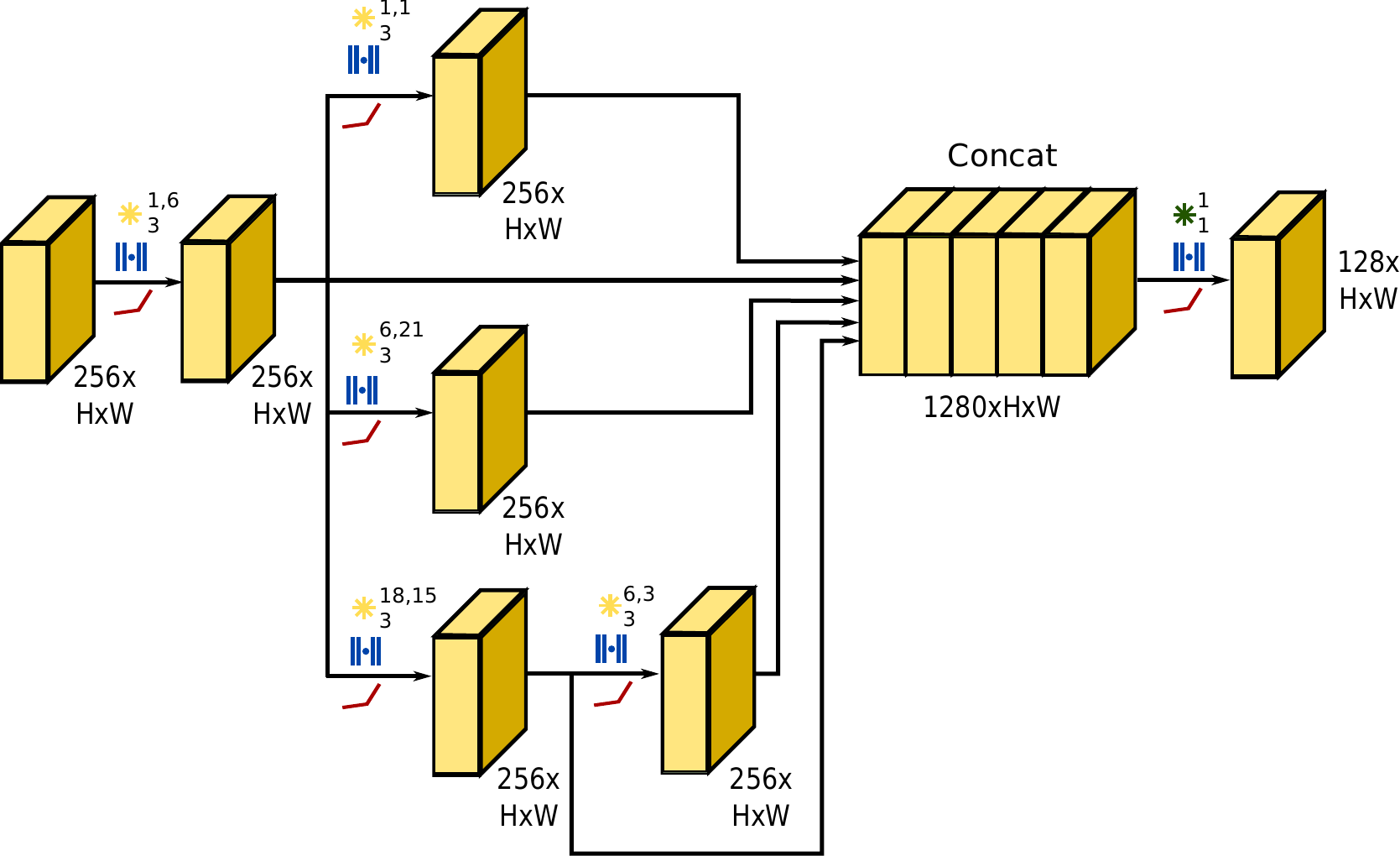}} \\
\multicolumn{3}{c}{(d) DPC Module} \\
\\
\multicolumn{3}{c}{\includegraphics[width=0.975\linewidth]{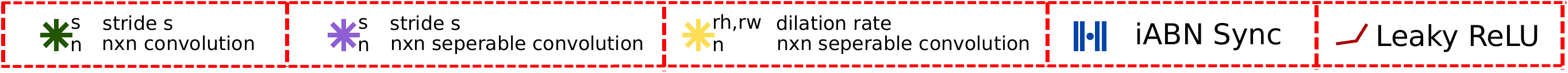}} \\
\end{tabular}}
\caption{Topologies of various architectural components in our proposed semantic head and instance head of our EfficientPS architecture.}
\label{fig:arch2}
\end{figure}

\subsection{Semantic Segmentation Head}
\label{sec:semanticHead}

Our proposed semantic segmentation head consists of three components, each aimed at targeting one of the critical requirements. First, at large-scale, the network should have the ability to capture fine features efficiently. In order to enable this, we employ our Large Scale Feature Extractor (LSFE) module that has two $3\times3$ depthwise separable convolutions with 128 output filters, each followed by an iABN sync and a Leaky ReLU activation function. The first $3\times3$ depthwise separable convolution reduces the number of filters to 128 and the second $3\times3$ depthwise separable convolution further learns deeper features.

The second requirement is that at small-scale, the network should be able to capture long-range context. Modules inspired by Atrous Spatial Pyramid Pooling (ASPP)~\cite{chen2017deeplab} that are widely used in state-of-the-art semantic segmentation architectures have been demonstrated to be effective for this purpose. Dense Prediction Cells (DPC)~\citep{chen2018searching} and Efficient Atrous Spatial Pyramid Pooling (eASPP)~\citep{valada19ijcv} are two variants of ASPP that are significantly more efficient and also yield a better performance. We find that DPC demonstrates a better performance with a minor increase in the number of parameters compared to eASPP. Therefore, we employ a modified DPC module in our semantic head as shown in \figref{fig:epsnet_arch}. We augment the original DPC topology by replacing batch normalization layers with iABN sync, and ReLUs with Leaky ReLUs. The DPC module consists of a $3\times3$ depthwise separable convolution with 256 output channels having a dilation rate of (1,6) and extends out to five parallel branches. Three of the branches, each consist of a $3\times3$ dilated depthwise separable convolution with 256 outputs, where the dilation rates are (1,1), (6,21), and (18,15) respectively. The fourth branch takes the output of the dilated depthwise separable convolution with a dilation rate of (18,15), as input and passes it through another $3\times3$ dilated depthwise separable convolution with 256 output channels and a dilation rate of (6,3). The outputs from all these parallel branches are then concatenated to yield a tensor with 1280 channels. This tensor is then finally passed through a $1\times1$ convolution with 256 output channels and forms the output of the DPC module. Note that each of the convolutions in the DPC module is followed by a iABN sync and a Leaky ReLU activation function.

The third and final requirement for the semantic head is that it should be able to mitigate the mismatch between large-scale and small-scale features while performing feature aggregation. To this end, we employ our Mismatch Correction Module (MC) that correlates the small-scale features with respect to large-scale features. It consists of cascaded $3\times3$ depthwise separable convolutions with 128 output channels, followed by iABN sync with Leaky ReLU and a bilinear upsampling layer that upsamples the feature maps by a factor of 2. Figures~\ref{fig:arch2}~(a), \ref{fig:arch2}~(c) and \ref{fig:arch2}~(d) illustrate the topologies of these main components of our semantic head.

The four different scaled outputs of our 2-way FPN, namely P\textsubscript{4}, P\textsubscript{8}, P\textsubscript{16} and P\textsubscript{32} are the inputs to our semantic head. The small-scale inputs, P\textsubscript{32} and P\textsubscript{16} with downsampling factors of $\times32$ and $\times16$ are each fed into two parallel DPC modules. While the large-scale inputs, P\textsubscript{8} and P\textsubscript{4} with downsampling factors of $\times8$ and $\times4$ are each passed through two parallel LSFE modules. Subsequently, the outputs from each of these parallel DPC and LSFE modules are augmented with feature alignment connections and each of them is upsampled to x4 scale. These upsampled feature maps are then concatenated to yield a tensor with 512 channels which is then input to a $1\times1$ convolution with N\textsubscript{‘stuff’+‘thing’} output filters. This tensor is then finally upsampled by a factor of 4 and passed through a softmax layer to yield the semantic logits having the same resolution as the input image. Now, the feature alignment connections from the DPC and LSFE modules interconnect each of these outputs by element-wise summation as shown in \figref{fig:epsnet_arch}. We add our MC modules in the interconnections between the second DPC and LSFE as well as between both the LSFE connections. These correlation connections aggregate contextual information from small-scale features and characteristic large-scale features for better object boundary refinement. We use the weighted per-pixel log-loss~\citep{bulo2017loss} for training which is given by
\begin{equation}
\mathcal{L}_{pp}(\Theta) = -\sum_{ij}w_{ij}(p^*_{ij})\log{p_{ij}},
\end{equation}
 $p^*_{i,j}$ is the groundtruth for a given image, $p_{i,j}$ is the predicted probability for the pixel $(i,j)$ being assigned class $c \in p$, $w_{ij} = \frac{4}{WH}$ if pixel $(i,j)$ belongs to $25\%$ of the worst prediction, and $w_{ij} = 0$ otherwise. $W$ and $H$ are the width and height of the given input image. The overall semantic head loss is given by
\begin{equation}\label{eq:Lsemantic}
\mathcal{L}_{semantic}(\Theta)=\frac{1}{n}\sum L_{pp},
\end{equation}
where $n$ is the batch size. We present in-depth analysis of our semantic head in comparison other semantic heads commonly used in state-of-the-art architectures in \secref{sec:detailedSemantic}.

\subsection{Instance Segmentation Head}
\label{sec:instanceHead}

The instance segmentation head of our EfficientPS network shown in \figref{fig:epsnet_arch} has a topology similar to Mask R-CNN~\citep{he2017mask} with certain modifications. More specifically, we replace all the standard convolutions, batch normalization layers, and ReLU activations with depthwise separable convolution, iABN sync, and Leaky ReLU respectively. Similar to the rest of our architecture, we use depthwise separable convolutions instead of standard convolutions to reduce the number of parameters consumed by the network. This enables us to conserve $\SI{2.09}{\million}$ parameters in comparison to the conventional Mask R-CNN.

Mask R-CNN consists of two stages. In the first stage, the Region Proposal Network (RPN) module shown in Figure \ref{fig:arch2}~(b) employs a fully convolutional network to output a set of rectangular object proposals and an objectness score for the given input FPN level. Subsequently, ROI align~\citep{he2017mask} uses object proposals to extract features from FPN encodings by directly pooling features from the n$^{\text{th}}$ channel with a $14\times14$ spatial resolution bounded within a bounding box proposal. The features that are extracted then serve as input to the bounding box regression, object classification and mask segmentation networks. The logits output from the mask segmentation networks for each candidate bounding box proposal is then fused with the semantic logits in our proposed panoptic fusion module described in \secref{sec:fusion}.

\begin{figure*}
\centering
\includegraphics[width=\linewidth]{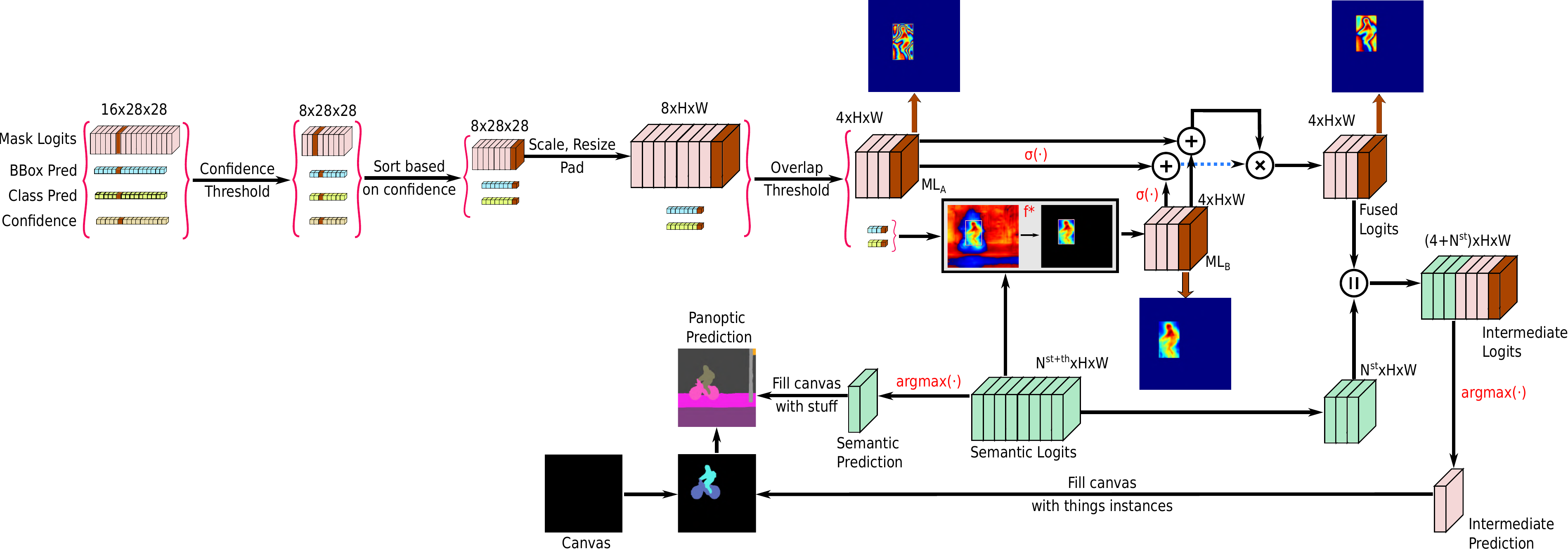}
\caption{Illustration of our proposed Panoptic Fusion Module. Here, ML\textsubscript{A} and ML\textsubscript{B} mask logits are fused as $(\sigma(ML_A) + \sigma(ML_B)) \odot (ML_A + ML_B))$, where ML\textsubscript{B} is output of the function $f^*$, $\sigma(\cdot)$ is the sigmoid function and $\odot$ is the Hadamard product. Here, the $f^*$ function for given class prediction $c$ (cyclist in this example), zeroes out the score of the $c$ channel of the semantic logits outside the corresponding bounding box. Please note that $16$ initial mask logits and $4$ instances are just arbitrary number taken for the sake of ease of explanation. The real values can and are much higher than these numbers.}
\label{fig:pfm}
\end{figure*}

In order to train the instance segmentation head, we adopt the loss functions proposed in Mask R-CNN, i.e. two loss functions for the first stage: objectness score loss and object proposal loss, and three loss functions for the second stage: classification loss, bounding box loss and mask segmentation loss. We take a set of randomly sampled positive matches and negative matches such that $|N_s|\leq256$. The objectness score loss $\mathcal{L}_{os}$ defined as log loss for a given $N_s$ is given by
\begin{align}\label{eq:os}
\mathcal{L}_{os}(\Theta) = &-\frac{1}{|N_s|}\sum_{(p^*_{os},p_{os})\in N_s} p^*_{os}\cdot\log p_{os}\nonumber\\ &+ (1-p^*_{os})\cdot\log (1-p_{os}),
\end{align}
where $p_{os}$ is the output of the objectness score branch of RPN and $p^*_{os}$ is the groundtruth label which is 1 if the anchor is positive, and 0 if the anchor is negative. We use the same strategy as Mask R-CNN for defining positive and negative matches. For a given anchor $a$, if the groundtruth box $b^*$ has the largest Intersection over Union (IoU) or IoU$(b^*,a) > T_H$, then the corresponding prediction $b$ is a positive match and $b$ is a negative match when IoU$(b^*,a) < T_L$. The thresholds $T_H$ and $T_L$ are pre-defined where $T_H > T_L$.

The object proposal loss $\mathcal{L}_{op}$ is a regression loss that is defined only on positive matches and is given by
\begin{equation}\label{eq:op}
\mathcal{L}_{op}(\Theta) = \frac{1}{|N_s|} \sum_{(t^*,t)\in N_p} \sum_{(i*,i)\in (t^*,t)} L_1(i*,i),
\end{equation}
where $L_1$ is the smooth L1 Norm, $N_p$ is the subset of $N_s$ positive matches, $t^* = (t_x^*, t_y^*, t_w^*, t_h^*)$ and $t = (t_x, t_y, t_w, t_h)$ are the parameterizations of $b^*$ and $b$ respectively, $b^* = (x^*, y^*, w^*, h^*)$ is the groundtruth box, $b^* = (x, y, w, h)$ is the predicted bounding box, $x, y, w$ and $h$ are the center coordinates, width and height of the predicted bounding box. Similarly,  $x^*, y^*, w^*$ and $h^*$ denote the center coordinates, width and height of the groundtruth bounding box. The parameterizations~\citep{girshick2015fast} are given by
\begin{equation}
t_x = \frac{(x-x_a)}{w_a},
t_y = \frac{(y-y_a)}{h_a}, \\
t_w = \log\frac{w}{w_a}, \\
t_h = \log\frac{h}{h_a}, \\
\end{equation}
\begin{equation}
t_x^* = \frac{(x^*-x_a)}{w_a},
t_y^* = \frac{(y^*-y_a)}{h_a}, \\
t_w^* = \log\frac{w^*}{w_a}, \\
t_h^* = \log\frac{h^*}{h_a}, \\
\end{equation}
where $x_a, y_a, w_a$ and $h_a$ denote the center coordinates, width and height of the anchor $a$.

Similar to the objectness score loss $\mathcal{L}_{os}$, the classification loss $\mathcal{L}_{cls}$ is defined for a set of $K_s$ randomly sampled positive and negative matches such that $|K_s| \leq 512$. The classification loss $\mathcal{L}_{cls}$ is given by
\begin{equation}
\mathcal{L}_{cls}(\Theta) = -\frac{1}{|K_s|}\sum_{c=1}^{N_{‘thing’+1}} Y^*_{o,c}\cdot\log Y_{o,c}, \textrm{for} (Y^*,Y)\in K_s,
\end{equation}
where $Y$ is the output of the classification branch, $Y^*$ is the one hot encoded groundtruth label, $o$ is the observed class, and $c$ is the correct classification for object $o$. For a given image, it is a positive match if IoU$(b^*,b) > T_n$ and otherwise a negative match, where $b^*$ is the groundtruth box, and $b$ is the object proposal from the first stage.

The bounding box loss $\mathcal{L}_{bbx}$ is a regression loss that is defined only on positive matches and is expressed as
\begin{equation}
\mathcal{L}_{bbx}(\Theta) = \frac{1}{|K_s|}\sum_{(T^*,T)\in K_p} \sum_{(i*,i)\in (T^*,T)}L_1(i*,i),
\end{equation}
where $L_1$ is the smooth L1 Norm~\citep{girshick2015fast}, $K_p$ is the subset of $K_s$ positive matches, $T^*$ and $T$ are the parameterizations of $B^*$ and $B$ respectively, similar to \eqref{eq:os} and~(\ref{eq:op}) where $B^*$ is the groundtruth box, and $B$ is the corresponding predicted bounding box.

Finally, the mask segmentation loss is also defined only for positive samples and is given by
\begin{equation}
\mathcal{L}_{mask}(\Theta) = -\frac{1}{|K_s|}\sum_{(P^*,P)\in K_s}L_{p}(P^*,P),
\end{equation}
where $L_{p}(P^*,P)$ is given as
\begin{align}
L_{p}(P^*,P) = &-\frac{1}{|T_p|}\sum_{(i,j)\in T_p} P_{i,j}^*\cdot\log P_{i,j}\nonumber\\ &+ (1-P_{i,j}^*)\cdot\log (1-P_{i,j}),
\end{align}
where $P$ is the predicted $28\times28$ binary mask for a class with $P_{i,j}$ denoting the probability of the mask pixel $(i,j)$, $P^*$ is the $28\times28$ groundtruth binary mask for the class, and $T_p$ is the set of non-void pixels in $P^*$.

All the five losses are weighed equally and the total instance segmentation head loss is given by
\begin{equation}\label{eq:Linstance}
\mathcal{L}_{instance} = \mathcal{L}_{os} + \mathcal{L}_{op} + \mathcal{L}_{cls} + \mathcal{L}_{bbx} + \mathcal{L}_{mask}.
\end{equation}
Similar to Mask R-CNN, the gradient that is computed w.r.t to the losses $\mathcal{L}_{cls}$, $\mathcal{L}_{bbx}$ and $\mathcal{L}_{mask}$ flow only through the network backbone and not through the region proposal network.

\subsection{Panoptic Fusion Module}
\label{sec:fusion}

In order to obtain the panoptic segmentation output, we need to fuse the prediction of the semantic segmentation head and the instance segmentation head. However, fusing both these predictions is not a straightforward task due to the inherent overlap between them. Therefore, we propose a novel panoptic fusion module to tackle the aforementioned problem in an adaptive manner in order to thoroughly exploit the predictions from both the heads congruously. \figref{fig:pfm} shows the topology of our panoptic fusion module. We obtain a set of object instances from the instance segmentation head of our network where for each instance, we have its corresponding class prediction, confidence score, bounding box and mask logits. First, we reduce the number of predicted object instances in two stages. We begin by discarding all object instances that have a confidence score of less than a certain confidence threshold. We then resize, zero pad and scale the $28\times28$ mask logits of each object instance to the same resolution as the input image. Subsequently, we sort the class prediction, bounding box and mask logits according to the respective confidence scores. In the second stage, we check each sorted instance mask logit for overlap with other object instances. To do so we compute the sigmoid of the mask logits and threshold it at 0.5 to obtain the corresponding binary mask. Then if the overlap between the binary masks is greater than a given overlap threshold, the mask logits with the highest confidence are retained and the other overlapping mask logits are discarded.

After filtering the object instances, we have the class prediction, bounding box prediction and mask logit $ML_A$ of each instance. We simultaneously obtain semantic logits with $N$ channels from the semantic head, where $N$ is the sum of $N_{‘stuff’}$ and $N_{‘thing’}$. We then compute a second mask logit $ML_{B}$ for each instance where we select the channel of the semantic logits based on its class prediction. We only keep the logit score of the selected channel for the area within the instance bounding box, while we zero out the scores that are outside this region. In the end, we have two mask logits for each instance, one from instance segmentation head and the other from the semantic segmentation head. We combine these two logits adaptively by computing the Hadamard product of the sum of sigmoid of $ML_{A}$ and sigmoid of $ML_{B}$, and the sum of $ML_{A}$ and $ML_{B}$ to obtain the fused mask logits $FL$ of instances expressed as
\begin{equation}
\label{eq}
FL = (\sigma(ML_A) + \sigma(ML_B)) \odot (ML_A + ML_B),
\end{equation}
where $\sigma(\cdot)$ is the sigmoid function and $\odot$ is the Hadamard product. We then concatenate the fused mask logits of the object instances with the ‘stuff’ logits along the channel dimension to generate intermediate panoptic logits. Subsequently, we apply the argmax operation along the channel dimension to obtain the intermediate panoptic prediction. In the final step, we take a zero-filled canvas and first copy the instance-specific ‘thing’ prediction from the intermediate panoptic prediction. We then fill the empty parts of the canvas with ‘stuff’ class predictions by copying them from the predictions of the semantic head while ignoring classes that have an area smaller than a predefined threshold called minimum stuff area. This gives us the final panoptic segmentation output.

We fuse $ML_{A}$ and $ML_{B}$ instance logits in the aforementioned manner due to the fact that if both logits for a given pixel conform with each other, the final instance score will increase proportionately to their agreement or vice-versa. In case of agreement, the corresponding object instance will dominate or be superseded by other instances as well as the ‘stuff’ classes score. Similarly, in case of disagreement, the score of the given object instance will reflect the extent of their difference. Simply put, the fused logit score is either adaptively attenuated or amplified according to the consensus. We evaluate the performance of our proposed panoptic fusion module in comparison to other existing methods in the ablation study presented in \secref{sec:fusionEval}.

\section{Experimental Results}
\label{sec:experimentalResults}

In this section, we first describe the standard evaluation metrics that we adopt for empirical evaluations, followed by brief descriptions of the datasets that we benchmark on in \secref{sec:datasets}. We then present extensive quantitative comparisons and benchmarking results in \secref{sec:benchmarking}, and detailed ablation studies on the various proposed architectural components in \secref{sec:ablation}. Finally, we present qualitative comparisons and visualizations of panoptic segmentation on each of the datasets that we evaluate on in \secref{sec:qualitative} and \secref{sec:visualization} respectively.

We use PyTorch~\citep{paszke2019pytorch} for implementing all our architectures and we trained our models on a system with an Intel Xenon@2.20GHz processor and NVIDIA TITAN X GPUs. We use the standard Panoptic Quality (PQ) metric~\citep{kirillov2019panoptic} for quantifying the performance of our models. The PQ metric is computed as 
\begin{equation}
PQ = \frac{\sum_{(p,g) \in TP} IoU(p,g)} {|TP|+\frac{1}{2}|FP|+\frac{1}{2}|FN|},
\end{equation}
where $TP, FP, FN$ and $IoU$ are true positives, false positives, false negatives and the intersection-over-union. The $IoU$ is computed as $IoU = TP/(TP + FP + FN)$. We also report the Segmentation Quality (SQ) and Recognition Quality (RQ) metrics computed as
\begin{align}
SQ = \frac{\sum_{(p,g) \in TP}IoU(p,g)}{|TP|},\\
RQ = \frac{|TP|}{|TP|+\frac{1}{2}|FP|+\frac{1}{2}|FN|}.
\end{align}

Following the standard benchmarking criteria for pantoptic segmentation, we report PQ, SQ and RQ over all the classes in the dataset, and we also report them for the ‘stuff’ classes (PQ\textsuperscript{St}, SQ\textsuperscript{St}, RQ\textsuperscript{St}) and the ‘thing’ classes (PQ\textsuperscript{Th}, SQ\textsuperscript{Th}, RQ\textsuperscript{Th}). Additionally, for the sake of completeness, we report the Average Precision (AP), mean Intersection-over-Union (mIoU) for both ‘stuff’ and ‘thing’ classes, as well as the inference time and FLOPs for comparisons. The implementation of our proposed EfficientPS model and a live demo on various datasets is publicly available at \url{https://rl.uni-freiburg.de/research/panoptic}.

\subsection{Datasets}
\label{sec:datasets}

We benchmark our proposed EfficientPS for panoptic segmentation on four challenging urban scene understanding datasets, namely, Cityscapes~\citep{cordts2016cityscapes}, KITTI~\citep{Geiger2013IJRR}, Mapillary Vistas~\citep{neuhold2017mapillary}, and Indian Driving Dataset~\citep{varma2019idd}. The KITTI benchmark does not provide panoptic annotations, therefore to facilitate this work, we publicly release manually annotated panoptic groundtruth segmentation labels for the popular KITTI benchmark. These four diverse datasets contain images that range from congested city driving scenarios to rural scenes and highways. They also contain scenes in challenging perceptual conditions including snow, motion blur and other seasonal visual changes. We briefly describe the characteristics of these datasets in this section.

\begin{figure}
\centering
\footnotesize
\setlength{\tabcolsep}{0.1cm}
{\renewcommand{\arraystretch}{0.5}
\begin{tabular}{P{0.4cm}P{3.68cm} P{3.68cm}}
& \raisebox{-0.4\height}{RGB} & \raisebox{-0.4\height}{Panoptic Groundtruth} \\
\\
\rot{(a) Cityscapes} & \raisebox{-0.4\height}{\includegraphics[width=\linewidth]{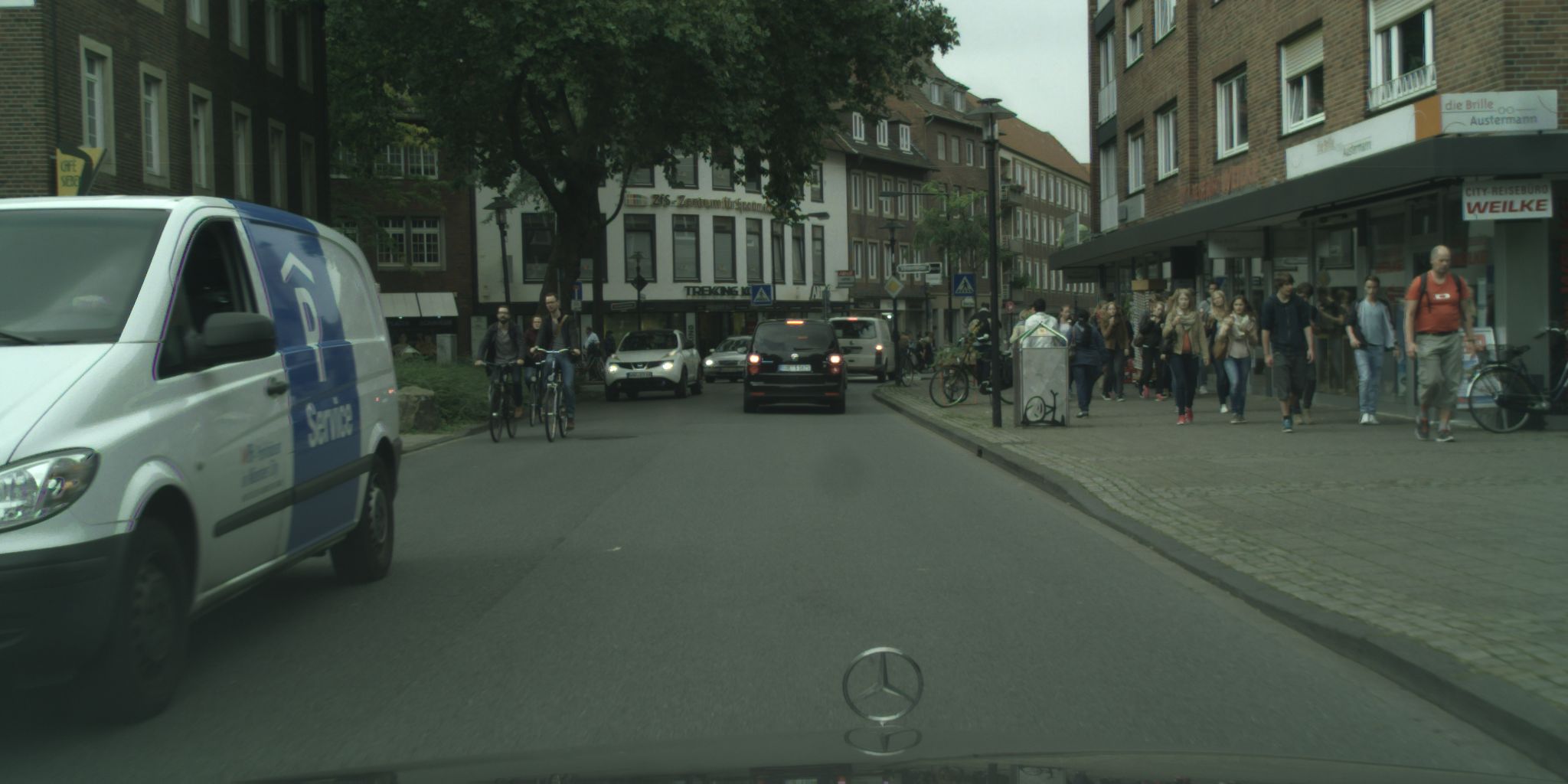}} & 
\raisebox{-0.4\height}{\includegraphics[width=\linewidth]{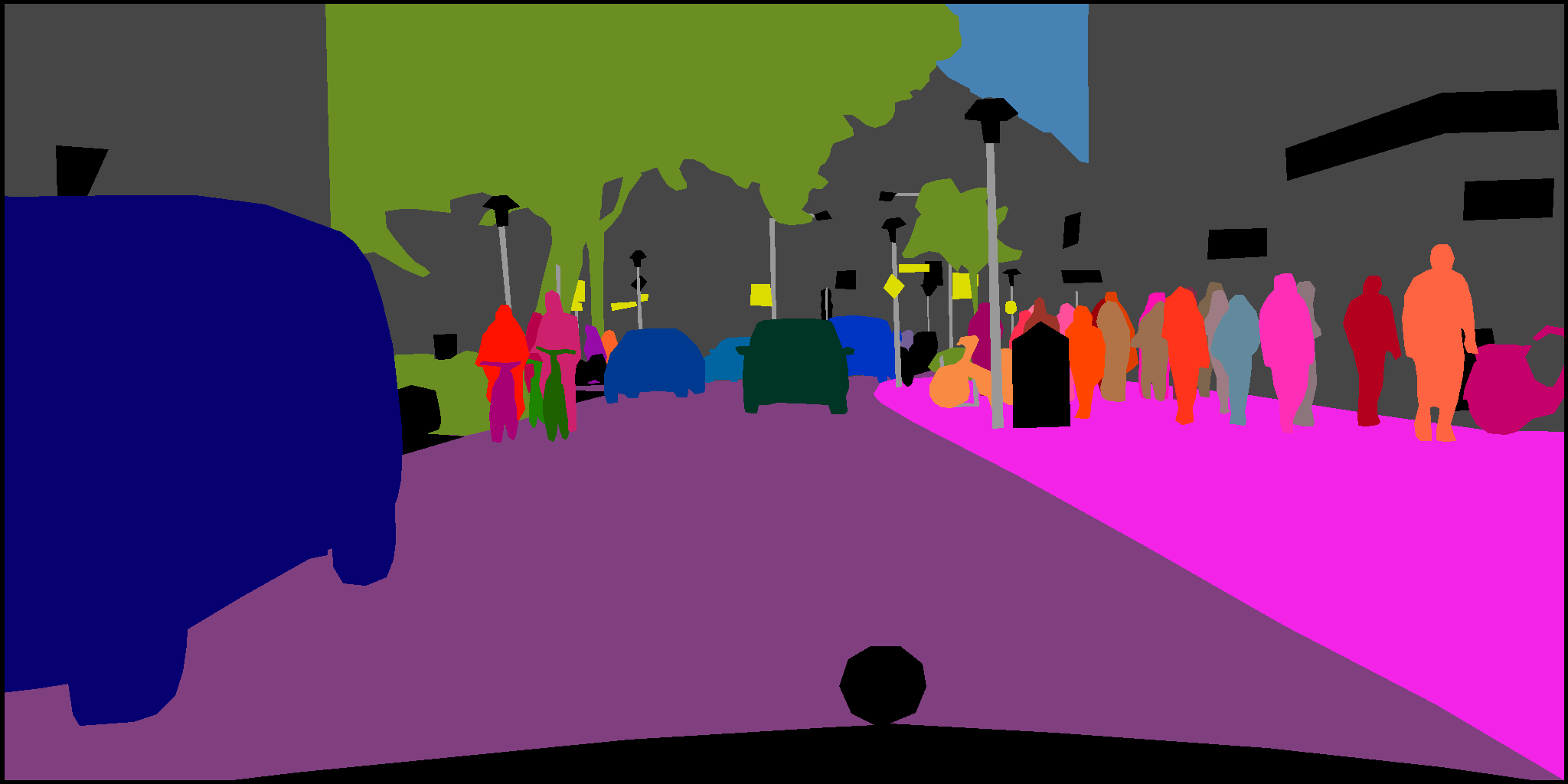}} \\
\\
\rot{(b) KITTI} & \raisebox{-0.4\height}{\includegraphics[width=\linewidth]{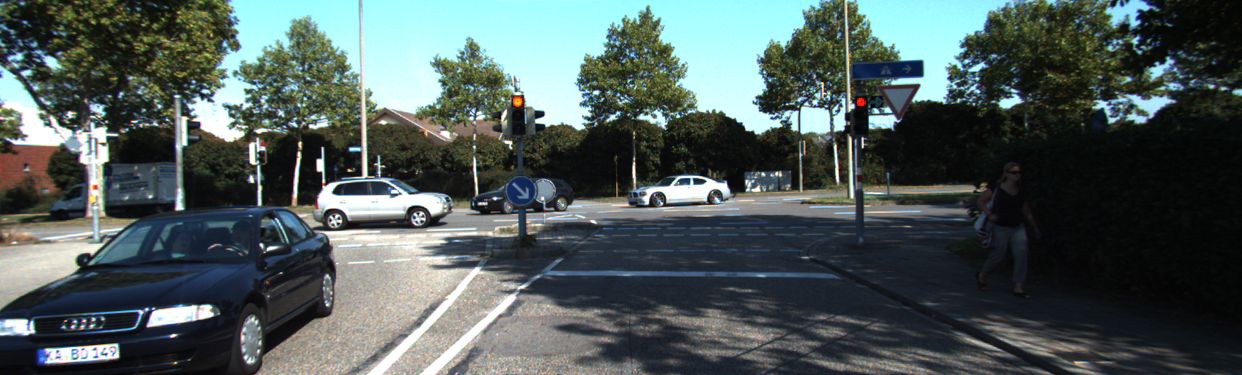}} & 
\raisebox{-0.4\height}{\includegraphics[width=\linewidth]{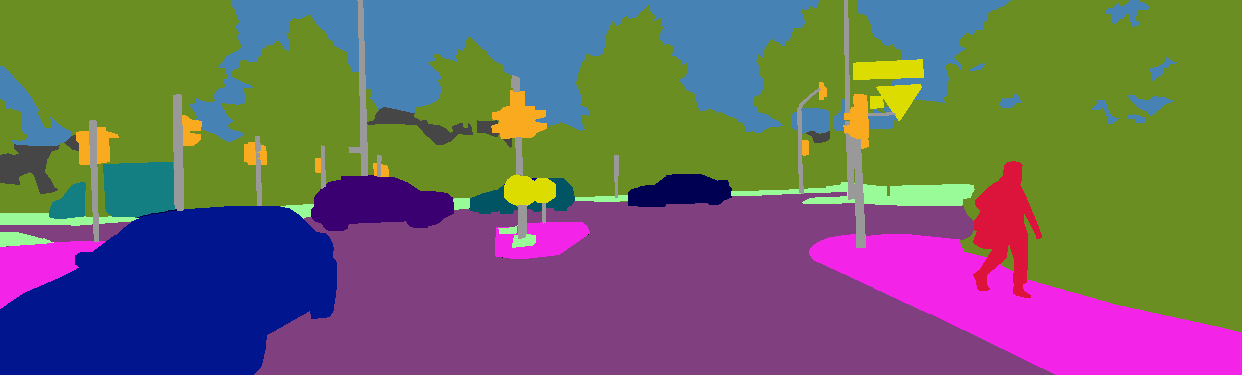}} \\
\\
\rot{(c) Mapillary Vistas} & \raisebox{-0.4\height}{\includegraphics[width=\linewidth]{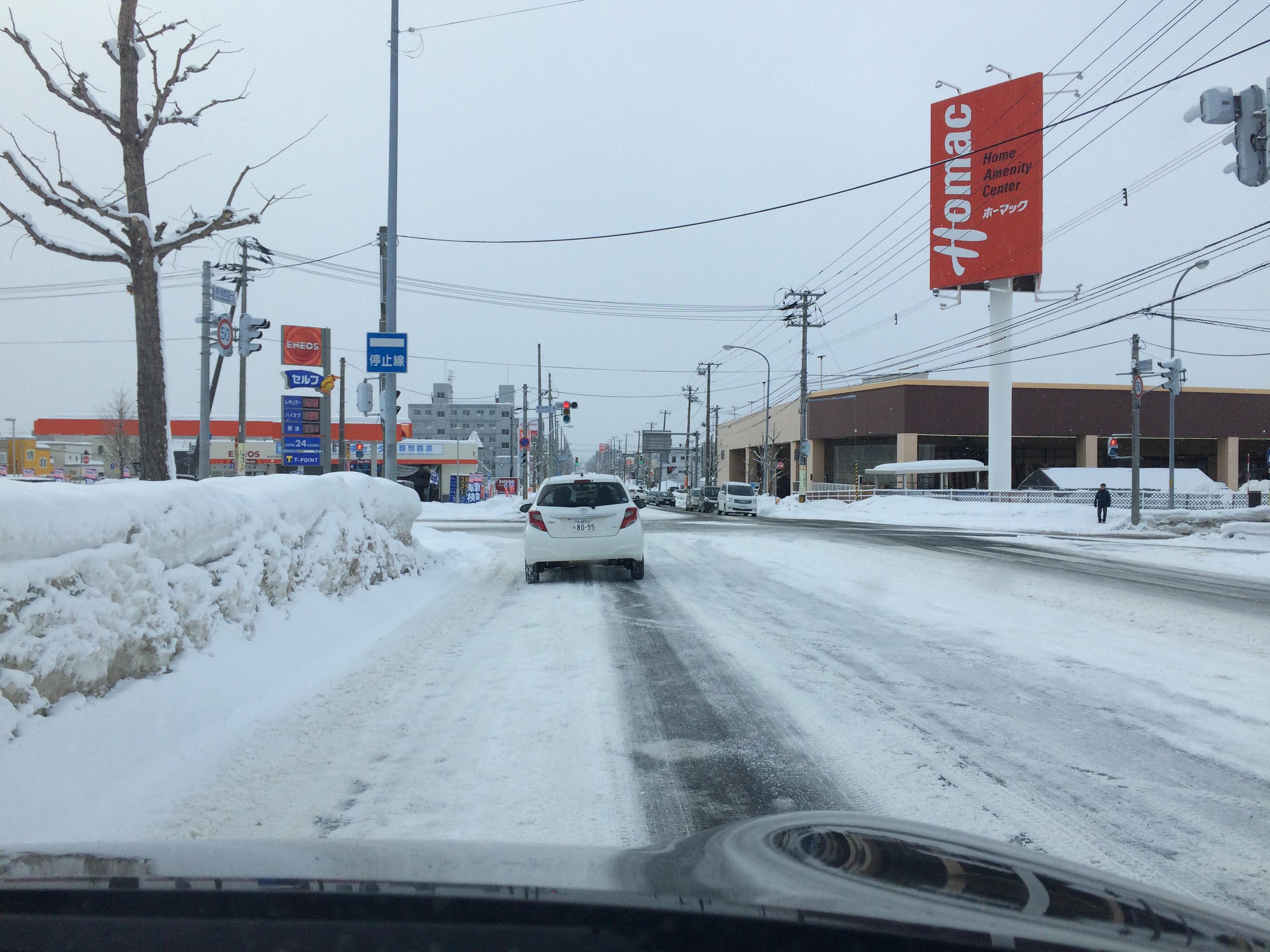}} &
\raisebox{-0.4\height}{\includegraphics[width=\linewidth]{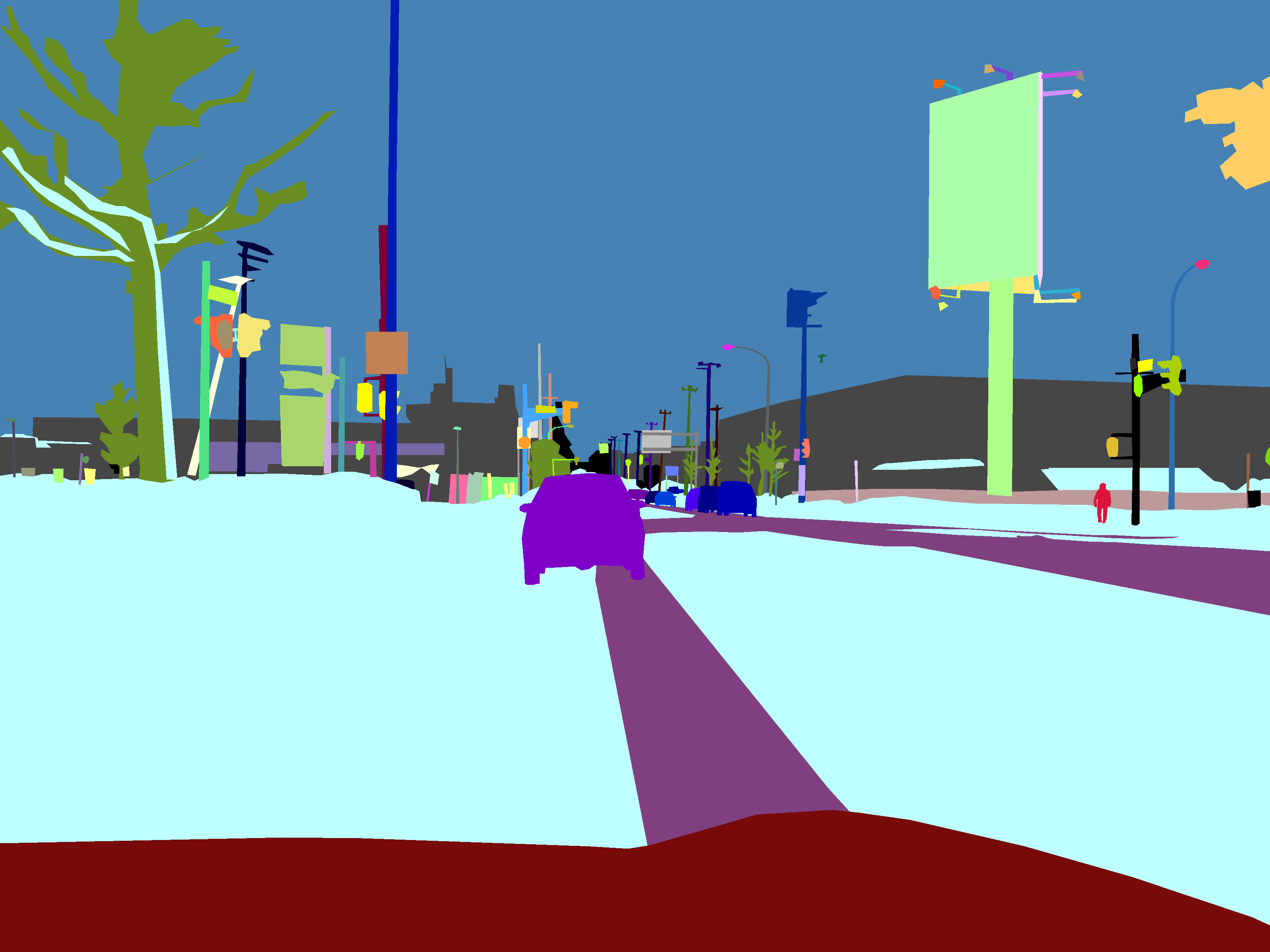}} \\
\\
\rot{(d) IDD} & \raisebox{-0.4\height}{\includegraphics[width=\linewidth]{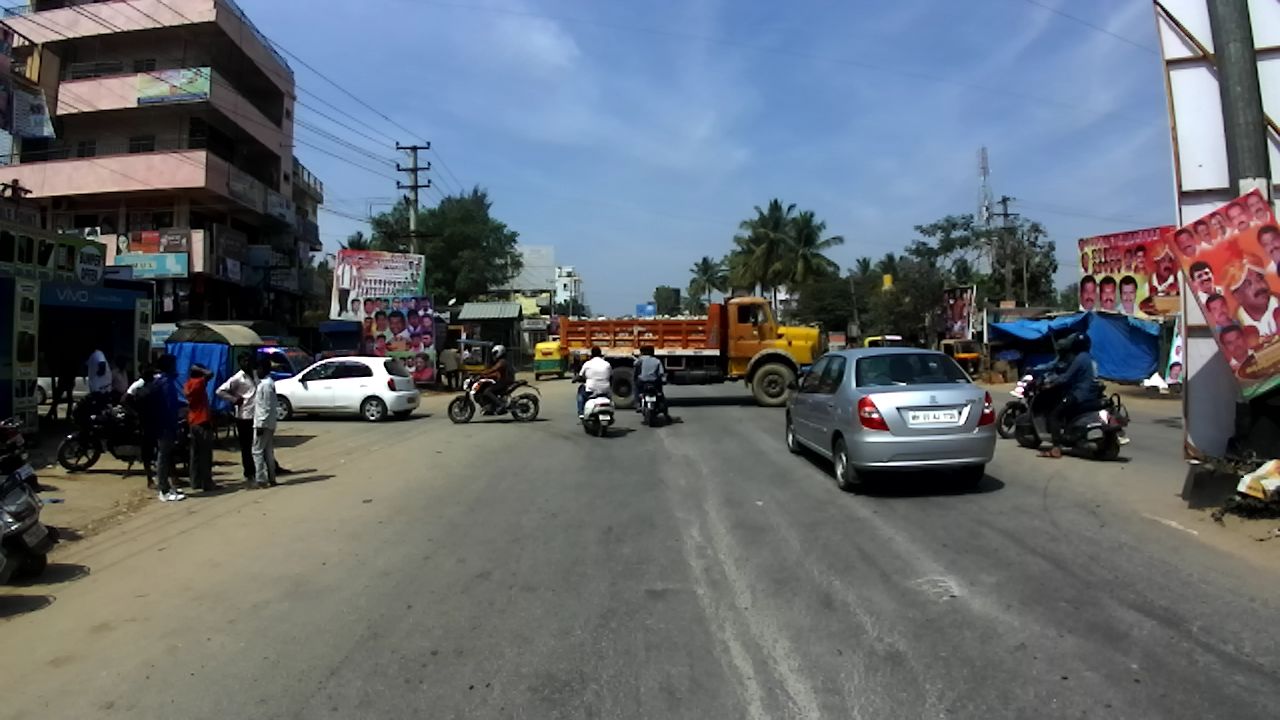}} &
\raisebox{-0.4\height}{\includegraphics[width=\linewidth]{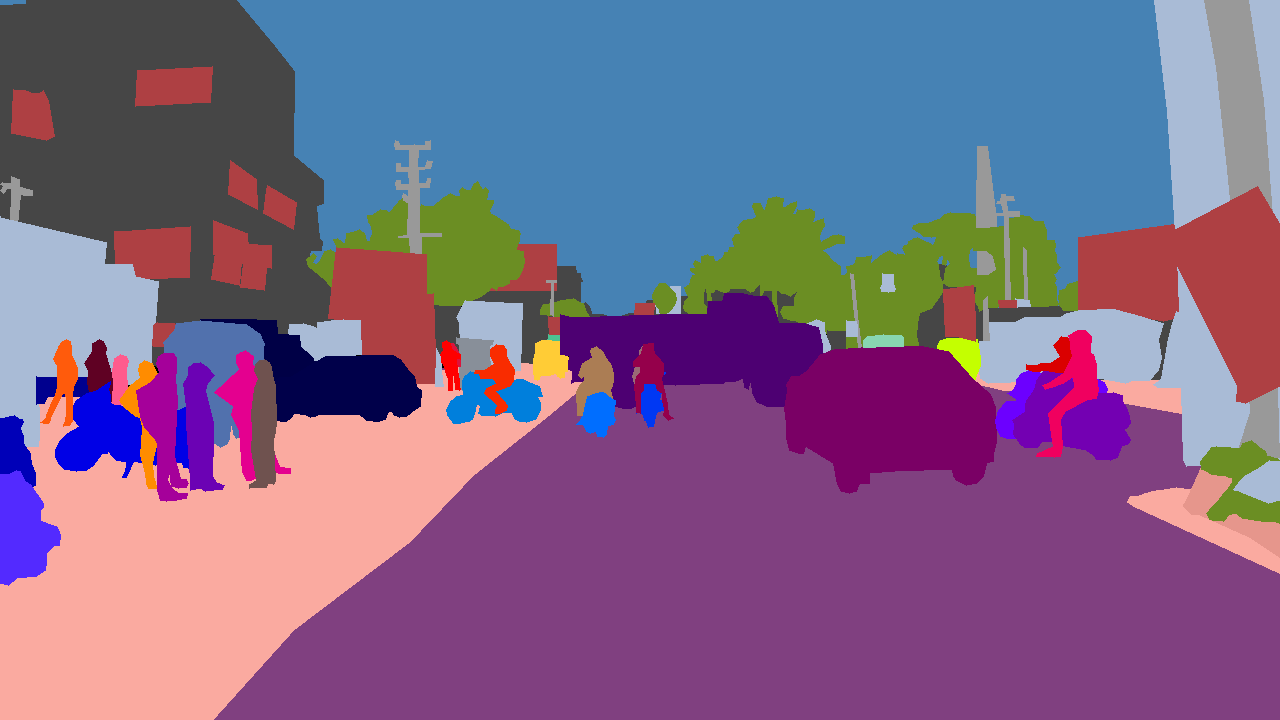}} \\
\end{tabular}}
\caption{Example images from the challenging urban scene understanding datasets that we benchmark on, namely, Cityscapes, KITTI, Mapillary Vistas, and Indian Driving Dataset (IDD). The images show cluttered urban scenes with many dynamic objects, occluded objects, perpetual snowy conditions and unstructured environments.}
\label{fig:datasetEx}
\end{figure}

{\parskip=5pt
\noindent\textbf{Cityscapes:} The Cityscapes dataset~\citep{cordts2016cityscapes} consists of urban street scenes and focuses on semantic understanding of common driving scenarios. It is one of the most challenging datasets for panoptic segmentation due to its sheer diversity as it covers scenes from over 50 European cities recorded over several seasons such as spring, summer and fall. The presence of a large number of dynamic objects further add to its complexity. \figref{fig:datasetEx}~(a) shows an example image and the corresponding panoptic groundtruth annotation from the Cityscapes dataset. As we see from this example, the scenes are extremely clutterd with many dynamic objects such as pedestrians and cyclists that are often grouped near one and another or partially occluded. These factors make panoptic segmentation, especially segmenting the ‘thing’ class exceedingly challenging.}

The widely used Cityscapes dataset recently introduced a benchmark for the task of panoptic segmentation. The dataset contains pixel-level annotations for 19 object classes of which 11 are ‘stuff’ classes and 8 are instance-specific ‘thing’ classes. It consists of 5000 finely annotated images and 20000 coarsely annotated images that were captured at a resolution of $2048\times1024$~pixels using an automotive-grade $\SI{22}{\centi\meter}$ baseline stereo camera. The finely annotated images are divided into 2975 for training, 500 for validation and 1525 for testing. The annotations for the test set are not publicly released, they are rather only available to the online evaluation server that automatically computes the metrics and publishes the results. We report the performance of our proposed EfficientPS on both the validation set as well as the test set. We also use the Cityscapes dataset for evaluating the improvement due to the various architectural contributions that we make in the ablation study. We report results on the validation set for our model trained only on the \textit{fine} annotations and we report the results on the test set from the benchmarking server for our model trained on both the \textit{fine} and \textit{coarse} annotations.

{\parskip=5pt
\noindent\textbf{KITTI:} The KITTI vision benchmark suite~\citep{Geiger2013IJRR} is one of the most comprehensive datasets that provides groundtruth for a variety of tasks such as semantic segmentation, scene flow estimation, optical flow estimation, depth prediction, odometry estimation, tracking and road lane detection. However, it still has not expanded its annotations to support the recently introduced panoptic segmentation task. The challenging nature of the KITTI scenes and its potential for benchmarking multi-task learning problems, makes extending this dataset to include panoptic annotations of great interest to the community. Therefore, in this work, we introduce the KITTI panoptic segmentation dataset for urban scene understanding that provides panoptic annotations for a subset of images from the KITTI vision benchmark suite. The annotations for the images that we provide do not intersect with the official KITTI semantic/instance segmentation test set, therefore in addition to panoptic segmentation, they can also be used as supplementary training data for benchmarking semantic or instance segmentation tasks individually.} 

Our dataset consists of a total of 1055 images, out of which 855 are used for the training set and 200 are used for the validation set. We provide annotations for 11~‘stuff’ classes and 8~‘thing’ classes adhering to the Cityscapes ‘stuff’ and ‘thing’ class distribution. In order to create panoptic annotations, we gathered semantic annotations from community driven extensions of KITTI~\citep{xu2016multimodal, ros2015vision} and combined them with the 200 training images from the KITTI semantic training set. We then manually annotated all the images with instance masks. We do so by manually drawing boundaries around the objects. We use an overlay of RGB and semantic segmentation image to guide the boundary drawing process. The pixels within the drawn boundaries in the semantic segmentation image are then labelled with a unique id to generate the corresponding instance segmentation mask. We create our simple annotation toolbox for labelling. We try to delineate objects as much as humanly possible otherwise treat the object as background or crowd in our annotations scheme. The instance masks are then merged with the semantic annotations to generate the panoptic segmentation ground truth labels. The images in our KITTI panoptic segmentation dataset are a resolution of $1280\times384$~pixels and contain scenes from both residential and inner city scenarios. \figref{fig:datasetEx}~(b) shows an example image from the KITTI panoptic segmentation dataset and its corresponding panoptic segmentation labels. We observe that the car denoted in teal color pixels and the van are both partially occluded by other ‘stuff’ classes such that they cause an object instance to be disjoint into two components. We find that scenarios such as these are extremely challenging for the task of panoptic segmentation as the disjoint object mask has to be assigned to the same instance ID. We hope that this dataset encourages innovative solutions to such real-world problems that are uncommon in other datasets and also accelerates research in multi-task learning for urban scene understanding.

{\parskip=5pt
\noindent\textbf{Mapillary Vistas:} Mapillary Vistas~\citep{neuhold2017mapillary} is one of the largest publicly available street-level imagery datasets that contains pixel-accurate and instance-specific semantic annotations. The novel aspects of this dataset include diverse scenes from over six continents and in a variety of weather conditions, season, time of day, cameras, and viewpoints. It consists of 18,000 images for training, 2,000 images for validation, and 5,000 images for testing. The dataset provides panoptic annotations for 37 ‘thing’ classes and 28 ‘stuff’ classes. The images in this dataset are of different resolutions, ranging from $1024\times768$~pixels to $4000\times6000$~pixels. \figref{fig:datasetEx}~(c) shows an example image and the corresponding panoptic segmentation groundtruth from the Mapillary Vistas dataset. We can see that due to the snowy condition, recognizing distant objects such as the car in this example becomes extremely difficult. Such drastic seasonal changes make this dataset one of the most challenging for panoptic segmentation.}

{\parskip=5pt
\noindent\textbf{Indian Driving Dataset:} The Indian Driving Dataset (IDD) \citep{varma2019idd} was recently introduced for scene understanding of unstructured environments. Unlike other urban scene understanding datasets, IDD consists of scenes that do not have well-delineated infrastructures such as lanes and sidewalks. It has a significantly more number of ‘thing’ instances in each scene compared to other datasets and it only has a small number of well-defined categories for traffic participants. The images in this dataset were captured with a front-facing camera mounted on a car and the data was gathered in two Indian cities as well as in their outskirts. IDD consists of a total of 10,003 images, where 6993 are used for training, 981 for validation and 2029 for testing. The images are a resolution of either $1920\times1080$~pixels or $720\times1280$~pixels. We train and evaluate all our models on 720p resolution on this dataset. The annotations are provided in four levels of hierarchy. Existing approaches primarily report their results for $level~3$, therefore we report the results of our model on the same to facilitate comparison. This level comprises of a total of 26 classes out of which 17 are ‘stuff’ classes and 9 are instance-specific ‘thing’ classes. An example image and the corresponding panoptic segmentation groundtruth from the IDD dataset is shown in \figref{fig:datasetEx}~(d). We observe that the transition between the road and the sidewalk class is structurally not well defined which often leads to misclassifications. Factors such as this, make evaluating on this dataset uniquely challenging.}

\subsection{Training Protocol}
\label{sec:training}

We train our network on crops of different resolutions of the input image, namely, $1024\times2048$, $1024\times1024$, $384\times1280$, and $720\times1280$~pixels. We take crops from the full resolution of the image provided in each of the datasets. We perform a limited set of random data augmentations including flipping and scaling within the range of $[0.5, 2.0]$. We initialize the backbone of our EfficientPS with weights from the EfficientNet model pre-trained on the ImageNet dataset~\citep{russakovsky2015imagenet} and initialize the weights of the iABN sync layers to 1. We use Xavier initialization~\citep{glorot2010understanding} for the other layers, zero constant initialization for the biases and we use Leaky ReLU with a slope of 0.01. We use the same hyperparameters as \cite{girshick2015fast} for our instance head and additionally set $T_H = 0.7$, $T_L = 0.3$, and $T_N = 0.5$. In our proposed panoptic fusion module, we use a confidence threshold of $c_t = 0.5$, overlap threshold of $o_t = 0.5$ and minimum stuff area of $min_{sa} = 2048$.

\begin{table*}
\footnotesize
\centering
\caption{Performance comparison of panoptic segmentation on the Cityscapes validation set. Superscripts St and Th refer to ‘stuff’ and ‘thing’ classes respectively. $-$ denotes that the metric has not been reported for the corresponding method.}
\label{tab:baselineCityscapes}
\begin{tabular}{p{0.5cm}p{2.7cm}p{1.6cm}|p{0.5cm}p{0.5cm}p{0.5cm}p{0.5cm}p{0.5cm}p{0.5cm}p{0.5cm}p{0.5cm}p{0.5cm}p{0.5cm}p{0.6cm}}
\toprule
Mode & Network & Pre-training & PQ & SQ & RQ &PQ\textsuperscript{Th} & SQ\textsuperscript{Th} & RQ\textsuperscript{Th} & PQ\textsuperscript{St} & SQ\textsuperscript{St} & RQ\textsuperscript{St}& AP & mIoU  \\
&  &  & $(\%)$ & $(\%)$ & $(\%)$ & $(\%)$ & $(\%)$ & $(\%)$ & $(\%)$ & $(\%)$ & $(\%)$ & $(\%)$ & $(\%)$ \\
\noalign{\smallskip}\hline\hline\noalign{\smallskip}
\multirow{16}{*}{\rotatebox[origin=0]{90}{Single-Scale}} & WeaklySupervised &  & $47.3$ & $-$ & $-$ & $39.6$ & $-$ & $-$ &$52.9$ & $-$ & $-$ &  $24.3$ & $71.6$  \\
& TASCNet &  & $55.9$ & $-$ & $-$ & $50.5$ & $-$ & $-$ & $59.8$ & $-$ & $-$ & $-$ & $-$ \\ 
& Panoptic FPN & & $58.1$ & $-$ & $-$ & $52.0$ & $-$ & $-$ &$62.5$ & $-$ & $-$ &  $33.0$ & $75.7$ \\
& AUNet & & $59.0$ &  $-$ &  $-$ & $54.8$ & $-$ & $-$ & $62.1$ & $-$ & $-$ & $34.4$ & $75.6$  \\
& UPSNet & & $59.3$ & $79.7$  & $73.0$ & $54.6$ &  $79.3$ & $68.7$& $62.7$ & $80.1$ &  $76.2$  & $33.3$ & $75.2$ \\ 
& DeeperLab & & $56.3$ & $-$ & $-$ & $-$ & $-$ & $-$ & $-$ & $-$ & $-$ & $-$ & $-$ \\ 
& Seamless & & $60.3$ & $-$ & $-$ & $56.1$ & $-$ & $-$ & $63.3$ & $-$ & $-$ & $33.6$ & $77.5$ \\  
& SSAP & & $61.1$ & $-$ & $-$ & $55.0$ & $-$ & $-$ & $-$ & $-$ & $-$ & $-$ & $-$ \\  
& AdaptIS & & $62.0$ & $-$ & $-$ & $58.7$ & $-$ & $-$ &$64.4$ & $-$ & $-$ & $36.3$ & $79.2$ \\
& Panoptic-DeepLab & & $63.0$ & $-$ & $-$ & $-$ & $-$ & $-$ & $-$ & $-$ & $-$ & $35.3$ & $\mathbf{80.5}$ \\
\cmidrule{2-14}
&\textbf{EfficientPS (ours)} & & $\mathbf{63.9}$ & $\mathbf{81.5}$  & $\mathbf{77.1}$ & $\mathbf{60.7}$ & $\mathbf{81.2}$ &$\mathbf{74.1}$&$\mathbf{66.2}$ &$\mathbf{81.8}$ & $\mathbf{79.2}$ &$\mathbf{38.3}$ & $79.3$ \\
\cline{2-14}
\cmidrule{2-14}
& TASCNet & COCO & $59.3$ & $-$ & $-$ &$56$ & $-$ & $-$ &  $61.5$ & $-$ & $-$ & $37.6$ & $78.1$  \\ 
& UPSNet & COCO & $60.5$ & $80.9$ & $73.5$ & $57.0$ & $-$ & $-$ & $63.0$ & $-$ & $-$ & $37.8$ & $77.8$  \\ 
& Seamless & Vistas & $65.0$ & $-$ & $-$ & $60.7$ & $-$ & $-$ & $68.0$  & $-$ & $-$ & $-$ & $80.7$  \\ 
& Panoptic-Deeplab & Vistas & $65.3$ & $-$ & $-$ & $-$ & $-$ & $-$ & $-$ & $-$ & $-$ & $38.8$ & $\mathbf{82.5}$  \\ 
\cmidrule{2-14}
& \textbf{EfficientPS (ours)} & Vistas & $\mathbf{66.1}$ & $\mathbf{82.5}$ & $\mathbf{78.9}$ & $\mathbf{62.7}$  & $\mathbf{81.9}$ &$\mathbf{75.2}$ & $\mathbf{68.5}$ &$\mathbf{82.9}$  & $\mathbf{81.6}$ &$\mathbf{41.9}$ & $81.0$ \\
\noalign{\smallskip}\hline\hline\noalign{\smallskip}
\multirow{8}{*}{\rotatebox[origin=0]{90}{Multi-Scale}} 
& Panoptic-DeepLab & & $64.1$ & $-$ & $-$ & $-$ & $-$ & $-$ & $-$ & $-$ & $-$ & $38.5$ & $\mathbf{81.5}$  \\
\cmidrule{2-14}
&\textbf{EfficientPS (ours)} & & $\mathbf{65.1}$ & $\mathbf{82.2}$ &$\mathbf{79.0}$ & $\mathbf{61.5}$& $\mathbf{81.4}$ &$\mathbf{75.4}$ &$\mathbf{67.7}$ &$\mathbf{82.8}$ &$\mathbf{81.7}$ &$\mathbf{39.7}$ & $80.3$ \\
\cline{2-14}
\cmidrule{2-14}
& TASCNet& COCO & $60.4$ & $-$ & $-$ & $56.1$ & $-$ & $-$ &$63.3$ & $-$ & $-$ & $39.1$ & $78.7$  \\
& M-RCNN + PSPNet   & COCO & $61.2$ & $80.9$ & $74.4$ &$54.0$ & $-$ & $-$ &$66.4$ & $-$ & $-$ & $36.4$ & $80.9$  \\
& UPSNet & COCO  & $61.8$ & $81.3$ & $74.8$ &$57.6$ & $77.7$ & $70.5$ & $64.8$& $81.4$ &  & $39.0$ & $79.2$  \\
& Panoptic-Deeplab & Vistas & $67.0$ & $-$ & $-$ & $-$ & $-$ & $-$ & $-$ & $-$ & $-$ & $42.5$ & $\mathbf{83.1}$  \\
\cmidrule{2-14}
&\textbf{EfficientPS (ours)} & Vistas & $\mathbf{67.5}$ & $\mathbf{83.2}$  & $\mathbf{80.2}$ & $\mathbf{63.5}$ & $\mathbf{82.2}$ &$\mathbf{77.2}$  &$\mathbf{70.4}$ &$\mathbf{83.9}$ &$\mathbf{82.4}$ &$\mathbf{43.8}$ & $82.1$  \\
\bottomrule
\end{tabular}
\end{table*}

We train our model with Stochastic Gradient Descent (SGD) with a momentum of $0.9$ using a multi-step learning rate schedule i.e. we start with an initial base learning rate and train the model for a certain number of iterations, followed by lowering the learning rate by a factor of 10 at each milestone and continue training until convergence. We denote the base learning rate $lr_{base}$, milestones and the total number of iterations $ti$ for each dataset in the following format: $\{lr_{base},\{milestone, milestone\}, ti\}$. The training schedule for Cityscapes, Mapillary Vistas, KITTI and IDD are \{0.07, \{32K, 44K\}, 50K\}, \{0.07, \{144K, 176K\}, 192K\}, \{0.07, \{16K, 22K\}, 25K\} and \{0.07 ,\{108K, 130K\}, 144K\} respectively. At the beginning of the training, we have a warm-up phase where the $lr_{base}$ is increased linearly from $\frac{1}{3}\cdot lr_{base}$ to $lr_{base}$ in 200 iterations. Aditionally, we freeze the iABN sync layers and further train the model for 10 epochs with a fixed learning rate of $lr = 10^{-4}$. The final loss $\mathcal{L}_{total}$ that we optimize is computed as
\begin{equation}
\mathcal{L}_{total} = \mathcal{L}_{semantic} + \mathcal{L}_{instance},
\end{equation}
where $\mathcal{L}_{semantic}$ and $\mathcal{L}_{instance}$ are given in \eqref{eq:Lsemantic} and \eqref{eq:Linstance} respectively. We train our EfficientPS with a batch size of 16 on 16 NVIDIA Titan X GPUs where each GPU tends to a single-image.

\subsection{Benchmarking Results}
\label{sec:benchmarking}

\begin{table*}
\footnotesize
\centering
\caption{Comparison of panoptic segmentation benchmarking results on the Cityscapes test set. Superscripts St and Th refer to ‘stuff’ and ‘thing’ classes respectively.}
\label{tab:baselineCityscapesTest}
\begin{tabular}{p{3cm}p{2.5cm}|p{0.7cm}p{0.7cm}p{0.7cm}p{0.7cm}p{0.6cm}}
\toprule
Network & Pre-training & PQ  & SQ & RQ & PQ\textsuperscript{Th}& PQ\textsuperscript{St} \\
 &  & $(\%)$ & $(\%)$ & $(\%)$ & $(\%)$ & $(\%)$ \\
\noalign{\smallskip}\hline\hline\noalign{\smallskip}
SSAP &  & $58.9$ & $82.4$ & $70.6$& $48.4$ &  $66.5$ \\
TASCNet & COCO & $60.7$ &  $81.0$ & $73.8$ & $53.4$& $66.0$ \\
Panoptic-Deeplab & & $62.3$ & $82.4$ & $74.8$ & $52.1$&$\mathbf{69.7}$ \\ 
Seamless & Vistas & $62.6$ & $82.1$ & $75.3$& $56.0$ & $67.5$ \\
Panoptic-Deeplab & Vistas & $66.5$ & $83.5$ & $78.8$ & $58.8$ & $\mathbf{72.0}$ \\ 
\midrule
\textbf{EfficientPS (ours)} & & $\mathbf{64.1}$ & $\mathbf{82.6}$ & $\mathbf{76.8}$ & $\mathbf{56.7}$ & $69.4$ \\
\textbf{EfficientPS (ours)} & Vistas & $\mathbf{67.1}$ & $\mathbf{83.4}$ & $\mathbf{79.6}$ & $\mathbf{60.9}$ & $71.6$ \\
\bottomrule
\end{tabular}
\end{table*}

\begin{table}
\footnotesize
\centering
\caption{Comparison of model efficiency with both state-of-the-art top-down and bottom-up panoptic segmentation architectures.}
\label{tab:runtimesCityscapes}
\begin{tabular}{p{2.3cm}p{1.7cm}p{0.8cm}p{0.8cm}p{0.5cm}}
\toprule
Network & Input Size & Params. & FLOPs & Time \\
& (pixels) & $(\si{\million})$ & $(\si{\billion})$ & $(\si{\milli\second})$ \\
\noalign{\smallskip}\hline\hline\noalign{\smallskip}
DeeperLab & $1025\times2049$ & $-$ & $-$ & $463$   \\
UPSNet & $1024\times2048$ & $45.05$ & $487.02$ & $202$   \\
Seamless & $1024\times2048$ &$51.43$ & $514.00$ & $168$  \\
Panoptic-Deeplab & $1025\times2049$ &$46.73$ & $547.49$ & $175$ \\ 
\midrule
\textbf{EfficientPS (ours)} & $1024\times2048$ & $\mathbf{40.89}$ & $\mathbf{433.94}$ & $\mathbf{166}$ \\
\bottomrule
\end{tabular}
\end{table}

In this section, we report results comparing the performance of our proposed EfficientPS architecture against current state-of-the-art panoptic segmentation approaches. For comparisons on the Cityscapes and Mapillary Vistas datasets, we directly report the performance metrics of the state-of-the-art methods as stated in their corresponding manuscripts. While for KITTI and IDD, we report results for the models that we trained using the official implementations that have been publicly released by the authors after further tuning of hyperparameters to the best of our ability. Note that existing methods have not reported results on KITTI and IDD validation sets. We report results on the validation sets for all the datasets and we additionally report results on the test set for the Cityscapes dataset by evaluating them on the official server. Note that at the time of submission, only the Cityscapes benchmark has the provision to evaluate the results on the test set. On each of the datasets, we report both the single-scale and multi-scale evaluation results. Following standard practise, we perform horizontal flipping and scaling (scales of \{0.75, 1, 1.25, 1.5, 1.75, 2\}) during the multi-scale evaluations.

\begin{table*}
\footnotesize
\centering
\caption{Performance comparison of panoptic segmentation on the Mapillary Vistas validation set. Note that no additional data was used for training EfficientPS on this dataset other than pre-training the encoder on ImageNet. Superscripts St and Th refer to ‘stuff’ and ‘thing’ classes respectively. $-$ denotes that the metric has not been reported for the corresponding method.}
\label{tab:baselineMapillary}
\begin{tabular}{p{0.7cm}p{2.7cm}|p{0.5cm}p{0.5cm}p{0.5cm}p{0.5cm}p{0.5cm}p{0.5cm}p{0.5cm}p{0.5cm}p{0.5cm}p{0.5cm}p{0.6cm}}
\toprule
Mode & Network & PQ & SQ & RQ & PQ\textsuperscript{Th} & SQ\textsuperscript{Th}&  RQ\textsuperscript{Th} & PQ\textsuperscript{St} & SQ\textsuperscript{St} &  RQ\textsuperscript{St} & AP & mIoU \\
 &  & $(\%)$ & $(\%)$ & $(\%)$ & $(\%)$ & $(\%)$ & $(\%)$ & $(\%)$ & $(\%)$ & $(\%)$ & $(\%)$ & $(\%)$ \\
\noalign{\smallskip}\hline\hline\noalign{\smallskip}
\multirow{7}{*}{\rotatebox[origin=0]{90}{Single-Scale}} & JSIS-Net & $17.6$ & $55.9$ & $23.5$ & $10.0$ &  $47.6$ & $14.1$ & $27.5$ &  $66.9$ & $35.8$ & $-$ & $-$ \\
& DeeperLab & $32.0$ & $-$ & $-$ & $-$ & $-$ & $-$ & $-$ & $-$ & $-$ & $-$ & $55.3$ \\
& TASCNet & $32.6$ & $-$ & $-$ & $31.1$ &$-$ & $-$ & $34.4$ &$-$ & $-$ & $18.5$ & $-$ \\  
& AdaptIS & $35.9$ &   $-$ & $-$ & $31.5$ &$-$ & $-$ & $-$ & $41.9$ &$-$ & $-$ & $-$ \\
& Seamless & $37.7$ &  $-$ & $-$ &$33.8$ & $-$ & $-$ &$42.9$ &  $-$ & $-$ &$16.4$ & $50.4$  \\
 & Panoptic-DeepLab  & $37.7$ &  $-$ & $-$ & $30.4$ & $-$ & $-$ &$\mathbf{47.4}$ & $-$ & $-$ &$14.9$ & $\mathbf{55.3}$ \\
\cmidrule{2-13}
& \textbf{EfficientPS (ours)} & $\mathbf{38.3}$ & $\mathbf{74.2}$ &$\mathbf{48.0}$ &$\mathbf{33.9}$ &$\mathbf{73.3}$ &  $\mathbf{43.0}$ &  $44.2$ &  $\mathbf{75.4}$ &  $\mathbf{54.7}$ &  $\mathbf{18.7}$ & $52.6$ \\
\noalign{\smallskip}\hline\hline\noalign{\smallskip}
\multirow{4}{*}{\rotatebox[origin=0]{90}{Multi-Scale}}
& & \\
& TASCNet & $34.3$ & $-$ & $-$ & $34.8$ & $-$ & $-$ & $33.6$ &  $-$ & $-$ & $20.4$ & $-$  \\ 
& Panoptic-DeepLab & $40.3$ &  $-$ & $-$ & $33.5$ & $-$ & $-$ & $\mathbf{49.3}$ & $-$ & $-$ &$17.2$ & $\mathbf{56.8}$  \\
\cmidrule{2-13}
& \textbf{EfficientPS (ours)} & $\mathbf{40.5}$ & $\mathbf{74.9}$ &$\mathbf{49.5}$ & $\mathbf{35.0}$ &$\mathbf{73.8}$ &$\mathbf{44.4}$ & $47.7$ & $\mathbf{76.2}$ &  $\mathbf{56.4}$ &$\mathbf{20.8}$ & $54.1$ \\
\bottomrule
\end{tabular}
\end{table*}

We compare the performance of our proposed EfficientPS against state-of-the-art models on the Cityscapes dataset including WeaklySupervised~\citep{li2018weakly}, TASCNet~\citep{li2018learning}, Panoptic FPN~\citep{kirillov2019bpanoptic}, AUNet~\citep{li2019attention}, UPSNet~\citep{xiong2019upsnet}, DeeperLab~\citep{yang2019deeperlab}, Seamless~\citep{porzi2019seamless}, SSAP~\citep{gao2019ssap}, AdaptIS~\citep{sofiiuk2019adaptis}, and Panoptic-DeepLab \citep{cheng2019panoptic}. \tabref{tab:baselineCityscapes} shows the results on the Cityscapes validation set. For a fair comparison, we categorize models in the table separately according to those that report single-scale and multi-scale evaluation, as well as without any pre-training and pre-training on other datasets, namely Mapillary Vistas~\citep{neuhold2017mapillary} denoted as Vistas and Microsoft COCO~\citep{lin2014microsoft} abbreviated as COCO. We report the performance of all the aforementioned variants of our EfficientPS model. Note that we do not use the Cityscapes \textit{coarse} annotations, depth data or exploit temporal data. Our EfficientPS model trained only on the Cityscapes \textit{fine} annotations and with single-scale evaluation outperforms the previous best proposal based approach AdaptIS by $1.9\%$ in PQ and $2.0\%$ in AP, while outperforming the best bottom-up approach Panoptic-Deeplab by $0.9\%$ in PQ and $3.0\%$ in AP. Furthermore, our EfficientPS model trained only on the Cityscapes \textit{fine} annotations and with multi-scale evaluation achieves an improvement of $1.0\%$ in PQ and $1.2\%$ in AP over Panoptic-Deeplab. We observe a similar trend while comparing with models that have been pre-trained with additional data, where our proposed EfficientPS outperforms the former state-of-the-art Panoptic-Deeplab in both single-scale evaluation and multi-scale evaluation. EfficientPS pre-trained on Mapillary Vistas and with single-scale evaluation outperforms Panoptic-Deeplab in the same configuration by $0.8\%$ in PQ and $3.1\%$ in AP, while for multi-scale evaluation it exceeds the performance of Panoptic-Deeplab by $0.5\%$ in PQ and $1.3\%$ in AP.

We report the benchmarking results on the Cityscapes test set in \tabref{tab:baselineCityscapesTest}, where the results were obtained directly from the leaderboard. Note that the official Cityscapes benchmark only reports the PQ,  PQ\textsuperscript{St}, PQ\textsuperscript{Th}, SQ and RQ metrics, and ranks the methods primarily based on the standard PQ metric. Our proposed EfficientPS without pre-training on any extra data achieves a PQ of $64.1\%$ which is an improvement of $1.8\%$ over the previous state-of-the-art Panoptic-Deeplab trained only using Cityscapes \textit{fine} annotations and an improvement of $1.5\%$ in PQ over the Seamless model that also uses extra data. More importantly, our proposed EfficientPS model pre-trained on Mapillary Vistas, sets the new state-of-art on the Cityscapes panoptic benchmark achieving a PQ score of $66.4\%$. This accounts for an improvement of $0.9\%$ in PQ compared to the previous state-of-the-art Panoptic Deeplab pre-trained on Mapillary Vistas. Moreover, our EfficientPS model ranks second in the semantic segmentation task with a mIoU of $84.2\%$ as well as second in the instance segmentation task with an AP of $39.1\%$, among all the published methods in the Cityscapes benchmark.

We compare the efficiency of our proposed EfficientPS architecture against state-of-the-art models in terms of the number of parameters and FLOPs that it consumes as well as the runtime on the Cityscapes dataset. Operations that involve addition and multiplication at their core are only considered while computing FLOPs. We compute the end-to-end runtime of inference for our architecture as well as for the state-of-the-art methods whose runtime is not reported in their respective paper. We use a single Nvidia Titan RTX GPU and an Intel Xenon@2.20GHz CPU. We average over 1000 runs on the same image with single scale test. In the case of parallel components in the architecture, maximum runtime among all the components contribute to the total runtime. \tabref{tab:runtimesCityscapes} shows the comparison with the top two top-down and bottom-up panoptic segmentation architectures. Our proposed EfficientPS has a runtime of $166\si{\milli\second}$ for an input image resolution of $1024\times2048$ pixels which makes it faster than the competing methods. We also observe that our EfficientPS architecture consumes the least amount of parameters and FLOPs, thereby making it the most efficient state-of-the-art panoptic segmentation model.

\begin{table*}
\footnotesize
\centering
\caption{Performance comparison of panoptic segmentation on the KITTI validation set. Note that no additional data was used for training EfficientPS on this dataset other than pre-training the encoder on ImageNet. Superscripts St and Th refer to ‘stuff’ and ‘thing’ classes respectively.}
\label{tab:baselineKITTI}
\begin{tabular}{p{0.7cm}p{2.7cm}|p{0.5cm}p{0.5cm}p{0.5cm}p{0.5cm}p{0.5cm}p{0.5cm}p{0.5cm}p{0.5cm}p{0.5cm}p{0.5cm}p{0.6cm}}
\toprule
Mode & Network & PQ & SQ & RQ & PQ\textsuperscript{Th} &SQ\textsuperscript{Th}& RQ\textsuperscript{Th} &PQ\textsuperscript{St}  &SQ\textsuperscript{St} &   RQ\textsuperscript{St} & AP & mIoU \\
 &  & $(\%)$ & $(\%)$ & $(\%)$ & $(\%)$ & $(\%)$ & $(\%)$ & $(\%)$ & $(\%)$ & $(\%)$ & $(\%)$ & $(\%)$ \\
\noalign{\smallskip}\hline\hline\noalign{\smallskip}
\multirow{4}{*}{\rotatebox[origin=0]{90}{Single-Scale}} & Panoptic FPN &  $38.6$ &$70.4$ & $51.2$ &$26.1$ & $68.3$ &$40.1$ &$47.6$ &  $71.9$ &$59.2$ & $24.4$ & $52.1$ \\
& UPSNet & $39.1$ & $70.7$ &$51.7$ &$26.6$ &$68.5$ & $40.6$ &$48.3$ &   $72.4$ &$59.8$ &$24.7$ & $52.6$ \\  
& Seamless & $41.3$ & $71.7$ &$52.3$ & $28.5$ &$69.2$ & $42.3$ &$50.6$ & $73.6$ &$59.6$ & $25.9$ & $53.8$ \\
\cmidrule{2-13}
&\textbf{EfficientPS (ours)} & $\mathbf{42.9}$ & $\mathbf{72.7}$ &$\mathbf{53.6}$ &$\mathbf{30.4}$ & $\mathbf{69.8}$ &$\mathbf{43.7}$ &$\mathbf{52.0}$ &  $\mathbf{74.9}$ &$\mathbf{60.9}$ & $\mathbf{27.1}$ & $\mathbf{55.3}$ \\
\noalign{\smallskip}\hline\hline\noalign{\smallskip}
\multirow{4}{*}{\rotatebox[origin=0]{90}{Multi-Scale}}& Panoptic FPN & $39.3$ &  $70.8$ &$51.6$ &$26.9$ &$68.7$ &$40.4$ &$48.3$ &  $72.4$ &$59.8$ & $24.8$ & $52.8$ \\
& UPSNet &  $39.9$ & $71.2$ &$52.0$ &$27.2$ &$68.8$ &$40.8$ &$49.1$ & $72.9$ &  $60.2$ & $25.2$ & $53.2$ \\
& Seamless & $42.2$ &  $72.3$ &$52.9$ &$29.1$ &$69.7$ & $42.9$ & $51.8$ & $74.2$ & $60.1$ & $26.6$ & $55.1$ \\
\cmidrule{2-13}
&\textbf{EfficientPS (ours)} & $\mathbf{43.7}$ &$\mathbf{73.2}$ & $\mathbf{54.1}$ &$\mathbf{30.9}$ & $\mathbf{70.2}$ & $\mathbf{44.0}$ &$\mathbf{53.1}$ & $\mathbf{75.4}$ &$\mathbf{61.5}$ & $\mathbf{27.9}$ & $\mathbf{56.4}$ \\
\bottomrule
\end{tabular}
\end{table*}

\begin{table*}
\footnotesize
\centering
\caption{Performance comparison of panoptic segmentation on the Indian Driving Dataset (IDD) validation set. Note that no additional data was used for training EfficientPS on this dataset other than pre-training the encoder on ImageNet. Superscripts St and Th refer to ‘stuff’ and ‘thing’ classes respectively.}
\label{tab:baselineIDD}
\begin{tabular}{p{0.7cm}p{2.7cm}|p{0.5cm}p{0.5cm}p{0.5cm}p{0.5cm}p{0.5cm}p{0.5cm}p{0.5cm}p{0.5cm}p{0.5cm}p{0.5cm}p{0.6cm}}
\toprule
Mode & Network & PQ & SQ & RQ &PQ\textsuperscript{Th} &SQ\textsuperscript{Th} & RQ\textsuperscript{Th} &  PQ\textsuperscript{St} & SQ\textsuperscript{St} &  RQ\textsuperscript{St} & AP & mIoU \\
 &  & $(\%)$ & $(\%)$ & $(\%)$ & $(\%)$ & $(\%)$ & $(\%)$ & $(\%)$ & $(\%)$ & $(\%)$ & $(\%)$ & $(\%)$ \\
\noalign{\smallskip}\hline\hline\noalign{\smallskip}
\multirow{4}{*}{\rotatebox[origin=0]{90}{Single-Scale}} & Panoptic FPN & $45.9$ & $75.9$ &$60.8$ & $46.1$ & $77.8$ &$60.9$ &$45.8$ &  $74.9$ &  $60.7$ & $27.8$ & $68.1$ \\
& UPSNet & $46.6$ &  $76.5$ &$60.9$ &$47.6$ & $78.9$ &$61.1$ & $46.0$ & $75.3$ &$60.8$ &  $28.2$ & $68.4$  \\ 
& Seamless  & $47.7$ & $77.2$ &$61.2$ &$48.9$ &$79.5$ & $61.5$ &$47.1$ &  $76.1$ & $61.1$ & $30.1$ & $69.6$  \\
\cmidrule{2-13}
&\textbf{EfficientPS (ours)}& $\mathbf{50.1}$ & $\mathbf{78.4}$ &$\mathbf{62.0}$ & $\mathbf{50.7}$ & $\mathbf{80.6}$ &$\mathbf{61.6}$ &$\mathbf{49.8}$ & $\mathbf{77.1}$ &   $\mathbf{62.2}$ &  $\mathbf{31.6}$ & $\mathbf{71.3}$ \\
\noalign{\smallskip}\hline\hline\noalign{\smallskip}
\multirow{4}{*}{\rotatebox[origin=0]{90}{Multi-Scale}}
& Panoptic FPN & $46.7$ & $77.0$ &$61.0$ &$47.3$ & $78.9$ &$61.1$ &$46.4$ &  $76.1$ & $61.0$ & $28.9$ & $70.1$ \\
& UPSNet & $47.1$ & $77.9$ &$60.9$ &$47.6$ & $79.8$ &$61.2$ &$46.8$ &  $76.9$ & $60.8$ &  $29.2$ & $70.6$  \\ 
& Seamless  & $48.5$ &  $78.2$ &$61.9$ &$49.5$ &$80.4$ &$62.2$ &$47.9$ & $77.1$ &  $61.7$ &  $31.4$ & $71.3$  \\
\cmidrule{2-13}
&\textbf{EfficientPS (ours)} & $\mathbf{51.1}$ &  $\mathbf{78.8}$ & $\mathbf{63.5}$ &$\mathbf{52.6}$ & $\mathbf{81.2}$ &$\mathbf{65.4}$ &$\mathbf{50.3}$ &  $\mathbf{77.5}$ & $\mathbf{62.5}$ &  $\mathbf{32.9}$ & $\mathbf{72.1}$ \\
\bottomrule
\end{tabular}
\end{table*}

In \tabref{tab:baselineMapillary}, we report results on the Mapillary Vistas validation set. The Mapillary Vistas dataset presents a substantial challenge as it contains images from varying seasons, weather conditions and time of day as well as the presence of 65 semantic object classes. Our proposed EfficientPS model exceeds the state-of-the-art for both single-scale and multi-scale evaluation. For single-scale evaluation, it achieves an improvement of $0.6\%$ in PQ over the top-down approach Seamless and the bottom-up approach Panoptic-DeepLab. While for multi-scale evaluation, it achieves an improvement of $0.4\%$ in PQ and $3.6\%$ in AP over the previous state-of-the-art Panoptic-DeepLab. Note that we do not use model ensembles. Our network falls short of the bottom-up approach Panoptic-Deeplab in PQ\textsuperscript{St} score primarily due to the output stride of 16 at which it operates which increases the computational complexity, whereas our EfficientPS uses an output stride of 32, hence is more efficient. On the one hand, bottom-up approaches tend to have a better semantic segmentation ability which is evident from the high PQ\textsuperscript{St} of Panoptic-Deeplab. While on the other hand, top-down approaches tend to have better instance segmentation ability as they can handle large-scale variations in object instances. It would be interesting to investigate architectures that can combine the strengths of the two in future.

We present results on the KITTI validation set in \tabref{tab:baselineKITTI}. Our proposed EfficientPS outperforms the previous state-of-the-art Seamless by $1.6\%$ in PQ, $1.2\%$ in AP and $1.5\%$ mIoU for single scale evaluation and $1.5\%$ in PQ, $1.3\%$ in AP and $1.3\%$ in mIoU for multi-scale evaluation. This dataset consists of cluttered and occluded objects that often have object masks split into two or more parts. In these cases context aggregation plays a major role. Hence, the improvement that we observe can be attributed to three factors: the multi-scale feature aggregation in our 2-way FPN due to the bidirectional flow of information, the long-range context being captured by our semantic head, and the adaptive fusion in our panoptic fusion module that effectively leverages the predictions from the individual heads.

\begin{table*}
\footnotesize 
\centering
\caption{Ablation study on various architectural contributions proposed in our EfficientPS model. The performance is shown for the models trained on Cityscapes \textit{fine} annotations and evaluated on the validation set. SIH, SH, and PFM denotes depthwise separable Instance Head, Semantic Head, and Panoptic Fusion Module respectively. '-' refers to the standard configuration as \cite{kirillov2019bpanoptic}, whereas '\checkmark' refers to our proposed configuration. Superscripts St and Th refer to ‘stuff’ and ‘thing’ classes respectively.}
\label{tab:epsnetArchitecureEvaluation}
\begin{tabular}{p{0.5cm}p{2.8cm}p{0.8cm}p{0.5cm}p{0.5cm}p{0.5cm}|p{0.4cm}p{0.4cm}p{0.4cm}p{0.4cm}p{0.4cm}p{0.4cm}p{0.4cm}p{0.4cm}p{0.4cm}p{0.4cm}p{0.4cm}}
\noalign{\smallskip}\hline\noalign{\smallskip}
Model & Encoder & 2-way & SIH & SH & PFM & PQ  &SQ &RQ &PQ\textsuperscript{Th} & SQ\textsuperscript{Th}&  RQ\textsuperscript{Th}& PQ\textsuperscript{St} &  SQ\textsuperscript{St} &  RQ\textsuperscript{St} & AP & mIoU \\
 & & FPN & & & & $(\%)$ & $(\%)$ & $(\%)$ & $(\%)$ & $(\%)$ & $(\%)$ & $(\%)$ & $(\%)$ & $(\%)$ & $(\%)$ & $(\%)$ \\
\noalign{\smallskip}\hline\hline\noalign{\smallskip}
M1 & ResNet-50 & - & - & - & -  & $57.8$ & $78.8$ &$71.7$ &$52.1$ &$78.5$ & $66.9$ &$61.8$ &  $79.0$ &$75.2$ & $31.1$ & $74.1$   \\
M2 & ResNet-50 & - & - & - & -  & $58.1$ & $79.0$ &$71.9$ &$52.3$ &$78.7$ & $67.0$ &$62.3$ &  $79.2$ &$75.4$ & $31.4$ & $74.3$   \\
M3 & ResNet-50 & - & - & - & -  & $58.2$ & $79.1$ &$72.0$ &$52.4$ &$78.8$ & $67.2$ &$62.4$ &  $79.4$ &$75.6$ & $31.6$ & $74.6$   \\
M4 & ResNet-50 & - & - & - & \checkmark  & $58.8$ &$79.5$ & $72.6$ &$53.4$ &$79.2$ &$62.8$ &$67.9$ &  $79.7$ &$76.1$ & $33.8$ & $75.1$ \\
M5 & ResNet-50 & - & \checkmark & - & \checkmark  & $58.6$ & $79.4$ &$72.4$ &  $53.1$ &$79.1$ & $67.5$ & $62.6$ &$79.6$ &$75.9$ & $33.7$ & $75.0$  \\
M6 & Mod. EfficientNet-B5 & - & \checkmark & - & \checkmark & $59.7$ &$79.9$ & $73.3$ & $54.7$ &$76.6$ &$68.1$ &$63.3$ &  $80.3$ &$79.5$ & $34.1$ & $76.3$  \\
M7 & Mod. EfficientNet-B5 & \checkmark & \checkmark & - & \checkmark  & $61.5$ &$80.7$ &$75.6$ & $57.2$
 &$80.6$ & $72.5$ & $64.6$ &   $80.9$ &$77.9$ &$36.8$ & $77.3$  \\

M8 & Mod. EfficientNet-B5 & \checkmark & \checkmark & \checkmark & \checkmark & $\mathbf{63.9}$ & $\mathbf{81.5}$  & $\mathbf{77.1}$ & $\mathbf{60.7}$ & $\mathbf{81.2}$ &$\mathbf{74.1}$&$\mathbf{66.2}$ &$\mathbf{81.8}$ & $\mathbf{79.2}$ &$\mathbf{38.3}$ & $\mathbf{79.3}$ \\
\noalign{\smallskip}\hline\noalign{\smallskip}
\end{tabular}
\end{table*}

Finally, we also report results on the Indian Driving Dataset (IDD) largely due to the fact that it contains images of unstructured urban environments and scenes that do not have clear delineated road infrastructure which makes it extremely challenging. \tabref{tab:baselineIDD} presents results on the IDD validation set. Our proposed EfficientPS substantially exceeds the state-of-the-art by achieving a PQ score of $50.1\%$ and $51.1\%$ for single-scale and multi-scale evaluation respectively. This amounts to an improvement of $2.6\%$ in PQ over Seamless and $4\%$ in PQ over UPSNet for multi-scale evaluation. The unstructured scenes in this dataset challenges the ability of models to detect object boundaries of ‘stuff’ classes such as road and sidewalk. Our EfficientPS achieves a PQ\textsuperscript{St} score of $49.8\%$ for single-scale evaluation which is an improvement of $2.7\%$ over Seamless and this can be attributed to the effectiveness of our proposed semantic head in capturing object boundaries.

\subsection{Ablation Studies}
\label{sec:ablation}

In this section, we present extensive ablation studies on the various architectural components that we propose in our EfficientPS architecture in comparison to their counterparts employed in state-of-the-art models. Primarily, we study the impact of our proposed network backbone, semantic head and panoptic fusion module on the overall panoptic segmentation performance of our network. We begin with a detailed analysis of various components of our EfficientPS architecture, followed by comparisons of different encoder network topologies and FPN architectures for the network backbone. We then study the impact of different parameter configurations in our proposed semantic head and its comparison with existing semantic head topologies. Finally, we assess the performance of our proposed panoptic fusion module by comparing with different panoptic fusion methods proposed in the literature. For all the ablative experiments, we train our models on the Cityscapes \textit{fine} annotations and evaluate it on the validation set. We use the PQ metric as the primary evaluation criteria for all the experiments presented in this section. Nevertheless, we also report the other metrics defined in the beginning of \secref{sec:experimentalResults}.
 
\subsubsection{Detailed Study on the EfficientPS Architecture}
\label{sec:detailedAblation}

We first study the improvement due to the various components that we propose in our EfficientPS architecture. Results from this experiment are shown in \tabref{tab:epsnetArchitecureEvaluation}. The basic model M1 employs the network configuration and panoptic fusion heuristics as~\cite{kirillov2019panoptic}. It uses the ResNet-50 with FPN as the backbone and incorporates Mask R-CNN for the instance head. It employs group norm ~\citep{wu2018group} for the normalization layer. The semantic head of this network is comprised of an upsampling stage which has a $3\times3$ convolution, group norm~\citep{wu2018group}, ReLU, and $\times2$ bilinear upsampling. At each FPN level, this upsampling stage is repeated until the feature maps are $1/4$ scale of the input. These resulting feature maps are then summed element-wise and passed through a $1\times1$ convolution, followed by $\times4$ bilinear upsampling, and softmax to yield the semantic segmentation output. This model M1 achieves a PQ of $57.8\%$, AP of $31.1\%$ and an mIoU score of~$74.1\%$. For the M2 and M3 model, we use BN sync and IABN sync as the normalization layer. Additionally in M3 ReLU is replaced with leakyReLU activation layer. We observe that M3 and M2 obtains a gain of $0.4\%$ and $0.3\%$ over M1 respectively, implying that with a higher batch size of 16 it is better to employ BN sync or iABN sync than group norm as the normalization layer. As M3 has a slight improvement over M2 we build subsequent models based on M3.

The next model M4 that incorporates our proposed panoptic fusion module achieves an improvement of $0.6\%$ in PQ, $2.2\%$ in AP and $0.8\%$ in the mIoU score without increasing the number of parameters. This increase in performance demonstrates that the adaptive fusion of semantic and instance head outputs is effective in resolving the inherent overlap conflict. In the M5 model, we replace all the standard convolutions in the instance head with depthwise separable convolutions which reduces the number of parameters of the model by $\SI{2.09}{\million}$ with a drop of $0.2\%$ in PQ, $0.1\%$ drop in AP and mIoU score. However, from the aspect of having an efficient model, a reduction of $5\%$ of the model parameters for a drop of $0.2\%$ in PQ can be considered as a reasonable trade-off. Therefore, we employ depthwise separable convolutions in the instance head of our proposed EfficientPS architecture.

\begin{table*}
\footnotesize 
\centering
\caption{Performance comparison of various encoder topologies employed in the M8 model. Results are shown for the models trained on the Cityscapes \textit{fine} annotations and evaluated on the validation set. Superscripts St and Th refer to ‘stuff’ and ‘thing’ classes respectively.}
\label{tab:encoderEvaluation}
\begin{tabular}{p{3.7cm}p{1.0cm}p{1.0cm}|p{0.5cm}p{0.5cm}p{0.5cm}p{0.5cm}p{0.5cm}p{0.5cm}p{0.5cm}p{0.5cm}p{0.5cm}p{0.5cm}p{0.6cm}}
\noalign{\smallskip}\hline\noalign{\smallskip}
Encoder & Params & FLOPs & PQ &SQ &RQ & PQ\textsuperscript{Th} &SQ\textsuperscript{Th}&RQ\textsuperscript{Th} & PQ\textsuperscript{St} &  SQ\textsuperscript{St} &   RQ\textsuperscript{St} & AP & mIoU \\
 & $(\si{\million})$ & $(\si{\billion})$ & $(\%)$ & $(\%)$ & $(\%)$ & $(\%)$ & $(\%)$ & $(\%)$ & $(\%)$ & $(\%)$ & $(\%)$ & $(\%)$ & $(\%)$ \\
\noalign{\smallskip}\hline\hline\noalign{\smallskip}
 MobileNetV3 & $5.40$ & $9.44$ &  $55.8$ & $78.1$ & $70.2$ & $50.4$ & $77.4$ & $67.1$ & $59.8$ & $78.6$ & $72.4$ & $29.1$ & $72.2$ \\
 ResNet-50 & $25.60$ & $172.19$ &  $60.3$ & $80.1$ & $72.6$ & $55.3$ & $79.9$ & $68.9$ & $63.9$ & $80.3$ & $75.3$ & $34.9$ & $76.1$ \\
 ResNet-101 & $44.50$ & $327.99$ & $61.1$ & $80.3$ & $75.1$ & $56.5$ & $80.1$ & $71.9$ & $64.2$ & $80.5$ & $77.4$ & $35.9$ & $77.2$ \\
 Xception-71 & $27.50$ & $210.38$ &  $62.1$ & $81.1$ & $75.4$ & $58.5$ & $80.9$ & $72.3$ & $64.7$ & $81.2$ & $77.7$ & $36.2$ & $78.1$ \\
 ResNeXt-101 &$86.74$ & $636.84$ &  $63.2$ & $81.2$ & $76.0$ & $59.6$ & $80.4$ & $72.9$ & $65.8$ & $81.7$ & $78.3$ & $36.9$ & $78.9$ \\
 \textbf{Mod. EfficientNet-B5 (Ours)} & $30.00$ & $250.97$ & $\mathbf{63.9}$ & $\mathbf{81.5}$  & $\mathbf{77.1}$ & $\mathbf{60.7}$ & $\mathbf{81.2}$ &$\mathbf{74.1}$&$\mathbf{66.2}$ &$\mathbf{81.8}$ & $\mathbf{79.2}$ &$\mathbf{38.3}$ & $\mathbf{79.3}$ \\
\noalign{\smallskip}\hline\noalign{\smallskip}
\end{tabular}
\end{table*}

\begin{table*}
\footnotesize
\centering
\caption{Performance comparison of various FPN architectures employed in the M8 model. Results are shown for the models trained on the Cityscapes \textit{fine} annotations and evaluated on the validation set. Superscripts St and Th refer to ‘stuff’ and ‘thing’ classes respectively.}
\label{tab:FPNEvaluation}
\begin{tabular}{p{2.5cm}|p{0.5cm}p{0.5cm}p{0.5cm}p{0.5cm}p{0.5cm}p{0.5cm}p{0.5cm}p{0.5cm}p{0.5cm}p{0.5cm}p{0.6cm}}
\noalign{\smallskip}\hline\noalign{\smallskip}
Architecture & PQ &SQ &RQ & PQ\textsuperscript{Th} &SQ\textsuperscript{Th}&RQ\textsuperscript{Th} & PQ\textsuperscript{St} &  SQ\textsuperscript{St} &   RQ\textsuperscript{St} & AP & mIoU \\
 &  $(\%)$ &  $(\%)$ & $(\%)$ & $(\%)$ & $(\%)$ & $(\%)$ & $(\%)$ & $(\%)$ & $(\%)$ & $(\%)$ & $(\%)$ \\
\noalign{\smallskip}\hline\hline\noalign{\smallskip}
Bottom-Up FPN & $60.4$ &$80.6$ &$73.7$ &  $56.3$ &$80.4$ &$69.9$ &$63.4$ &  $80.8$ &$76.4$ & $35.2$ & $75.3$  \\
Top-Down FPN &  $62.2$ & $80.9$ &$75.7$ & $58.1$ &$80.1$ &  $72.4$ &$65.1$ & $81.4$ &$78.0$ &$36.5$ & $78.2$  \\
PANet FPN & $63.1$ & $81.1$ &$75.5$ &$59.4$ &$80.3$ & $72.3$ &$65.8$ &   $81.6$ &$77.8$ &$37.1$ & $78.8$  \\
\textbf{2-way FPN (Ours)} &  $\mathbf{63.9}$ & $\mathbf{81.5}$  & $\mathbf{77.1}$ & $\mathbf{60.7}$ & $\mathbf{81.2}$ &$\mathbf{74.1}$&$\mathbf{66.2}$ &$\mathbf{81.8}$ & $\mathbf{79.2}$ &$\mathbf{38.3}$ & $\mathbf{79.3}$ \\
\noalign{\smallskip}\hline\noalign{\smallskip}
\end{tabular}
\end{table*}

In the M6 model, we replace the ResNet-50 encoder with our modified EfficientNet-B5 encoder that does not have any squeeze-and-excitation connections, and we replace all the normalization layers and ReLU activations with iABN sync and leaky ReLU. This model achieves a PQ of $59.7\%$ which is an improvement of $1.1\%$ in PQ over the M3 model and a larger improvement is also observed in the mIoU score. The improvement in performance can be attributed to the richer representational capacity of the EfficientNet-B5 architecture. Subsequently in the M7 model, we replace the standard FPN with our proposed 2-way FPN which additionally improves the performance by $1.8\%$ in PQ and $2.7\%$ in AP. The addition of the parallel bottom-up branch in our 2-way FPN enables bidirectional flow of information, thus breaking away from the limitation of the standard FPN. 

Finally, we incorporate our proposed semantic head into the M8 model that fuses and aligns multi-scale features effectively which enables it to achieve a PQ of $63.9\%$. Although our semantic head contributes to this improvement of $2.4\%$ in the PQ score, it cannot not be solely attributed to the semantic head. This is due to the fact that if we employ standard panoptic fusion heuristics, an improvement in semantic segmentation would only contribute to an increase in PQ\textsuperscript{st} score. However, our proposed adaptive panoptic fusion yields an improvement in PQ\textsuperscript{th} as well, which is evident from the overall improvement in the PQ score. We denote this M8 model configuration as EfficientPS in this work. In the following sections, we further analyze the individual architectural components of the M6 model in more detail. 

\subsubsection{Comparison of Encoder Topologies}
\label{sec:encoderAblation}

There are numerous network architectures that have been proposed for addressing the task of image classification. Typically, these networks serve as the encoder or feature extractor for more complex tasks such as panoptic segmentation. In this section, we evaluate the performance of our proposed modified EfficientNet-B5 in comparison to five widely employed encoder architectures. For a fair comparison, we keep all the other components of our EfficientPS network the same and only replace encoder. More specifically, we compare with MobileNetV3~\citep{howard2019searching}, ResNet-50~\citep{he2016deep}, ResNet-101~\citep{he2016deep}, Xception-71~\citep{chollet2017xception}, ResNeXt-101~\citep{xie2017aggregated}, and EfficientNet-B5~\citep{tan2019efficientnet}. Results from this experiment are presented in \tabref{tab:encoderEvaluation}. We observe that our modified EfficientNet-B5 architecture yields the highest PQ score, closely followed by the ResNeXt-101 architecture. However, ResNext-101 has an additional $\SI{56.74}{\million}$ parameters which is more than twice the number of parameters consumed by our modified EfficientNet-B5 architecture. Similarly, ResNeXt-101 in FLOPs is $\SI{385.87}{\billion}$ more. We can see that the other encoder models, especially MobileNetV3, ResNet-50 and Xception-71 have a comparable or fewer parameters and FLOPs  than our modified EfficientNet-B5. However they also yield a substantially lower PQ score. Therefore, we employ our modified EfficientNet-B5 as the encoder backbone in our proposed EfficientPS architecture. The computation of FLOPs presented in \tabref{tab:encoderEvaluation} architectures is only for the encoder part of the network.

\subsubsection{Evaluation of the 2-way FPN}
\label{sec:fpnAblation}

In this section, we compare the performance of our novel 2-way FPN with other existing FPN variants. For a fair comparison, we keep all the other components of our EfficientPS network the same and only replace the 2-way FPN in the backbone. We compare with the top-down FPN~\citep{lin2017feature}, bottom-up FPN and PANet FPN variants. We refer to the FPN architecture described in~\cite{liu2018path} as PANet FPN in which the top-down path is followed by a bottom-up path. For each of the FPN variants we use iABN sync and leaky ReLU layers instead of BN and Relu layers. The results from comparing with various FPN architectures are shown in \tabref{tab:FPNEvaluation}.

\begin{table*}
\footnotesize
\centering
\renewcommand{\arraystretch}{1.4}
\caption{Ablation study on our semantic head topology incorporated into the M6 model. Results are shown for the models trained on the Cityscapes \textit{fine} annotations and evaluated on the validation set. \textbf{$P_{os}$} is the output of our 2-way FPN at the $os$ pyramid scale level, \textbf{$c^k_f$} refers to a convolution layer with $f$ number of filters and $k\times k$ kernel size, LSFE refers to Large Scale Feature Extractor and DPC refers to Dense Prediction Cells. Superscripts St and Th refer to ‘stuff’ and ‘thing’ classes respectively.}
\label{tab:semArchEval}
\begin{tabular}{p{0.5cm}p{1.0cm}p{1.0cm}p{1.0cm}p{1.0cm}p{1.2cm}|p{0.4cm}p{0.4cm}p{0.4cm}p{0.4cm}p{0.4cm}p{0.4cm}p{0.4cm}p{0.4cm}p{0.4cm}p{0.4cm}p{0.5cm}}
\noalign{\smallskip}\hline\noalign{\smallskip}
Model & P\textsubscript{32} &  P\textsubscript{16} &  P\textsubscript{8} &  P\textsubscript{4} & Feature  &PQ &SQ &RQ & PQ\textsuperscript{Th} &SQ\textsuperscript{Th}&RQ\textsuperscript{Th} & PQ\textsuperscript{St} &  SQ\textsuperscript{St} &   RQ\textsuperscript{St} & AP & mIoU \\
 & & & & & Correlation & $(\%)$ & $(\%)$ & $(\%)$ & $(\%)$ & $(\%)$ & $(\%)$ & $(\%)$ & $(\%)$ & $(\%)$ & $(\%)$ & $(\%)$ \\
\noalign{\smallskip}\hline\hline\noalign{\smallskip}
M81 & $[c^3_{128}c^3_{128}]$ & $[c^3_{128}c^3_{128}]$ & $[c^3_{128}c^3_{128}]$ & $[c^3_{128}c^3_{128}]$ & -  & $61.6$ &$80.6$ &$75.7$ & $57.3$ &$80.4$ & $72.6$ &$64.7$ &   $80.7$ &$77.9$ &$36.7$ & $77.2$ \\
M82 & LSFE & LSFE & LSFE & LSFE & -& $61.7$ &  $80.5$ &$75.4$ &$57.9$
 &$80.5$ &$72.6$ &$64.5$ &  $80.6$ &$77.4$ & $36.6$ & $77.4$ \\
M83 & DPC & LSFE & LSFE & LSFE & -& $62.3$ & $80.9$ &$75.9$ &$57.9$ &$80.4$ & $72.0$ & $65.6$ &  $81.3$ &$78.8$ &$36.8$ & $78.1$ \\
M84 & DPC & DPC & LSFE & LSFE & -& $62.9$ &  $81.0$ &$75.7$ &$59.0$ &$80.5$ &$71.4$ &$65.8$ &  $81.4$ &$78.7$ & $37.0$ & $78.6$ \\
M85 & DPC & DPC & DPC & LSFE & -   & $62.4$ &$80.8$ & $75.4$ &$58.8$ &$80.3$ &$72.4$ &$65.1$ &   $81.1$ &$77.6$ & $36.7$ & $78.2$  \\
M86 & DPC & DPC  & LSFE & LSFE & \checkmark  &$\mathbf{63.9}$ & $\mathbf{81.5}$  & $\mathbf{77.1}$ & $\mathbf{60.7}$ & $\mathbf{81.2}$ &$\mathbf{74.1}$&$\mathbf{66.2}$ &$\mathbf{81.8}$ & $\mathbf{79.2}$ &$\mathbf{38.3}$ & $\mathbf{79.3}$ \\
\noalign{\smallskip}\hline\noalign{\smallskip}
\end{tabular}
\end{table*}

The top-down FPN model predominantly propagates semantically high-level features which describe entire objects, whereas the bottom-up FPN model propagates low-level information such as local textures and patterns. The EfficientPS model with the bottom-up FPN achieves a PQ of $60.4\%$, while the model with the top-down FPN achieves a PQ of $62.2\%$. Both these models achieve a performance which is $3.2\%$ and $1.4\%$ lower in PQ than our 2-way FPN respectively. A similar trend can also be observed in the other metrics. The lower PQ score of the individual bottom-up FPN and top-down FPN models substantiate the limitation of the unidirectional flow of information in the standard FPN topology. Both the PANet FPN and our proposed 2-way FPN aim to mitigate this problem by adding another bottom-up path to the standard FPN in a sequential or parallel manner respectively. We observe that the model with our proposed 2-way FPN demonstrates an improvement of $0.5\%$ in PQ over the model with the PANet FPN. This implies that the parallel information pathways are more likely to capture better multi-scale features to predict stuff regions at varying resolutions as well as are able to encode sufficiently rich semantics to precisely predict class labels.
 
\subsubsection{Detailed Study on the Semantic Head}
\label{sec:detailedSemantic}

We construct the topology of our proposed semantic head considering two critical factors. First, since large-scale outputs comprise of characteristic features and small-scale outputs consist of contextual features, they both should be captured distinctly by the semantic head. Second, while fusing small and large-scale outputs, the contextual features need to be aligned to obtain semantically reinforced fine features. In order to demonstrate that these two critical factors are essential, we perform ablative experiments on various configurations of our semantic head incorporated into the M8 model described in \secref{sec:detailedSemantic}. Results from this experiment are presented in \tabref{tab:semArchEval}.

The output at each level of the 2-way FPN, P\textsubscript{32}, P\textsubscript{16}, P\textsubscript{8} and P\textsubscript{4} are the inputs to our semantic head. In the first M81 model configuration, we employ two cascaded $3\times3$ convolutions, iABN sync and leaky ReLU activation sequentially at each level of the 2-way FPN. The aforementioned series of layers constitute the LSFE module which is followed by a bilinear upsampling layer at each level of the 2-way FPN to yield an output which is $1/4$ scale of the input image. These upsampled features are then concatenated and passed through a $1\times1$ convolution and bilinear upsamplig to yield an output which is the same scale as the input image. This M61 model achieves a PQ of $61.6\%$. In the subsequent M82 model configuration, we replace all the standard $3\times3$ convolutions with $3\times3$ depthwise separable convolutions in the LSFE module to reduce the number of parameters. This also yields a minor improvement in performance compared to the M81 model, therefore we employ depthwise separable convolutions in all the experiments that follow.

In the M83 model, we replace the LSFE module in the P\textsubscript{32} level of the 2-way FPN with dense prediction cells (DPC) described in \secref{sec:semanticHead}. This M83 model achieves an improvement of $0.6\%$ in PQ and $0.7\%$ in the mIoU score. This can be attributed to the ability of DPC to effectively capture long-range context. In the M84 model, we replace the LSFE module in the P\textsubscript{16} level with DPC and in the subsequent M85 model, we introduce DPC at both P\textsubscript{16} and P\textsubscript{8} levels. We find that the M84 model achieves an improvement of $0.6\%$ in PQ over M63, however the performance drops in the M85 model by $0.5\%$ in PQ when we add the DPC module at the P\textsubscript{8} level. This can be attributed to the fact that DPC consisting of dilated convolutions do not capture characteristic features effectively at this large-scale. The final M86 model is derived from the M84 model to which we add our mismatch correction (MC) module along with the feature correlation connections as described in \secref{sec:semanticHead}. This model achieves the highest PQ score of $63.9\%$ which is an improvement of $1.0\%$ compared to the M84 model. This can be attributed to the MC module that correlates the semantically rich contextual features with fine features and subsequently merges them along the feature correlation connection to obtain semantically reinforced features that results in better object boundary refinement.

\begin{table*}
\footnotesize
\centering
\caption{Performance comparison of various existing semantic head topologies employed in the M8 model. Results are reported for the model trained on the Cityscapes \textit{fine} annotations and evaluated on the validation set. Superscripts St and Th refer to ‘stuff’ and ‘thing’ classes respectively.}
\label{tab:segComparison}
\begin{tabular}{p{2cm}|p{0.5cm}p{0.5cm}p{0.5cm}p{0.5cm}p{0.5cm}p{0.5cm}p{0.5cm}p{0.5cm}p{0.5cm}p{0.5cm}p{0.6cm}}
\noalign{\smallskip}\hline\noalign{\smallskip}
Semantic Head & PQ &SQ &RQ & PQ\textsuperscript{Th} &SQ\textsuperscript{Th}&RQ\textsuperscript{Th} & PQ\textsuperscript{St} &  SQ\textsuperscript{St} &   RQ\textsuperscript{St} & AP & mIoU \\
 & $(\%)$ & $(\%)$ & $(\%)$ & $(\%)$ & $(\%)$ & $(\%)$ & $(\%)$ & $(\%)$ & $(\%)$ & $(\%)$ & $(\%)$ \\
\noalign{\smallskip}\hline\hline\noalign{\smallskip}
Baseline & $61.5$ & $80.7$ &$75.6$ &$57.2$ &$80.6$ & $72.5$ &$64.6$ &  $80.9$ &$77.9$ & $36.8$ & $77.3$ \\
UPSNet & $62.0$ & $81.0$ &$74.7$ & $58.5$ & $80.5$&$70.9$ &$64.5$ & $81.3$ & $77.5$ & $35.9$ & $76.1$ \\
Seamless &$62.9$ & $81.1$ &$75.5$ & $58.9$ &$80.4$ & $71.3$ &$65.6$ &  $81.6$ &$78.5$ &$36.8$ & $78.5$  \\
\textbf{Ours} &  $\mathbf{63.9}$ & $\mathbf{81.5}$  & $\mathbf{77.1}$ & $\mathbf{60.7}$ & $\mathbf{81.2}$ &$\mathbf{74.1}$&$\mathbf{66.2}$ &$\mathbf{81.8}$ & $\mathbf{79.2}$ &$\mathbf{38.3}$ & $\mathbf{79.3}$ \\
\noalign{\smallskip}\hline\noalign{\smallskip}
\end{tabular}
\end{table*}

Additionally, we present experimental comparisons of our proposed semantic head against those that are used in other state-of-the-art panoptic segmentation architectures. Specifically, we compare against the semantic head proposed by~\cite{kirillov2019bpanoptic} which we denote as the baseline, UPSNet~\citep{xiong2019upsnet} and Seamless~\citep{porzi2019seamless}. For a fair comparison, we keep all the other components of the EfficientPS architecture the same across different experiments while only replacing the semantic head. \tabref{tab:segComparison} presents the results of this experiment.

The semantic head of UPSNet which is essentially a subnetwork comprising of sequential deformable convolution layers~\citep{dai2017deformable} achieves a PQ score of $62.0\%$ which is an improvement of $0.5\%$ over the baseline model. The semantic head of the Seamless model employs their MiniDL module at each level of the 2-way FPN that further improves the PQ by $0.9\%$ over semantic head of UPSNet. The semantic heads of all these models use the same module at each level of the 2-way FPN output which are of different scales. In contrast, our proposed semantic head that employs a combination of LSFE and DPC modules at different levels of the 2-way FPN achieves the highest PQ score of $63.9\%$ and consistently outperforms the other semantic head topologies in all the evaluation metrics.
 
\subsubsection{Evaluation of Panoptic Fusion Module}
\label{sec:fusionEval}

\begin{table}
\footnotesize 
\centering
\caption{Performance comparison of our proposed adaptive fusion $(\sigma(ML_A) + \sigma(ML_B)) \odot (ML_A + ML_B)$, with Multiply: $(ML_A \odot ML_B)$ and Add: $(ML_A + ML_B)$ , employed in the M8 model where $\sigma(\cdot)$ is the sigmoid function and $\odot$ is the Hadamard product. Results are reported for the model trained on the Cityscapes \textit{fine} annotations and evaluated on the validation set. Superscripts St and Th refer to ‘stuff’ and ‘thing’ classes respectively.}
\label{tab:fusionEqEvalution}
\begin{tabular}{p{1.5cm}|p{0.3cm}p{0.3cm}p{0.3cm}p{0.3cm}p{0.3cm}p{0.3cm}p{0.3cm}p{0.3cm}p{0.5cm}}
\noalign{\smallskip}\hline\noalign{\smallskip}
Model & PQ &SQ &RQ & PQ\textsuperscript{Th} &SQ\textsuperscript{Th}&RQ\textsuperscript{Th} & PQ\textsuperscript{St} &  SQ\textsuperscript{St} &   RQ\textsuperscript{St} \\
 & $(\%)$ & $(\%)$ & $(\%)$ & $(\%)$ & $(\%)$ & $(\%)$ & $(\%)$ & $(\%)$ & $(\%)$ \\
\noalign{\smallskip}\hline\hline\noalign{\smallskip}
Multiply & $62.3$ & $80.7$ &$76.0$ &$56.9$ & $79.1$ & $71.9$&$66.3$ &  $81.9$ &$79.0$  \\
Add & $63.4$ & $81.4$ &$76.9$ &$59.3$ &$80.4$ &$73.5$&$\mathbf{66.4}$ &   $\mathbf{82.0}$ &$\mathbf{79.3}$   \\
\textbf{Ours} & $\mathbf{63.9}$ & $\mathbf{81.5}$  & $\mathbf{77.1}$ & $\mathbf{60.7}$ & $\mathbf{81.2}$ &$\mathbf{74.1}$&$66.2$ &$81.8$ & $79.2$ \\
\noalign{\smallskip}\hline\noalign{\smallskip}
\end{tabular}
\end{table}

\begin{table}
\footnotesize 
\centering
\caption{Performance comparison of our proposed panoptic fusion module with various other panoptic fusion mechanisms employed in the M8 model. Results are reported for the model trained on the Cityscapes \textit{fine} annotations and evaluated on the validation set. Superscripts St and Th refer to ‘stuff’ and ‘thing’ classes respectively.}
\label{tab:fusionEvalution}
\begin{tabular}{p{1.5cm}|p{0.3cm}p{0.3cm}p{0.3cm}p{0.3cm}p{0.3cm}p{0.3cm}p{0.3cm}p{0.3cm}p{0.5cm}}
\noalign{\smallskip}\hline\noalign{\smallskip}
Model & PQ &SQ &RQ & PQ\textsuperscript{Th} &SQ\textsuperscript{Th}&RQ\textsuperscript{Th} & PQ\textsuperscript{St} &  SQ\textsuperscript{St} &   RQ\textsuperscript{St} \\
 & $(\%)$ & $(\%)$ & $(\%)$ & $(\%)$ & $(\%)$ & $(\%)$ & $(\%)$ & $(\%)$ & $(\%)$ \\
\noalign{\smallskip}\hline\hline\noalign{\smallskip}
Baseline & $62.4$ & $80.8$ &$75.4$ &$58.7$ & $80.4$ & $72.6$&$65.1$ & $81.1$ &$77.4$  \\
TASCNet & $62.5$ &$80.9$ &$75.6$ &$58.6$ &$80.5$ &$72.8$& $65.3$ & $81.2$ &$77.7$   \\
UPSNet & $63.1$ & $81.3$ &$76.1$ &$59.5$ &$80.6$ &$73.2$&$65.7$ & $\mathbf{81.8}$ &$78.2$   \\
\textbf{Ours} & $\mathbf{63.9}$ & $\mathbf{81.5}$  & $\mathbf{77.1}$ & $\mathbf{60.7}$ & $\mathbf{81.2}$ &$\mathbf{74.1}$&$\mathbf{66.2}$ &$\mathbf{81.8}$ & $\mathbf{79.2}$ \\
\noalign{\smallskip}\hline\noalign{\smallskip}
\end{tabular}
\end{table}
 
In this section, we evaluate our proposed Fusion \eqref{eq} to fuse $ML_A$ and $ML_B$ to its simple addition and multiplication counterpart. Here,  $ML_A$ and $ML_B$ are the same entity as defined in \secref{sec:fusion}. At a glance, addition and multiplication operations might seem like a logical choice for fusing the logits to attain adaptive attenuation or amplification according to the consensus. But they are in fact sub-optimal choices with respect to \eqref{eq}. \tabref{tab:fusionEqEvalution} shows the results from this experiment. We observe our proposed fusion strategy achieves the highest performance of $63.9\%$ in $PQ$. It is $0.5\%$ higher than addition and $1.6\%$ higher than multiplication. In the case of multiplication, the resulting thing logits attain high values in comparison to stuff logits when concatenated together to form intermediate panoptic logits. This leads to over-representation of thing classes, as a result, PQ\textsuperscript{Th} suffers a lot due to an increase in false positives. PQ\textsuperscript{Th} of $56.9\%$ for multiplication is the lowest out of all the strategies.

Similarly, in the case of addition, the different range values of $ML_A$ and $ML_B$ results in biased fused logits. Generally, semantic logits have higher values out of the two and hence the fused logits are biased towards $ML_B$. This again doesn't allow optimal adaptive attenuation or amplification. PQ\textsuperscript{Th} for this strategy is $59.3\%$ which is $2.4\%$ higher than multiplication. Clearly, addition is a better strategy than multiplication but is not the best. In contrast to the above strategies, our proposed strategy addresses the aforementioned shortcomings by normalizing the sum of the two logits $(ML_A + ML_B)$ based on the sum of their individual confidence $((\sigma(ML_A) + \sigma(ML_B))$  where $\sigma(\cdot)$ is the sigmoid function. This enables the proposed fusion module to be adaptive, achieving a gain of $1.4\%$ in PQ\textsuperscript{Th} while remaining relatively equal in stuff.

Next, we evaluate the performance of our proposed panoptic fusion module in comparison to other existing panoptic fusion mechanisms. First, we compare with the panoptic fusion heuristics introduced by~\cite{kirillov2019panoptic} which we consider as a baseline as it is extensively used in several panoptic segmentation networks. We then compare with Mask-Guided fusion~\citep{li2018learning} and the panoptic fusion heuristics proposed in~\citep{xiong2019upsnet} which we refer to as TASCNet and UPSNet in the results respectively. Once again for a fair comparison, we keep all the other network components the same across different experiments and only change the panoptic fusion mechanism. 

\tabref{tab:fusionEvalution} presents results from this experiment. Combining the outputs of the semantic head and instance head that have an inherent overlap is one of the critical challenges faced by panoptic segmentation networks. The baseline approach directly chooses the output of the instance head, i.e, if there is an overlap between predictions of the ‘thing’ and ‘stuff’ classes for a given pixel, the baseline heuristic classifies the pixel as a ‘thing’ class and assigns it an instance ID. This baseline approach achieves the lowest performance of $62.4\%$ in PQ demonstrating that this fusion problem is more complex than just assigning the output from one of the heads. The Mask-Guided fusion method of TASCNet seeks to address this problem by using a segmentation mask. The mask selects which pixel to consider from the instance segmentation output and which pixel to consider from the semantic segmentation output. This fusion approach achieves a PQ of $62.5\%$ which is comparable to the baseline method. Subsequently, the model that employs the UPSNet fusion heuristics achieves a larger improvement with a PQ score of $63.1\%$. This method computes the panoptic logits by adding the non-overlapping instance segmentation logits $ML_A$ to $ML_{B}$ that is obtained using the semantic logits as described in Section 3.4 while concatenating it to stuff logits from semantic segmenation logits. As shown, in previous experiment this is sub-optimal. However, our proposed adaptive fusion method that dynamically fuses the outputs from both the heads while refining the stuff segmentation using semantic head predictions achieves the highest PQ score of $63.9\%$ which is an improvement of $0.8\%$ over the UPSNet method. We also observe a consistently higher performance in all the other metrics.

\subsection{Qualitative Evaluations}
\label{sec:qualitative}

\begin{figure*}
\centering
\footnotesize
\setlength{\tabcolsep}{0.1cm}
{\renewcommand{\arraystretch}{1}
\begin{tabular}{P{0.4cm}P{3.9cm}P{3.9cm}P{3.9cm}P{3.9cm}}
& \raisebox{-0.4\height}{Input Image} & \raisebox{-0.4\height}{Seamless Output} & \raisebox{-0.4\height}{EfficientPS Output} & \raisebox{-0.4\height}{Improvement\textbackslash{Error Map}} \\
\\
\rot{(a) Cityscapes} & \raisebox{-0.4\height}{\includegraphics[width=\linewidth]{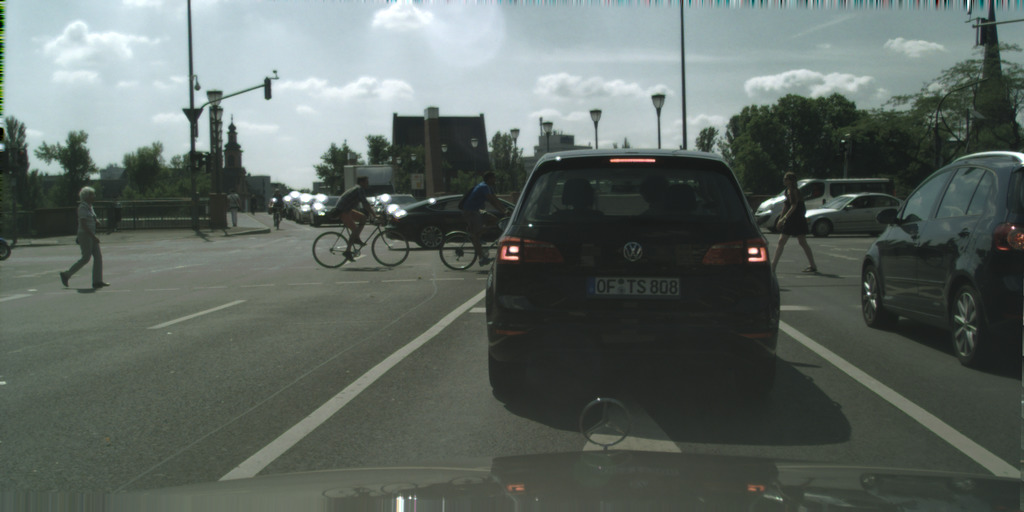}} & \raisebox{-0.4\height}{\includegraphics[width=\linewidth]{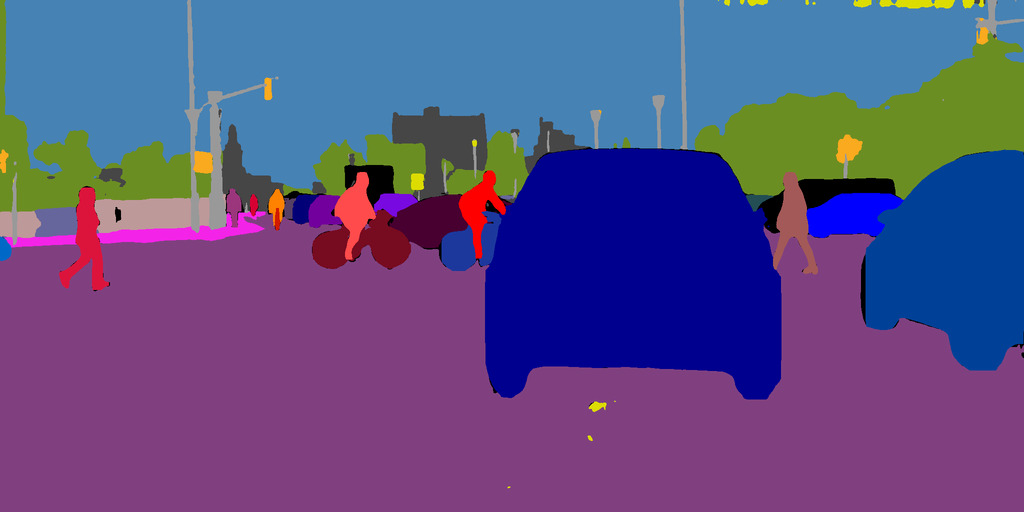}} & \raisebox{-0.4\height}{\includegraphics[width=\linewidth]{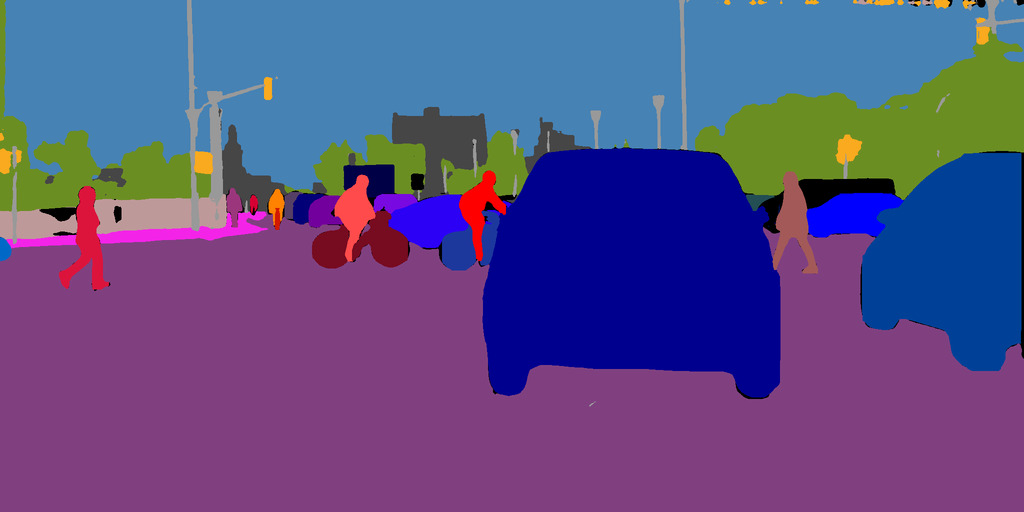}} & \raisebox{-0.4\height}{\fbox{\includegraphics[width=\linewidth]{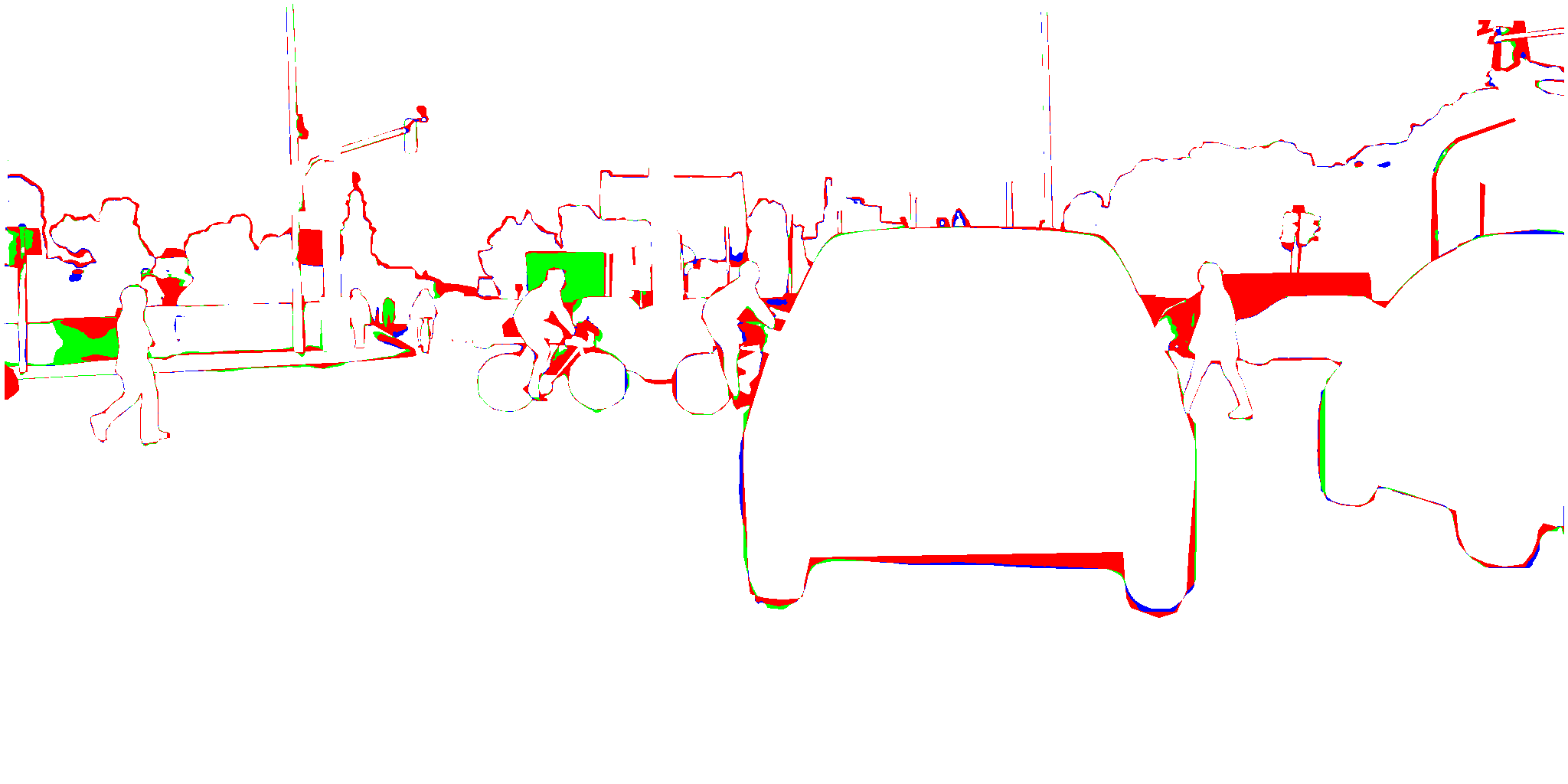}}} \\
\\
\rot{(b) Cityscapes} & \raisebox{-0.4\height}{\includegraphics[width=\linewidth]{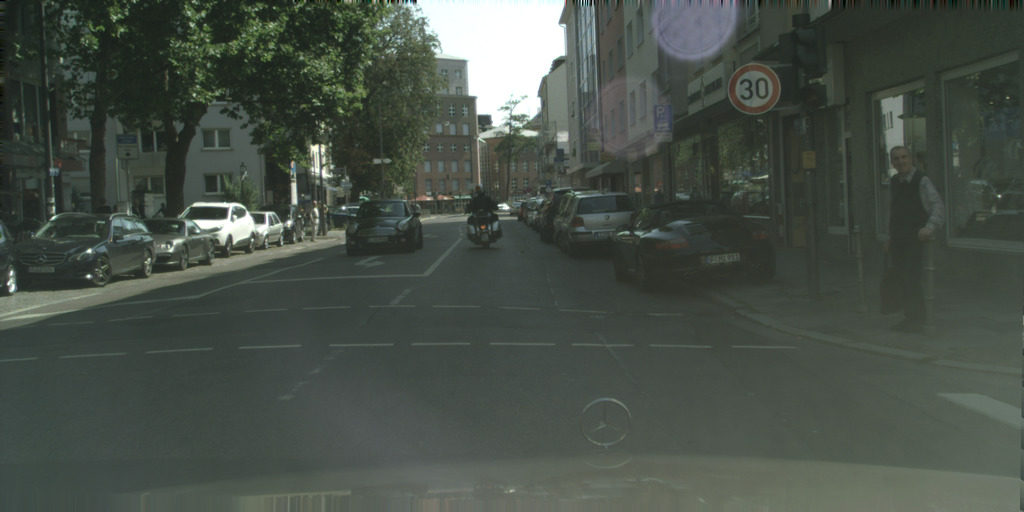}} & \raisebox{-0.4\height}{\includegraphics[width=\linewidth]{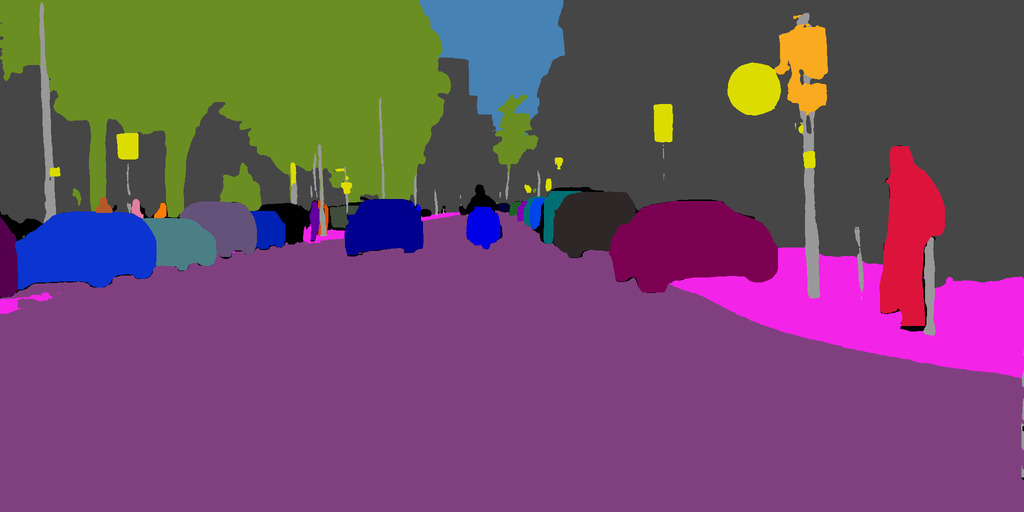}} & \raisebox{-0.4\height}{\includegraphics[width=\linewidth]{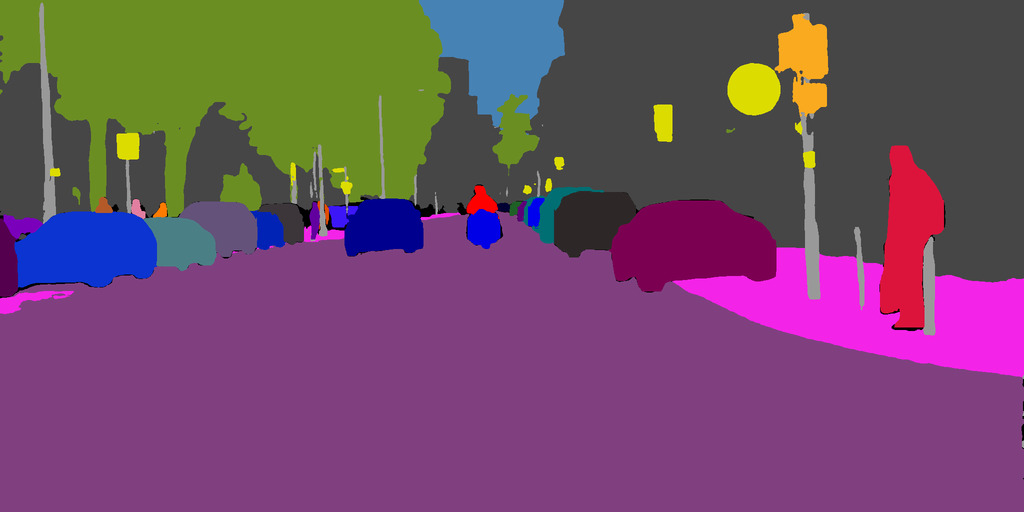}} & \raisebox{-0.4\height}{\fbox{\includegraphics[width=\linewidth]{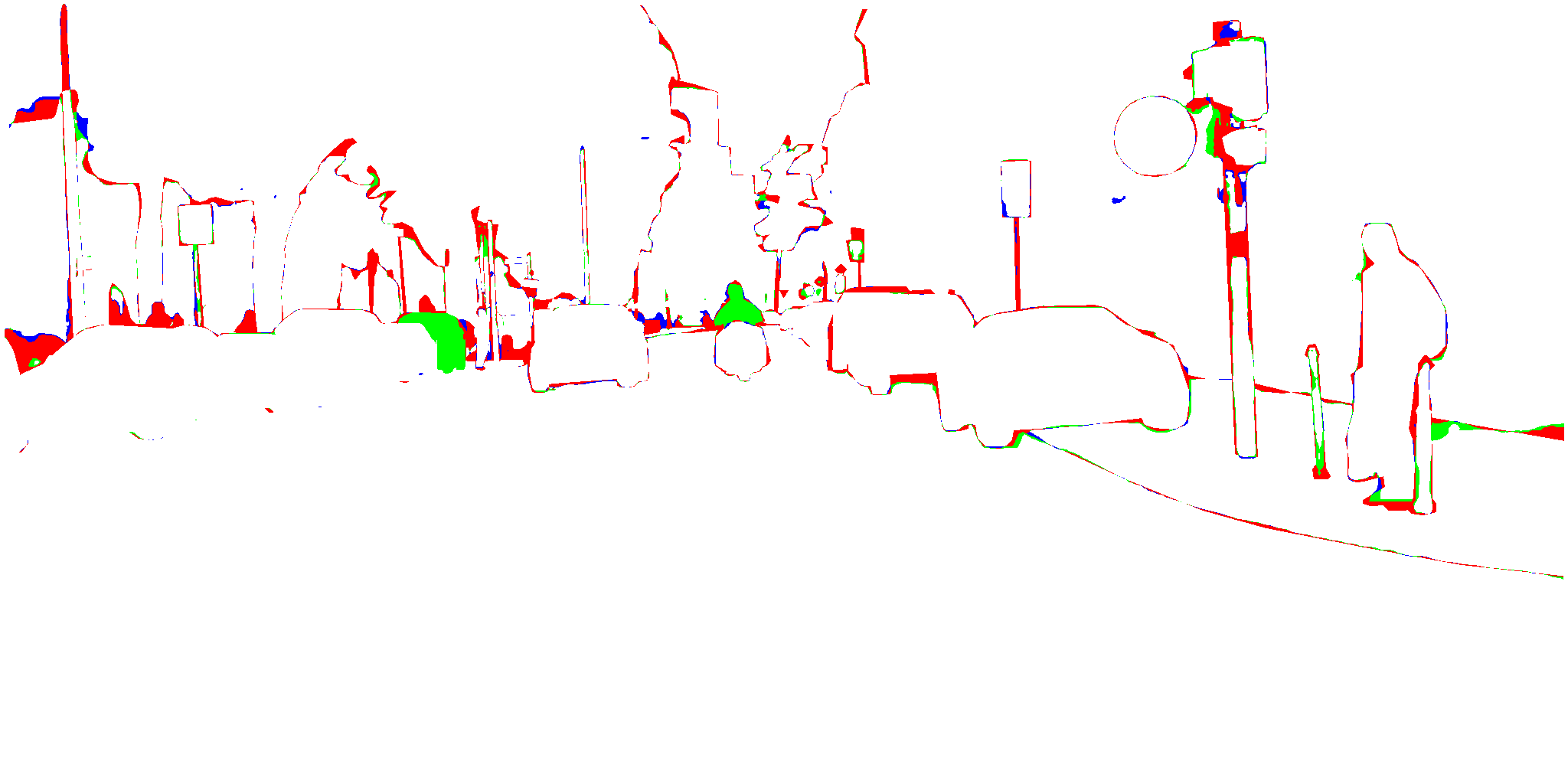}}} \\
\\
\rot{(c) Mapillary Vistas} & \raisebox{-0.4\height}{\includegraphics[width=\linewidth]{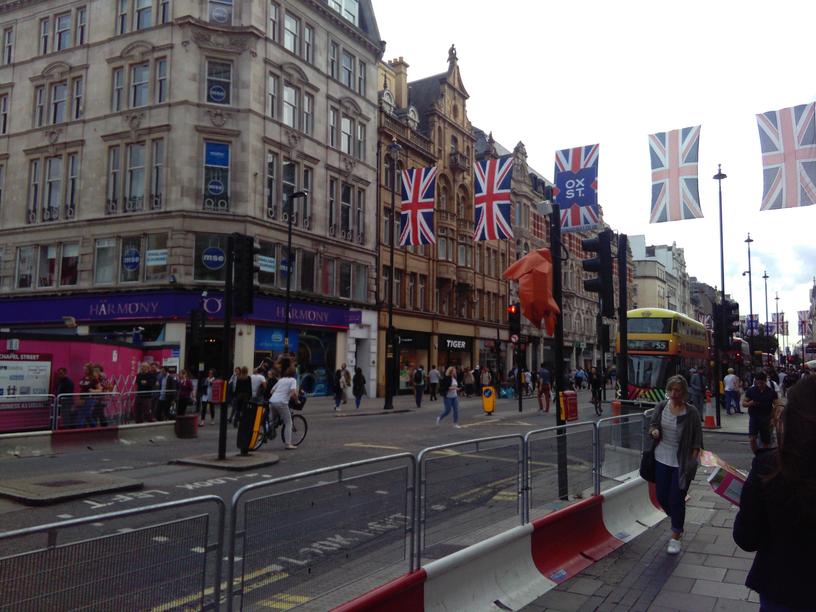}} & \raisebox{-0.4\height}{\includegraphics[width=\linewidth]{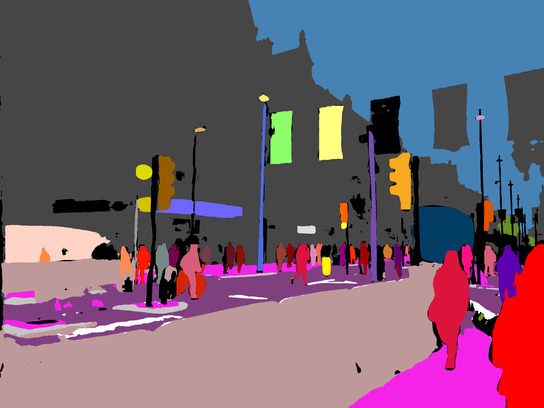}} & \raisebox{-0.4\height}{\includegraphics[width=\linewidth]{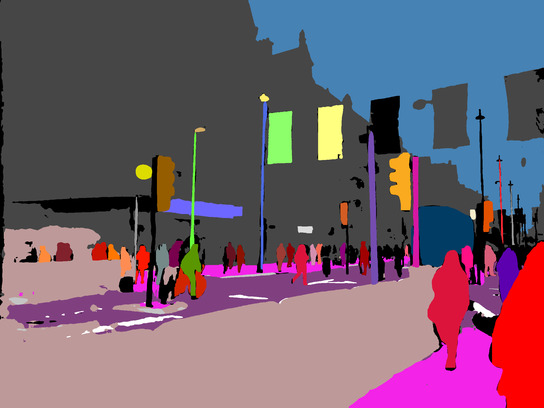}} & \raisebox{-0.4\height}{\fbox{\includegraphics[width=\linewidth]{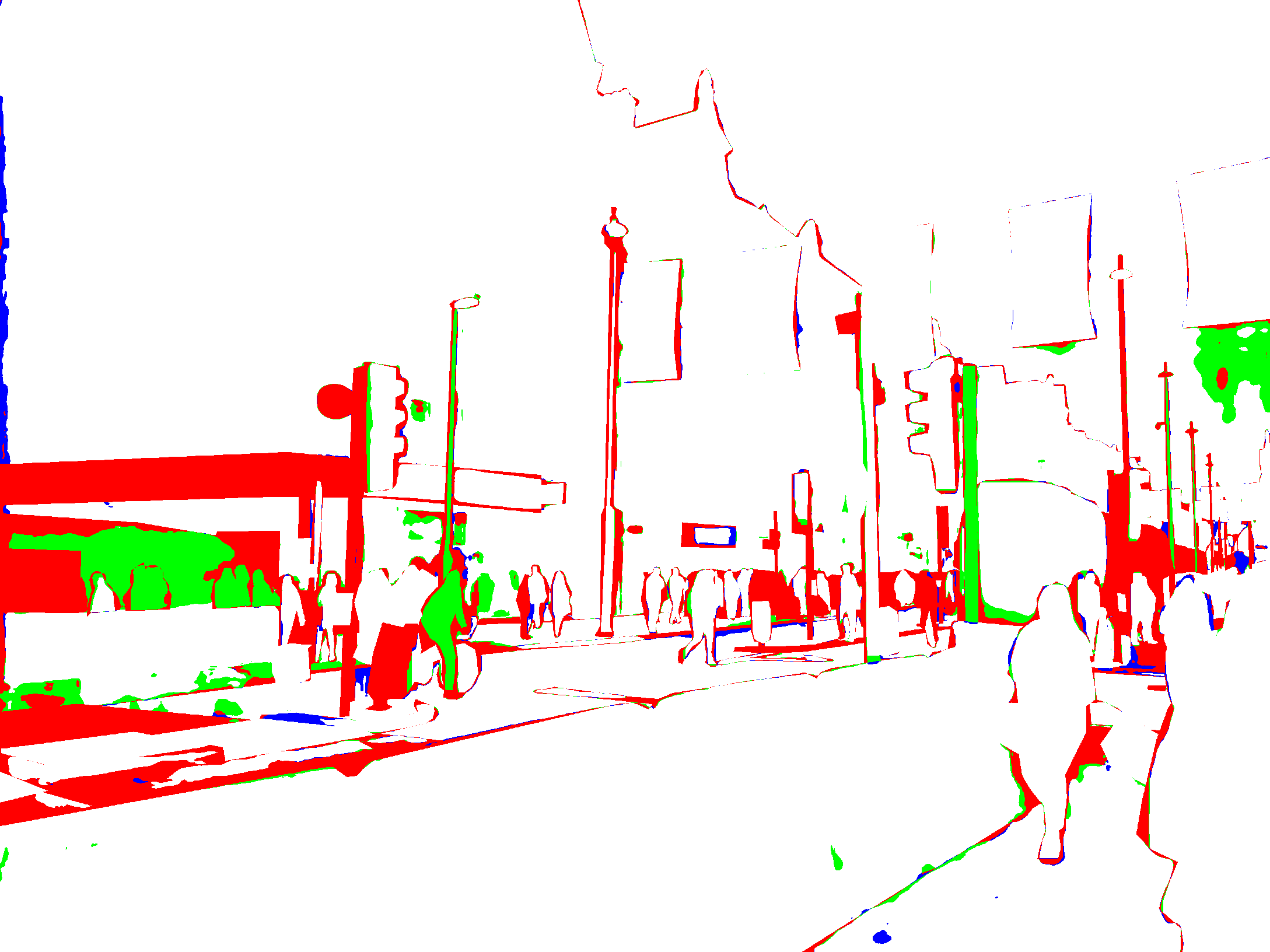}}} \\
\\
\rot{(d) Mapillary Vistas} &
\raisebox{-0.4\height}{\includegraphics[width=\linewidth]{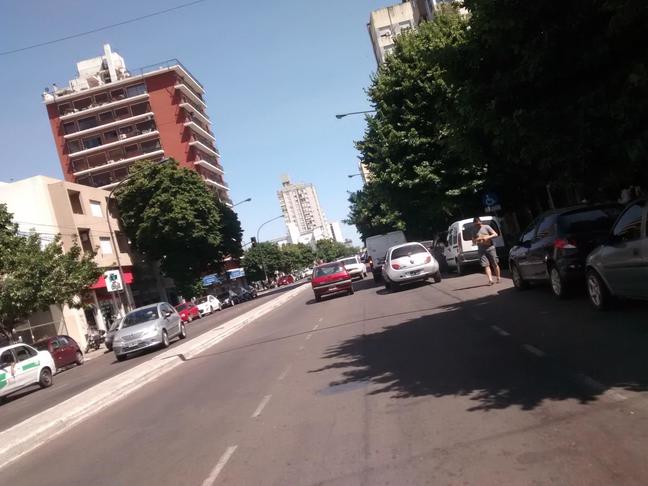}} & \raisebox{-0.4\height}{\includegraphics[width=\linewidth]{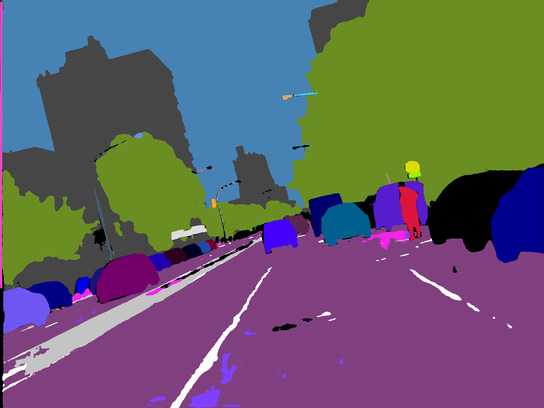}} & \raisebox{-0.4\height}{\includegraphics[width=\linewidth]{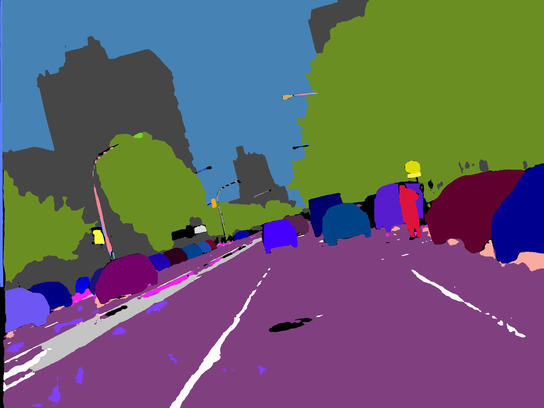}} & \raisebox{-0.4\height}{\fbox{\includegraphics[width=\linewidth]{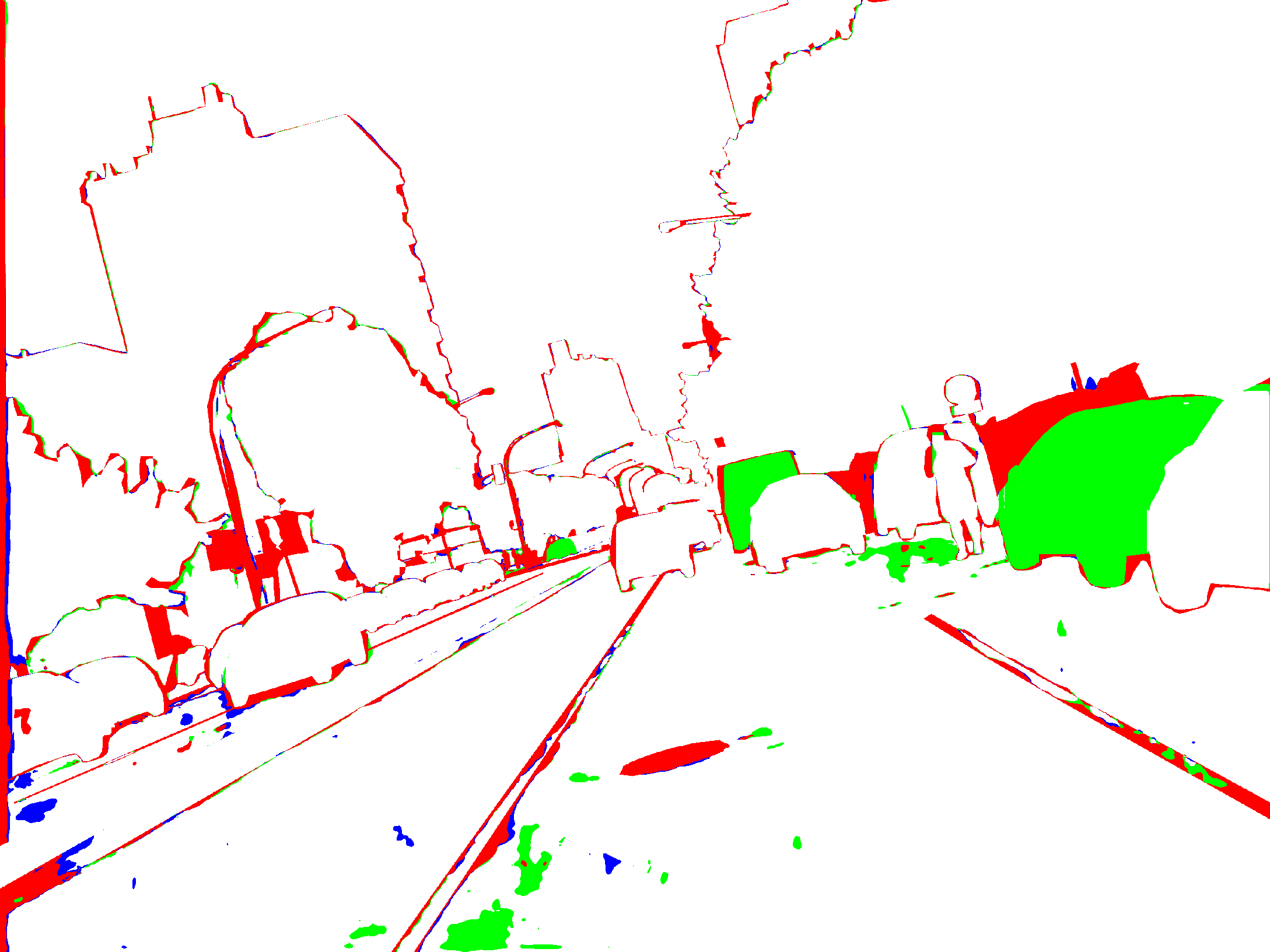}}} \\
\\
\rot{(e) KITTI} & \raisebox{-0.4\height}{\includegraphics[width=\linewidth]{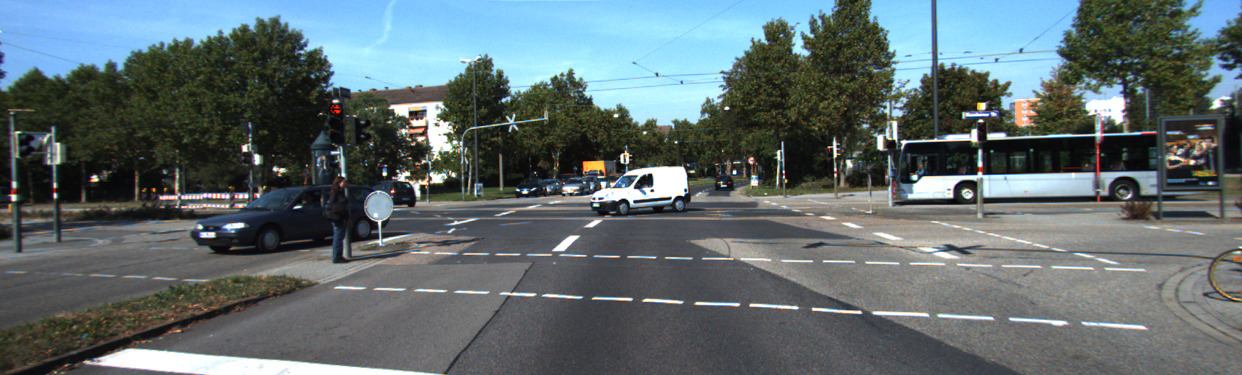}} & \raisebox{-0.4\height}{\includegraphics[width=\linewidth]{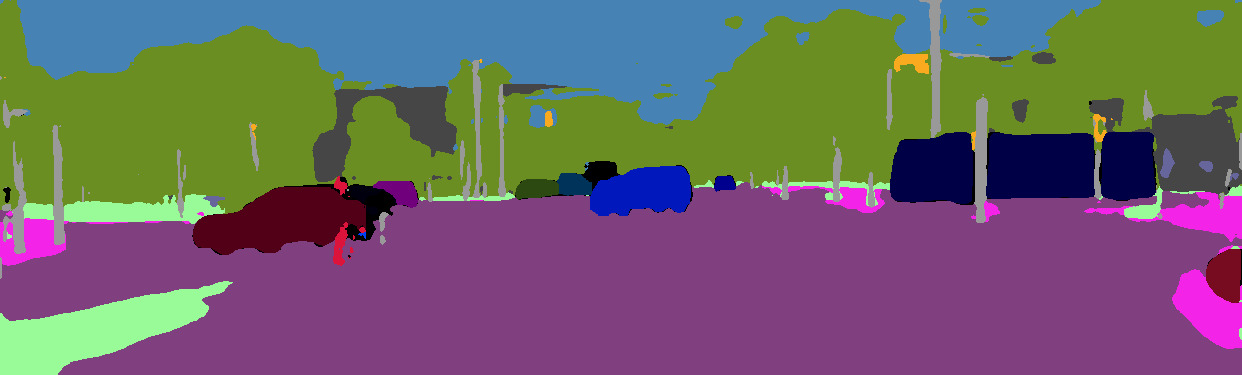}} & \raisebox{-0.4\height}{\includegraphics[width=\linewidth]{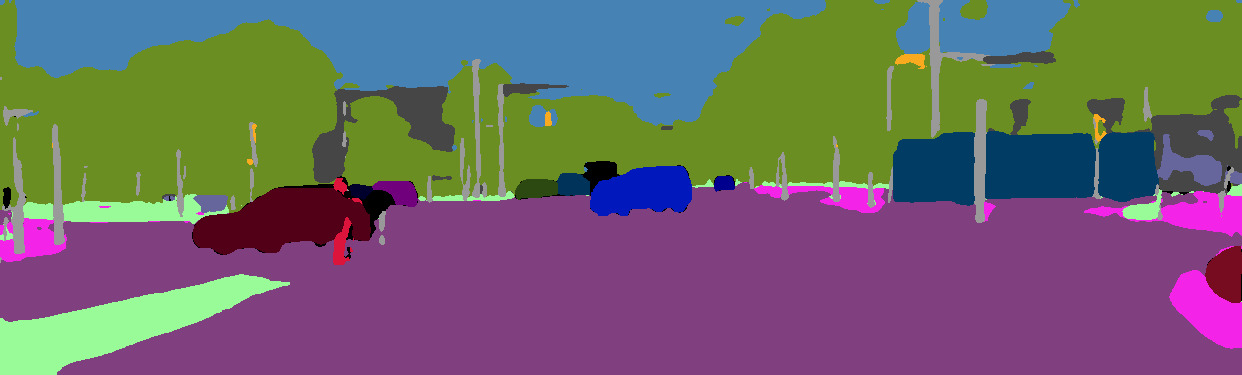}} & \raisebox{-0.4\height}{\fbox{\includegraphics[width=\linewidth]{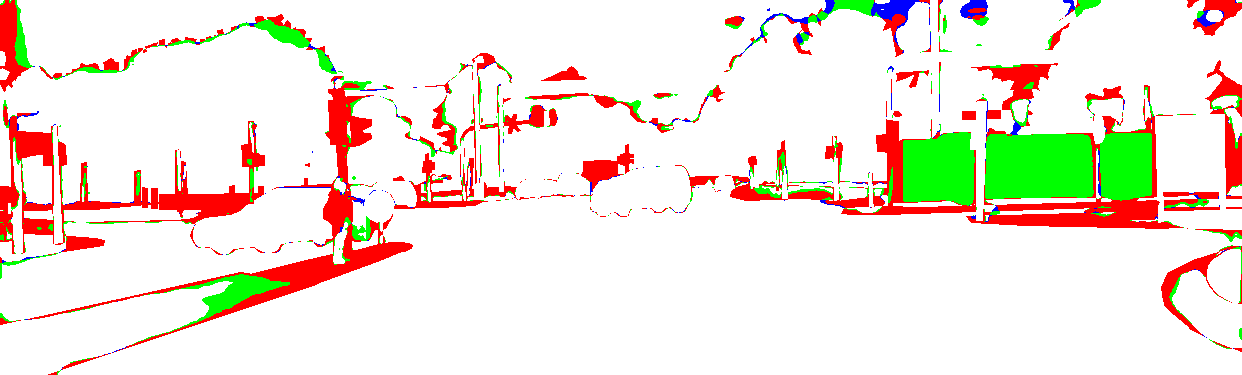}}} \\
\\
\rot{(f) KITTI} & \raisebox{-0.4\height}{\includegraphics[width=\linewidth]{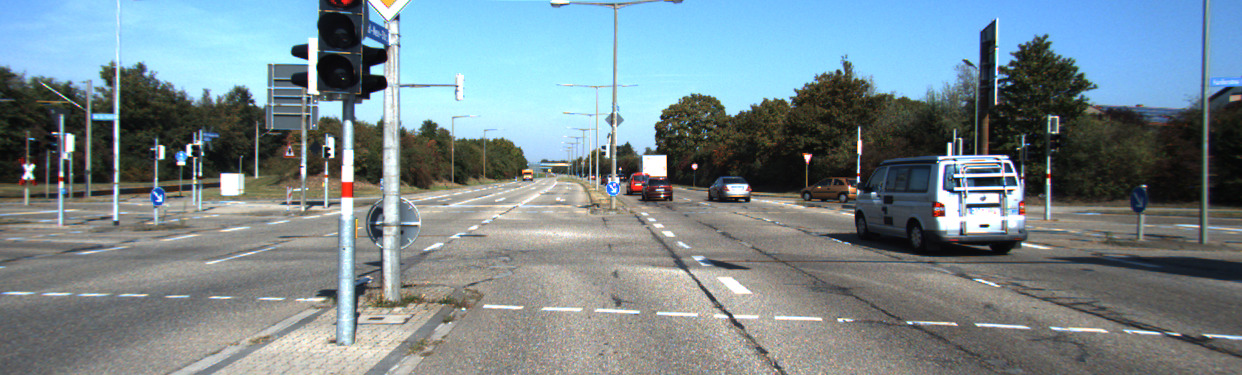}} & \raisebox{-0.4\height}{\includegraphics[width=\linewidth]{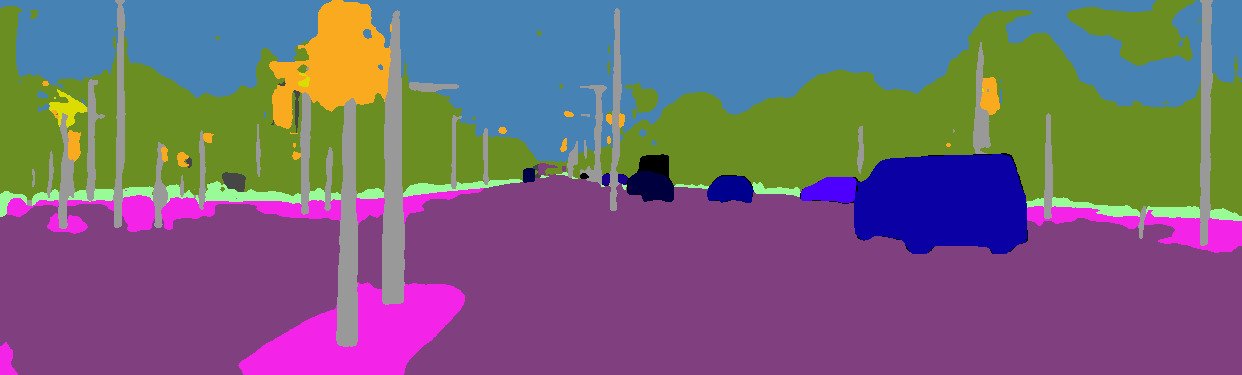}} & \raisebox{-0.4\height}{\includegraphics[width=\linewidth]{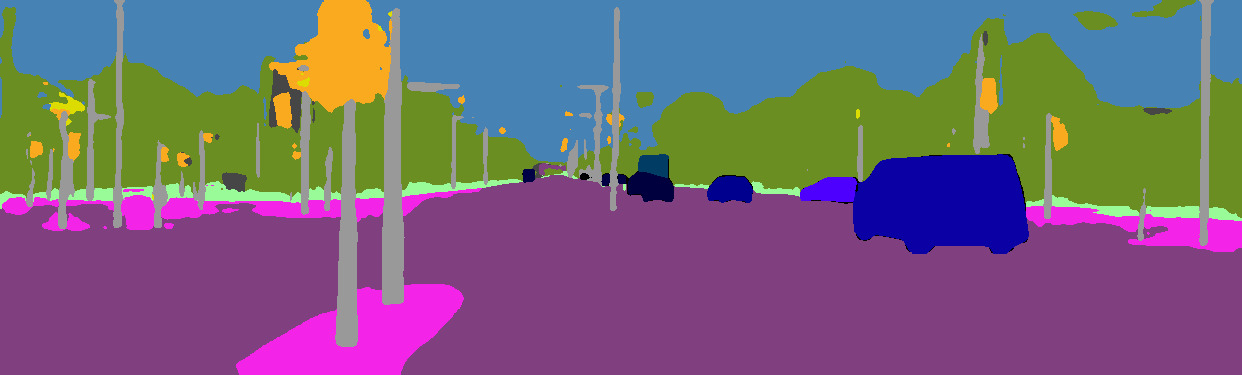}} & \raisebox{-0.4\height}{\fbox{\includegraphics[width=\linewidth]{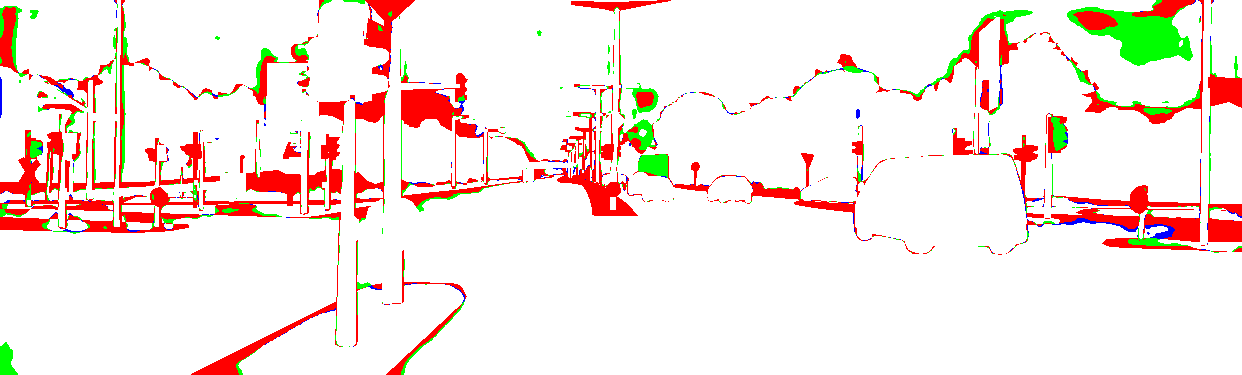}}} \\
\\
\rot{(g) IDD} & \raisebox{-0.4\height}{\includegraphics[width=\linewidth]{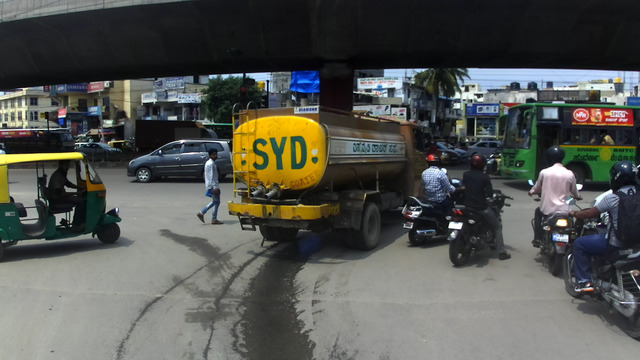}} & \raisebox{-0.4\height}{\includegraphics[width=\linewidth]{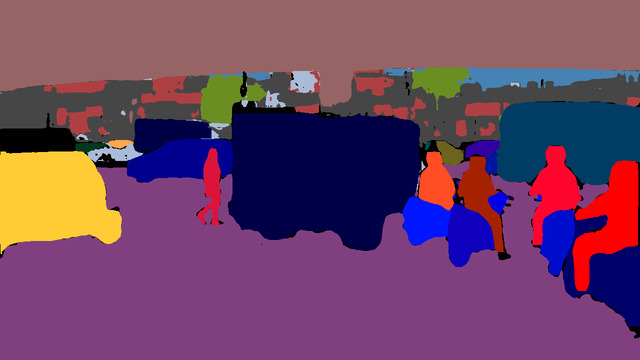}} & \raisebox{-0.4\height}{\includegraphics[width=\linewidth]{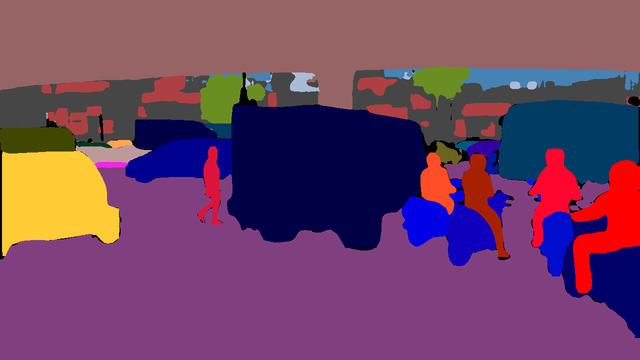}} & \raisebox{-0.4\height}{\fbox{\includegraphics[width=\linewidth]{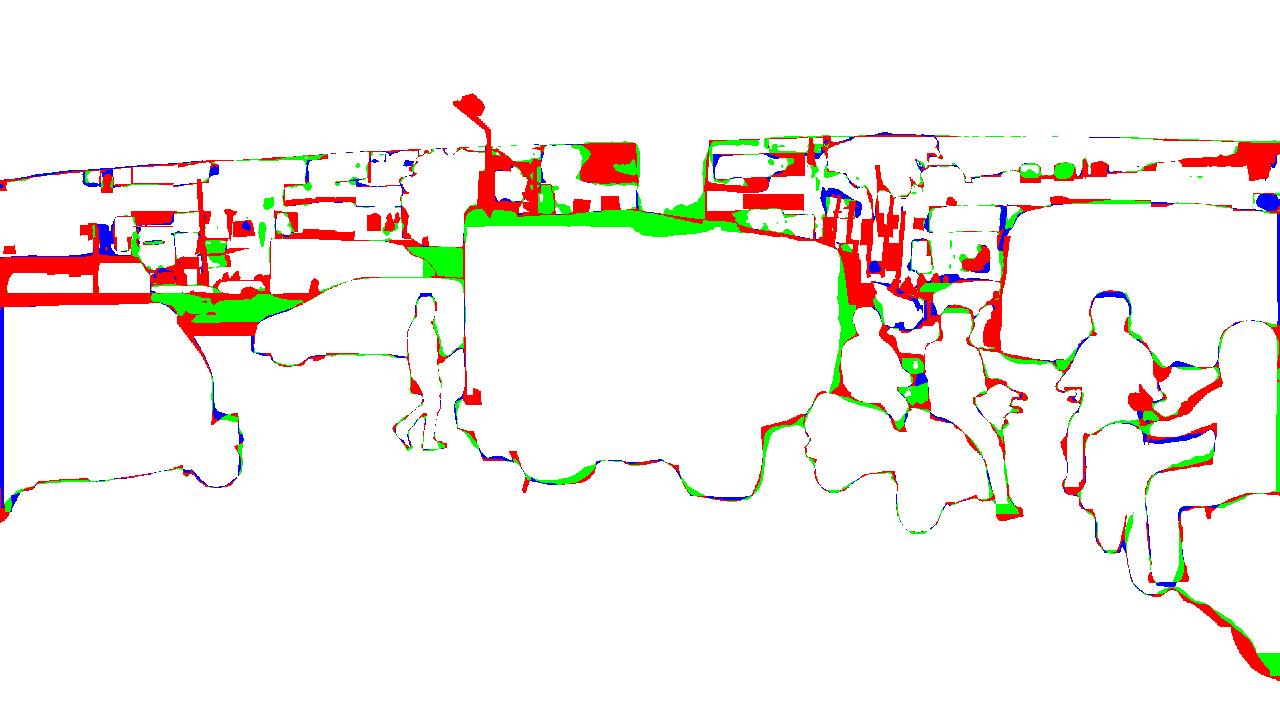}}} \\
\\
\rot{(h) IDD} & \raisebox{-0.4\height}{\includegraphics[width=\linewidth]{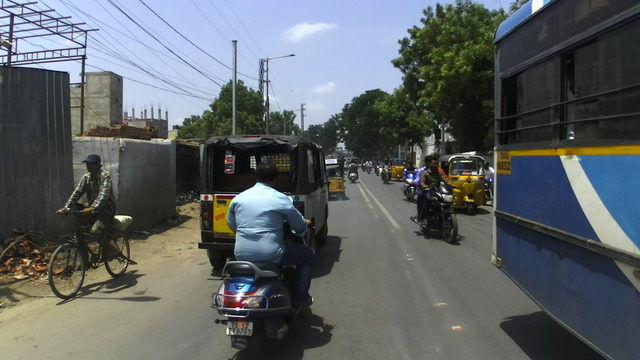}} & \raisebox{-0.4\height}{\includegraphics[width=\linewidth]{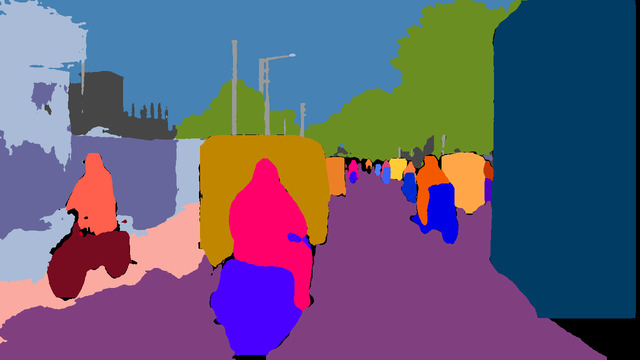}} & \raisebox{-0.4\height}{\includegraphics[width=\linewidth]{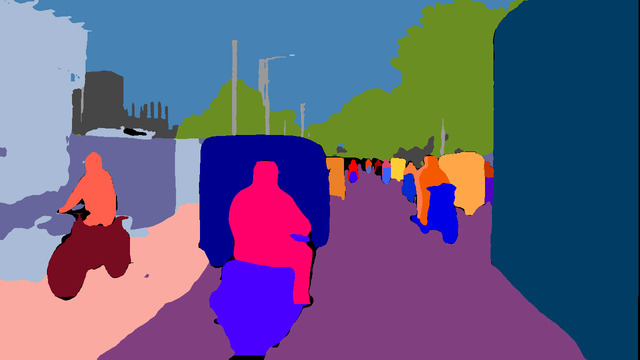}} & \raisebox{-0.4\height}{\fbox{\includegraphics[width=\linewidth]{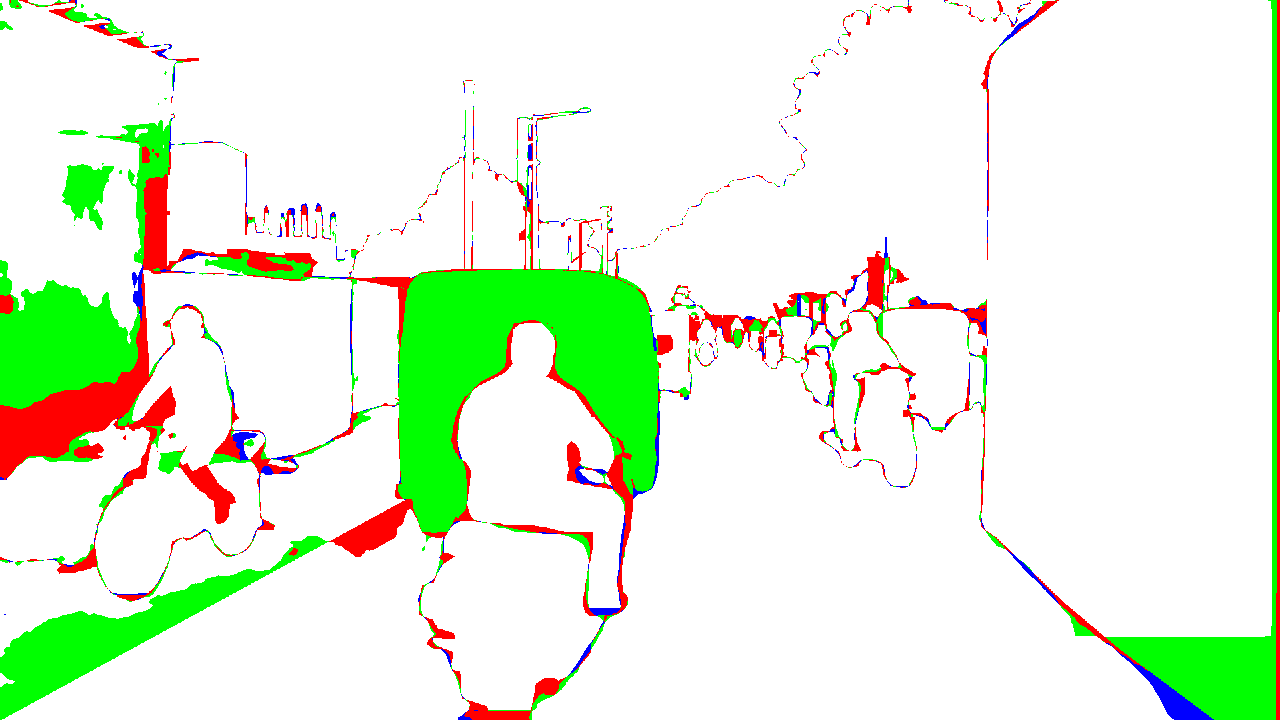}}} \\
\end{tabular}}
\caption{Qualitative panoptic segmentation results of our proposed EfficientPS network in comparison to the state-of-the-art Seamless architecture~\citep{porzi2019seamless} on different benchmark datasets. In addition to the panoptic segmentation output, we also show the improvement\textbackslash{error map} which denotes the pixels that are misclassified by the Seamless model but correctly predicted by the EfficientPS model in green, the pixels that are misclassified by the EfficientPS model but correctly predicted by the Seamless model in blue, and the pixels that are misclassified by both the EfficientPS model and the Seamless model in red.}
\label{fig:qualitativeEPSNet}
\end{figure*}

In this section, we qualitatively evaluate the panoptic segmentation performance of our proposed EfficientPS architecture in comparison to the state-of-the-art Seamless~\citep{porzi2019seamless} model on each of the datasets that we benchmark on. We use the publicly available official implementation of the Seamless architecture to obtain the outputs for the qualitative comparisons. The best performing state-of-the-art model Panoptic-Deeplab does not provide any publicly available implementation or pre-trained models which makes such comparisons infeasible. \figref{fig:qualitativeEPSNet} presents two examples from the validation sets of each of the urban scene understanding dataset. For each example, we show the input image, the corresponding panoptic segmentation output from the Seamless model and our proposed EfficientPS model. Additionally, we show the improvement and error map where a green pixel indicates that our EfficientPS made the right prediction but the Seamless model misclassified it (improvement of EfficientPS over Seamless), a blue pixel indicates that Seamless model made the right prediction but EfficientPS misclassified it, and a red pixel denotes that both models misclassified it with respect to the groundtruth.

\figref{fig:qualitativeEPSNet}~(a) and (b) show examples from the Cityscapes dataset in which the improvement over the Seamless model can be seen in the ability to segment heavily occluded ‘thing’ class instances. In the first example, the truck far behind on the bridge is occluded by cars and a cyclist, and in the second example, the distant car parked on the left side of the image is only partially visible as the car in the front occludes it. We observe from the improvement maps that our proposed EfficientPS model accurately detect, classify and segment these instances, while the Seamless model misclassifies these pixels. This can be primarily attributed to our 2-way FPN that effectively aggregates multi-scale features to learn semantically richer representations and the panoptic fusion module that addresses the instance overlap ambiguity in an adaptive manner.

In \figref{fig:qualitativeEPSNet}~(c) and (d), we qualitatively compare the performance on the challenging Mapillary Vistas dataset. We observe that in \figref{fig:qualitativeEPSNet}~(c) the group of people towards left side of the image who are behind the fence are misclassified in the output of the Seamless model and the instances of these people are not detected. Whereas, our EfficientPS model accurately segments each of the instances of the people. Similarly, the distant van on the right side of the image shown in \figref{fig:qualitativeEPSNet}~(d) is partially occluded by the neighboring cars and is entirely misclassified by the Seamless model. However, our EfficientPS model accurately captures this heavily occluded object instance. In \figref{fig:qualitativeEPSNet}~(c), interestingly, the Seamless model misclassifies the cyclist on the road as a pedestrian. We hypothesize that this might be due to the fact that one of the legs of the cyclist is touching the ground and the other leg which is on the pedal of the bicycle is barely visible. Hence, this causes the Seamless model to misclassify the object instance. Whereas, our EfficientPS model effectively leverages both the semantic and instance prediction in our panoptic fusion module to accurately address this ambiguity in the scene. We also observe in \figref{fig:qualitativeEPSNet}~(c) that the EfficientPS model misclassifies the traffic sign fixed on the fence and only partially segments the advertisement board attached to the building near the fence while it accurately segments all the other instances of this class. This is primarily due to the fact that there is a lack of relevant examples for this type of traffic sign which is atypical of those found in the training set.

\figref{fig:qualitativeEPSNet}~(e) and (f) show qualitative comparisons on the KITTI dataset. In \figref{fig:qualitativeEPSNet}~(e), we see that the Seamless model misclassifies the bus that is towards the right of the image as a truck although it segments the object coherently. This is primarily due to the fact that there are poles as well as an advertisement board in front of the bus which divides the it into different subregions. This leads the model to predict it as a truck that has a transition between the tractor unit and the trailer. However, our proposed EfficientPS model mitigates this problem with its bidirectional aggregation of multi-scale features that effectively captures contextual information. In \figref{fig:qualitativeEPSNet}~(f), we observe that a distant truck on the right lane is partially occluded by cars behind it which causes the Seamless model to not detect the truck as a new instance, rather it detects the truck and the car behind it as being the same object. This  is similar to the scenario observed on the Cityscapes dataset in \figref{fig:qualitativeEPSNet}~(a). Nevertheless, our proposed EfficientPS model yields accurate predictions in such challenging scenarios consistently across different datasets.

\begin{figure*}
\centering
\footnotesize
\setlength{\tabcolsep}{0.1cm}
{\renewcommand{\arraystretch}{1}
\begin{tabular}{P{0.4cm}P{3.9cm}P{3.9cm}P{3.9cm}P{3.9cm}}
& \raisebox{-0.4\height}{(i)} & \raisebox{-0.4\height}{(ii)} & \raisebox{-0.4\height}{(iii)} & \raisebox{-0.4\height}{(iv)} \\
\\
\rot{(a) Cityscapes} & \raisebox{-0.4\height}{\includegraphics[width=\linewidth]{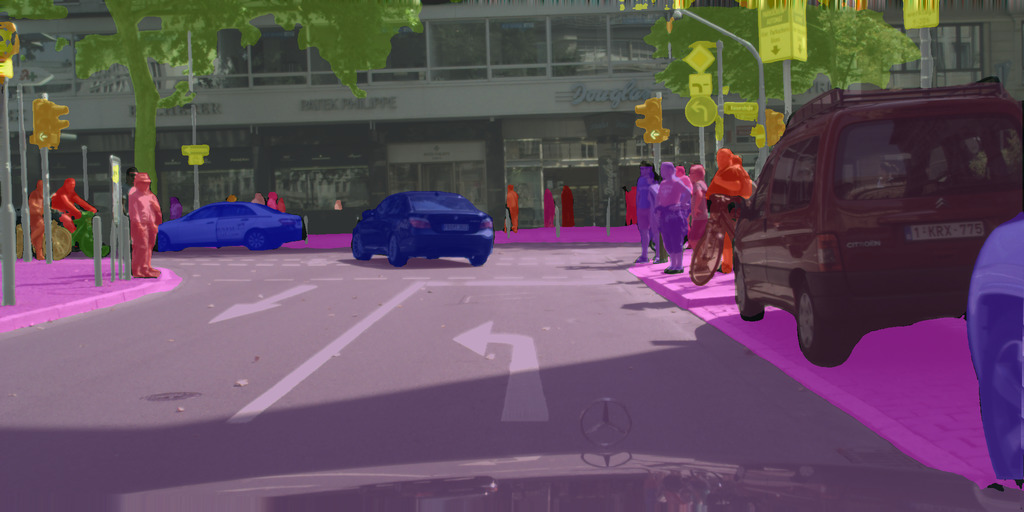}} & \raisebox{-0.4\height}{\includegraphics[width=\linewidth]{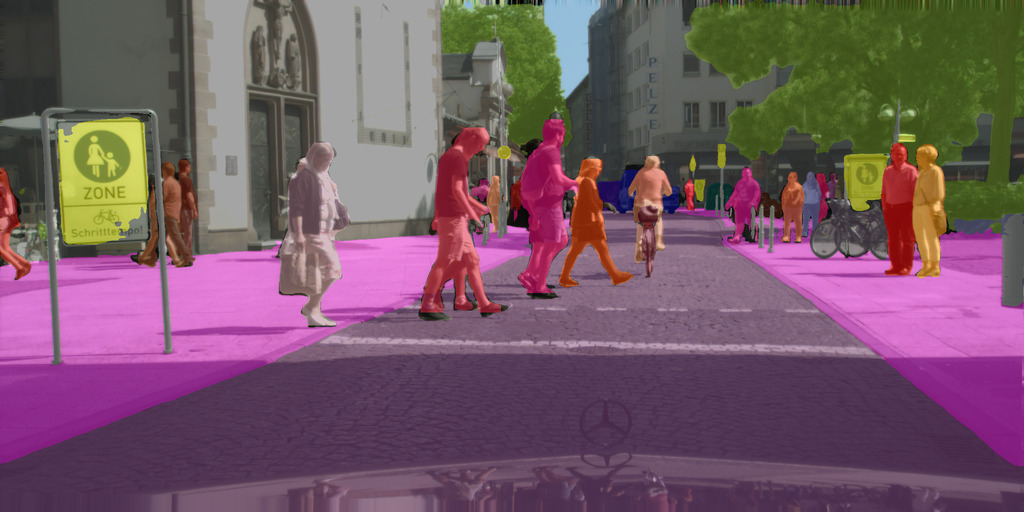}} & \raisebox{-0.4\height}{\includegraphics[width=\linewidth]{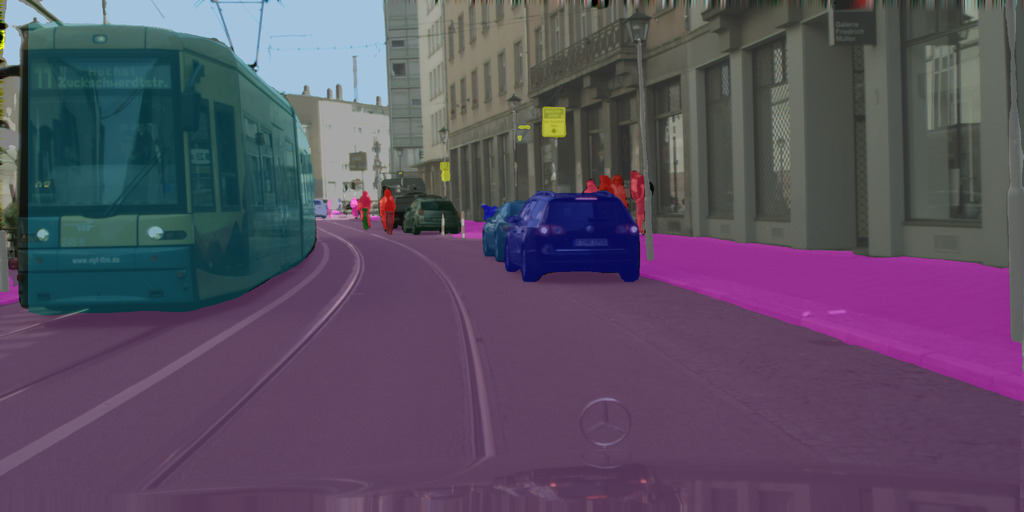}} & \raisebox{-0.4\height}{\fbox{\includegraphics[width=\linewidth]{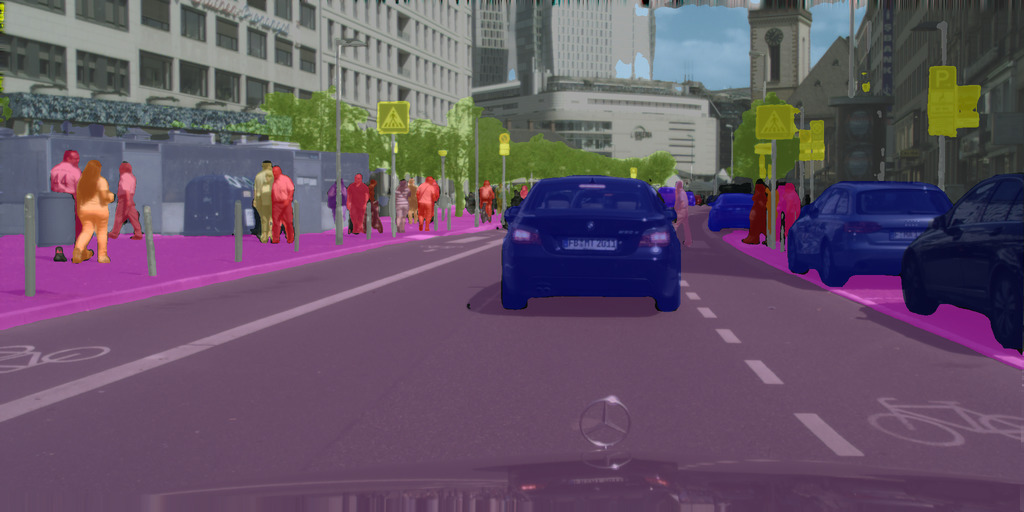}}} \\
\\
\rot{(b) Cityscapes} & \raisebox{-0.4\height}{\includegraphics[width=\linewidth]{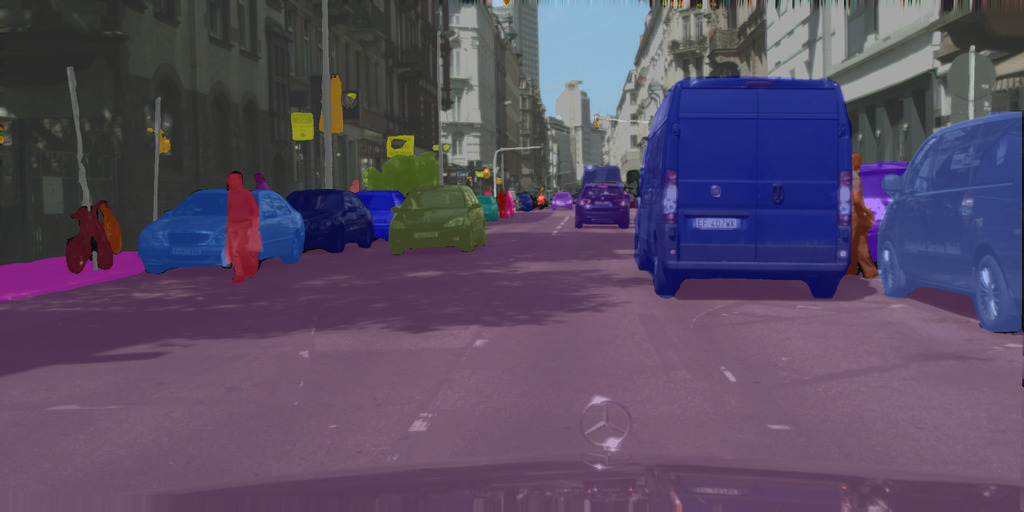}} & \raisebox{-0.4\height}{\includegraphics[width=\linewidth]{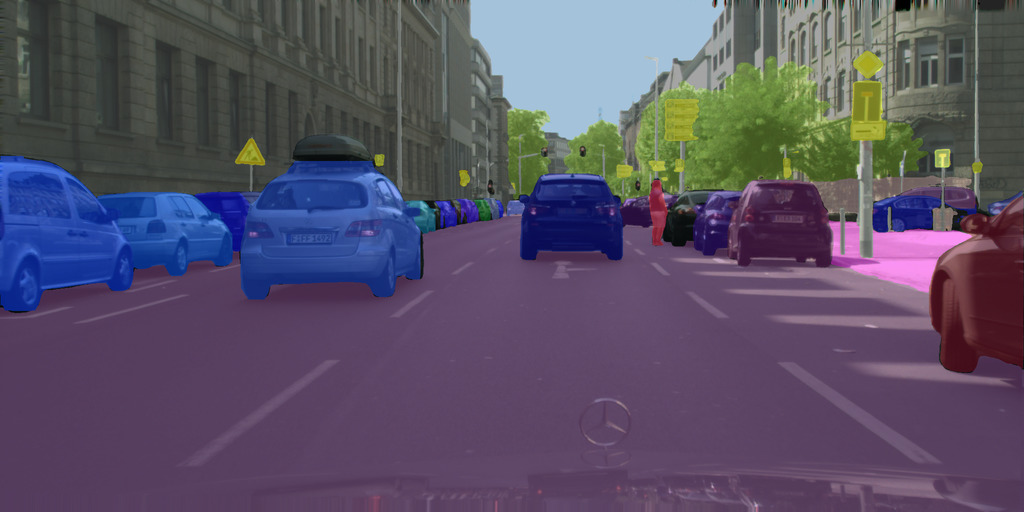}} & \raisebox{-0.4\height}{\includegraphics[width=\linewidth]{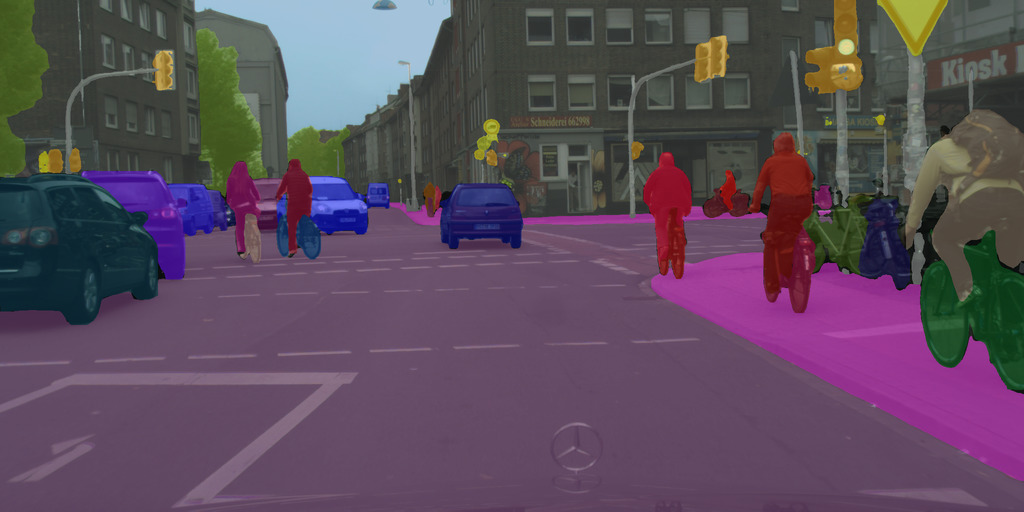}} & \raisebox{-0.4\height}{\fbox{\includegraphics[width=\linewidth]{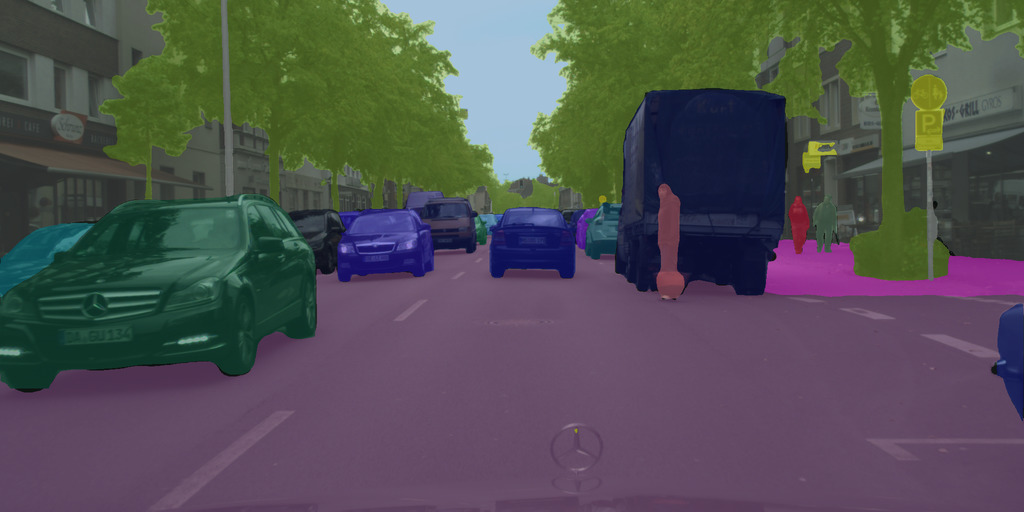}}} \\
\\
\rot{(c) Mapillary Vistas} & \raisebox{-0.4\height}{\includegraphics[width=\linewidth]{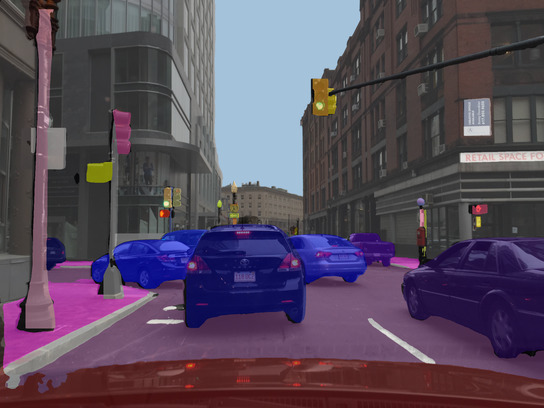}} & \raisebox{-0.4\height}{\includegraphics[width=\linewidth]{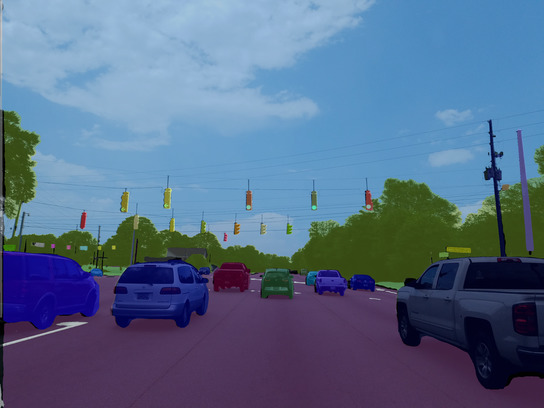}} & \raisebox{-0.4\height}{\includegraphics[width=\linewidth]{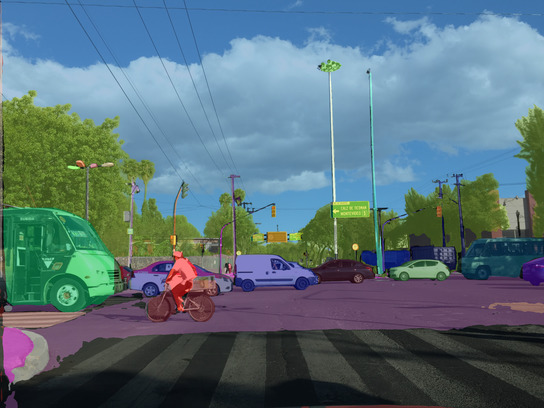}} & \raisebox{-0.4\height}{\fbox{\includegraphics[width=\linewidth]{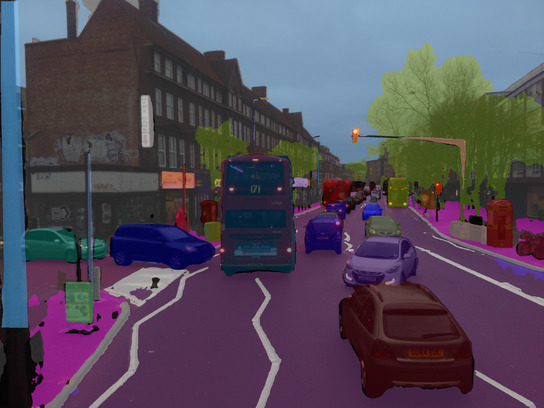}}} \\
\\
\rot{(d) Mapillary Vistas} &
\raisebox{-0.4\height}{\includegraphics[width=\linewidth]{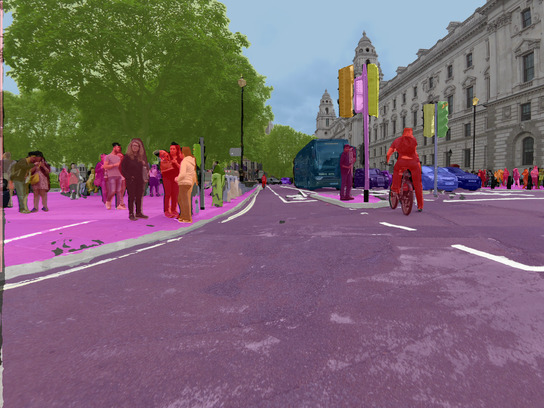}} & \raisebox{-0.4\height}{\includegraphics[width=\linewidth]{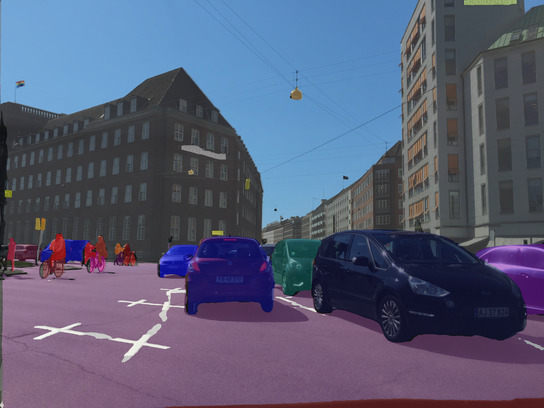}} & \raisebox{-0.4\height}{\includegraphics[width=\linewidth]{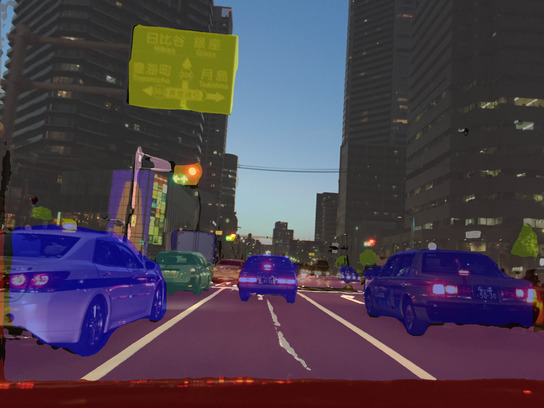}} & \raisebox{-0.4\height}{\fbox{\includegraphics[width=\linewidth]{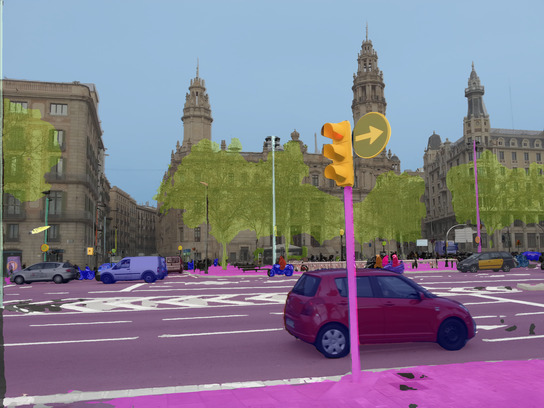}}} \\
\\
\rot{(e) KITTI} & \raisebox{-0.4\height}{\includegraphics[width=\linewidth]{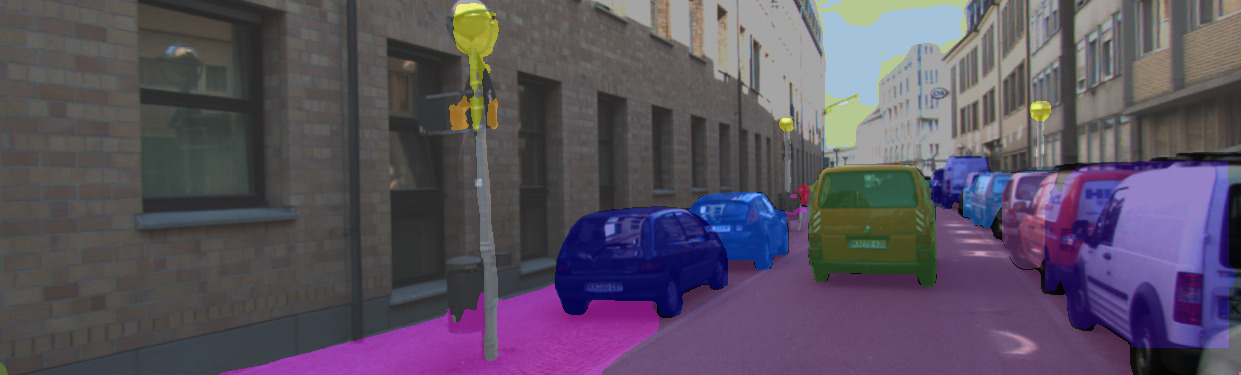}} & \raisebox{-0.4\height}{\includegraphics[width=\linewidth]{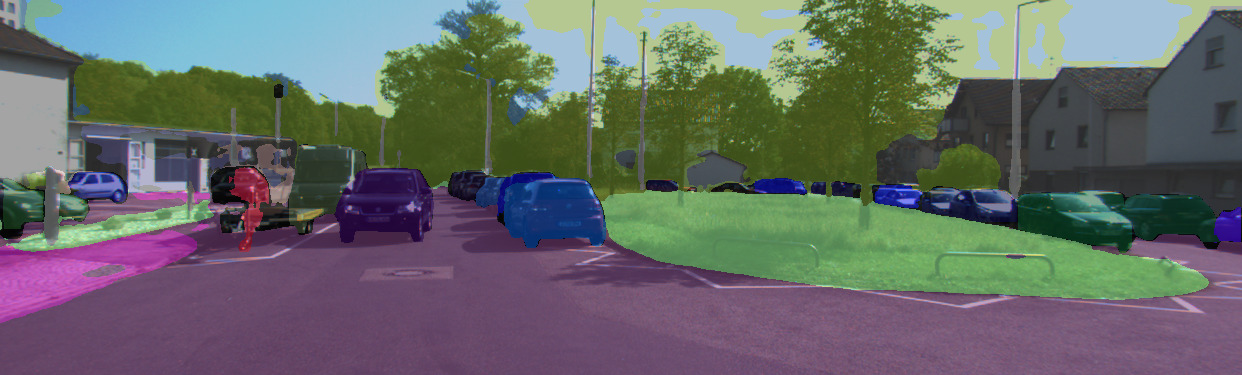}} & \raisebox{-0.4\height}{\includegraphics[width=\linewidth]{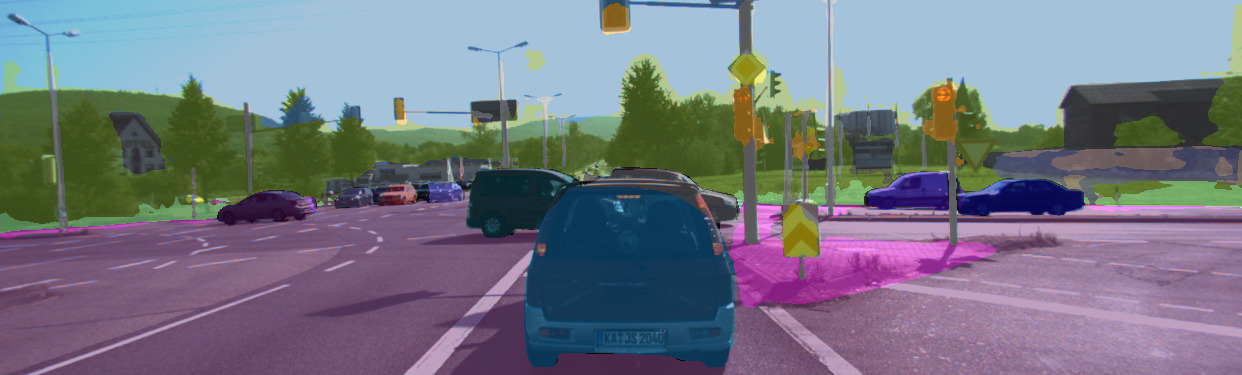}} & \raisebox{-0.4\height}{\fbox{\includegraphics[width=\linewidth]{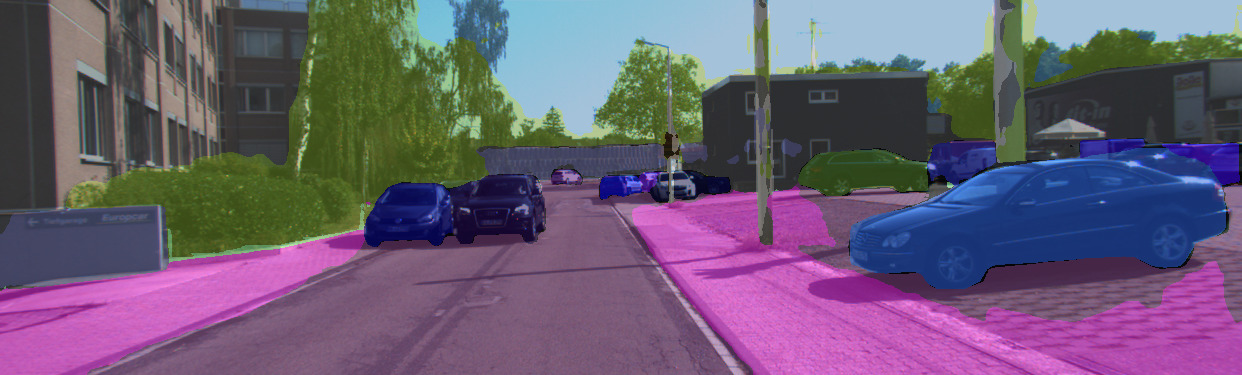}}} \\
\\
\rot{(f) KITTI} & \raisebox{-0.4\height}{\includegraphics[width=\linewidth]{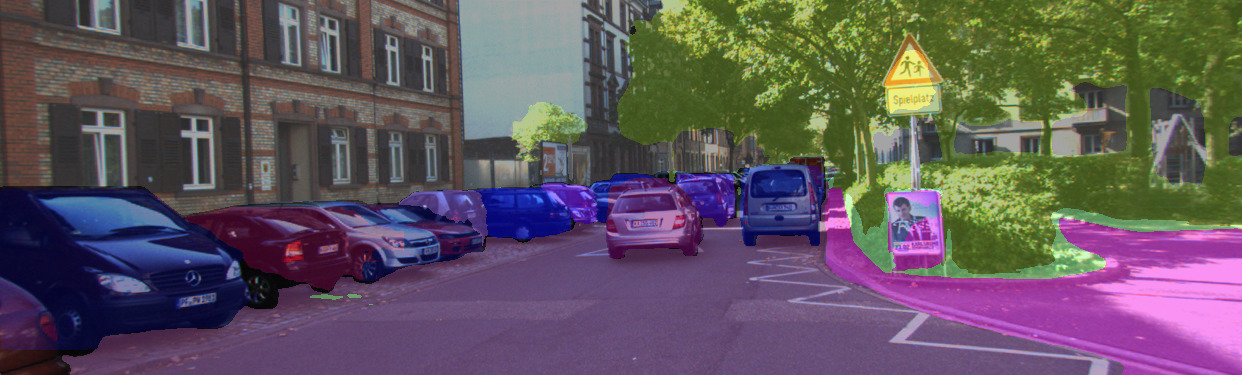}} & \raisebox{-0.4\height}{\includegraphics[width=\linewidth]{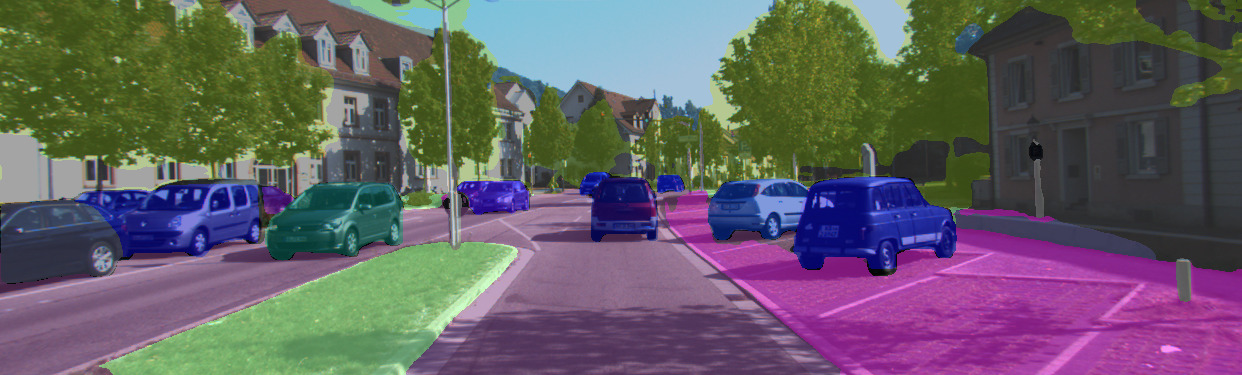}} & \raisebox{-0.4\height}{\includegraphics[width=\linewidth]{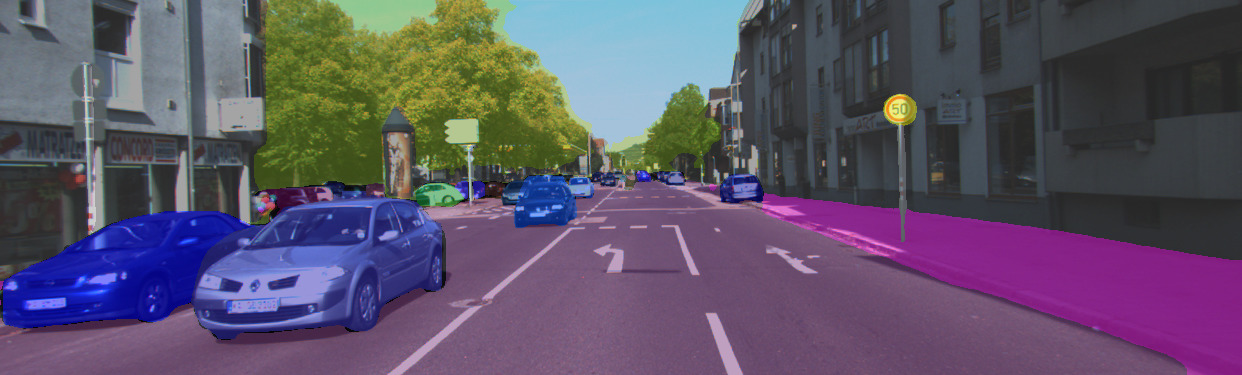}} & \raisebox{-0.4\height}{\fbox{\includegraphics[width=\linewidth]{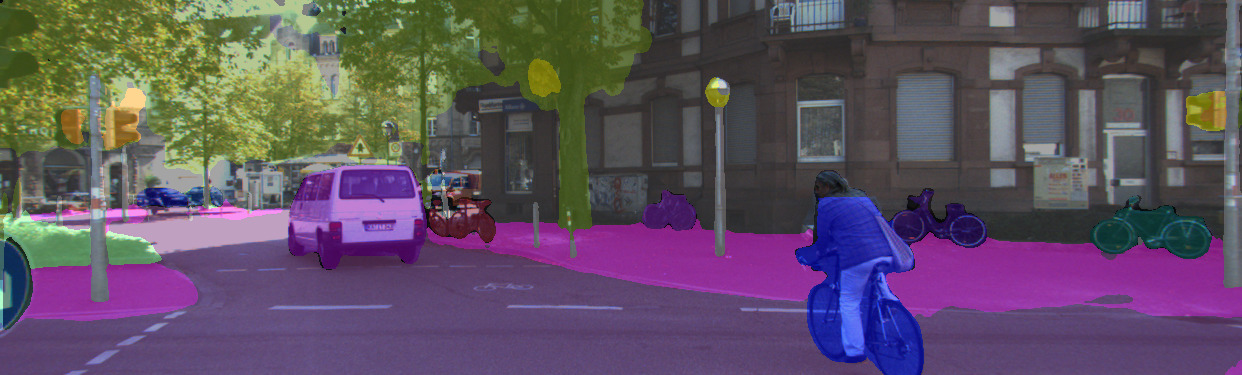}}} \\
\\
\rot{(g) IDD} & \raisebox{-0.4\height}{\includegraphics[width=\linewidth]{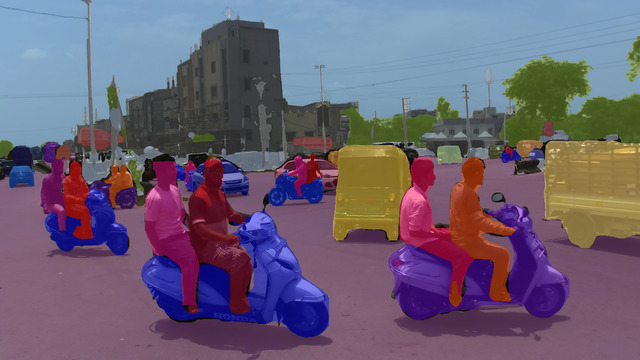}} & \raisebox{-0.4\height}{\includegraphics[width=\linewidth]{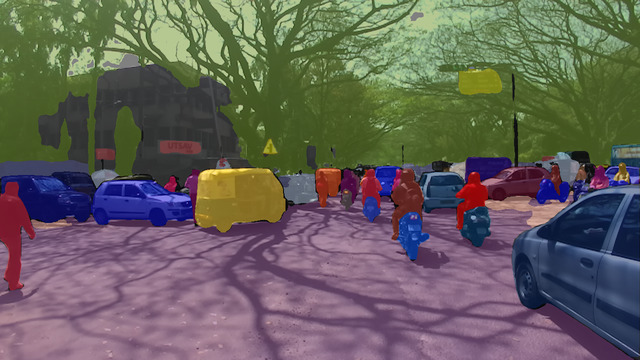}} & \raisebox{-0.4\height}{\includegraphics[width=\linewidth]{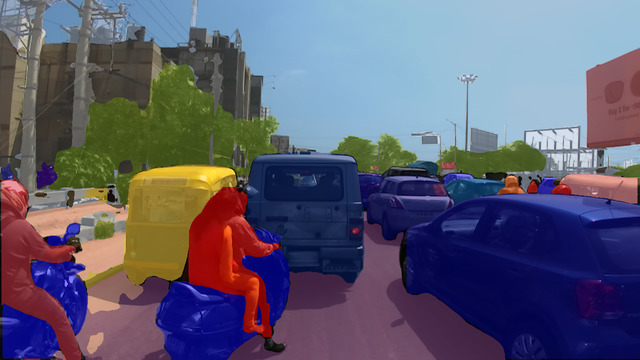}} & \raisebox{-0.4\height}{\fbox{\includegraphics[width=\linewidth]{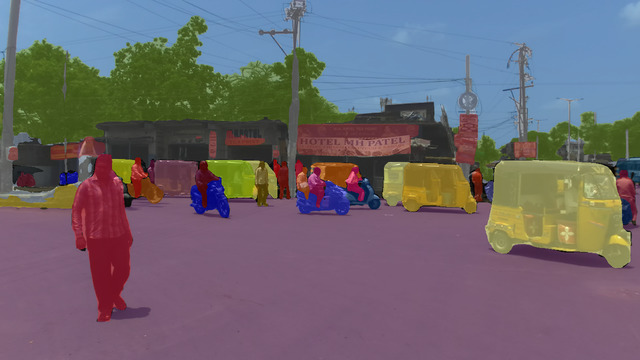}}} \\
\\
\rot{(h) IDD} & \raisebox{-0.4\height}{\includegraphics[width=\linewidth]{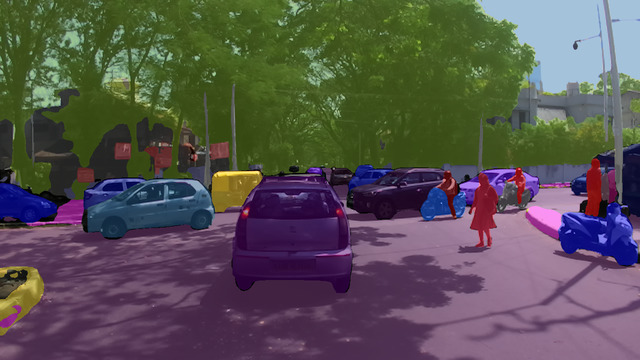}} & \raisebox{-0.4\height}{\includegraphics[width=\linewidth]{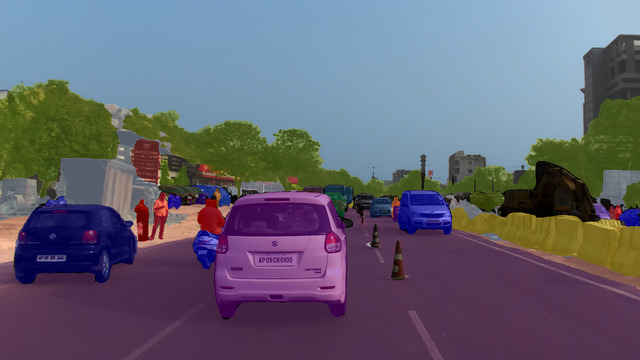}} & \raisebox{-0.4\height}{\includegraphics[width=\linewidth]{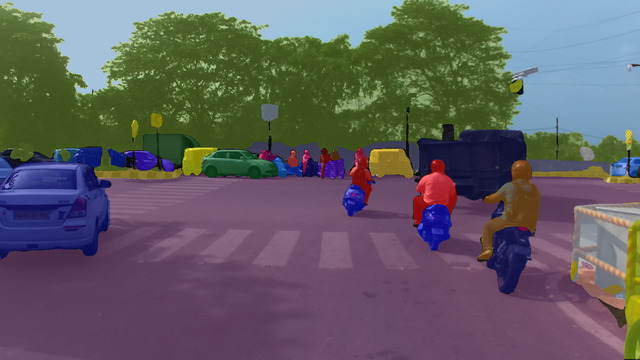}} & \raisebox{-0.4\height}{\fbox{\includegraphics[width=\linewidth]{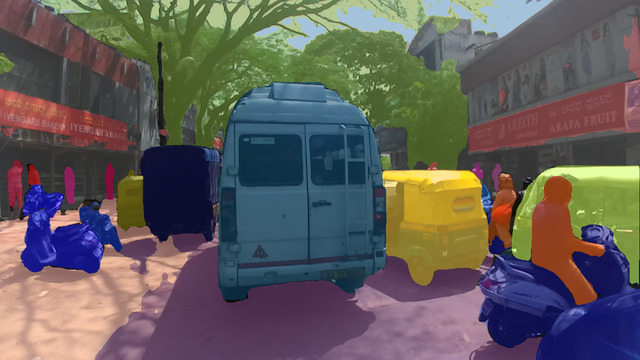}}} \\
\end{tabular}}
\caption{Visual panoptic segmentation results of our proposed EfficientPS model on each of the challenging urban scene understanding datasets that we benchmark on which in total encompasses scenes from over 50 countries. These examples show complex urban scenarios with numerous object instances in multiple scales and with partial occlusion. These scenes also show diverse lighting conditions from dawn to dusk as well as seasonal changes.}
\label{fig:visualizationEpsnet}
\end{figure*}
  
In \figref{fig:qualitativeEPSNet}~(g) and (h), we present examples from the IDD dataset. We can see that our EfficientPS model captures the boundaries of ‘stuff’ classes more precisely than the Seamless model in both the examples. For instance, the pillar of the bridge in \figref{fig:qualitativeEPSNet}~(g) and the extent of the sidewalk in \figref{fig:qualitativeEPSNet}~(h) are more well defined in the panoptic segmentation output of our EfficientPS model. This can be attributed to the object boundary refinement ability of our semantic head that correlates features of different scales before fusing them. In \figref{fig:qualitativeEPSNet}~(h), the Seamless model misclassifies the auto-rickshaw as a caravan due to the similar visual appearances of these two objects, however our proposed EfficientPS model with our novel panoptic backbone has an extensive representational capacity which enables it to accurately classify objects even with such subtle differences. We observe that although the upper half of the cyclist towards the left of the image is accurately segmented, the front leg of the cyclist is misclassifies as being part of the bicycle. This is a challenging scenario due to the high contrast in this region. We also observe that the boundary of the sidewalk towards the left of the auto rickshaw is misclassified. However, on visual inspection of the groundtruth, it appears that the sidewalk boundary in this region is mislabeled in groundtruth mask, while the model is making a reasonable prediction.

\subsection{Visualizations}
\label{sec:visualization}

We present visualizations of panoptic segmentation results from our proposed EfficientPS architecture on Cityscapes, Mapillary Vistas, KITTI, and Indian Driving Dataset (IDD) in \figref{fig:visualizationEpsnet}. The figures show the panoptic segmentation output of our EfficientPS model using single scale evaluation, which is overlaid on the input image. \figref{fig:visualizationEpsnet}~(a) and (b) show examples from the Cityscapes dataset which exhibit complex road scenes consisting of a large number of traffic participants. These examples show challenging scenarios with dynamic as well as static pedestrian groups in close proximity to each other and distant parked cars that are barely visible due to their neighbouring ‘thing’ class instances. Our proposed EfficientPS architecture effectively addresses these challenges and yields reliable panoptic segmentation results. In \figref{fig:visualizationEpsnet}~(c) and (d), we present results on the Mapillary Vistas dataset that show drastic viewpoint variations and scenes in different times of day. \figref{fig:visualizationEpsnet}~(c.iv), (d.i) and (d.iv) show scenes that were captured from uncommon viewpoints from those observed in the training data and \figref{fig:visualizationEpsnet}~(d.iii) shows a scene that was captured during nighttime. Nevertheless, our EfficientPS model demonstrates substantial robustness against these perceptual variations.

In \figref{fig:visualizationEpsnet}~(e) and (f), we present results on the KITTI dataset which show residential and highway road scenes consisting of several parked and dynamic cars, as well as a large amount of thin structures such as poles. We observe that our EfficientPS model generalizes effectively to these complex scenes even when the network was only trained on the relatively small dataset. \figref{fig:visualizationEpsnet}~(g) and (h) show examples from the IDD dataset that highlight challenges of an unstructured environment. One such challenge is the accurate segmentation of sidewalks, as the transition between the road and the sidewalk is not well delineated often caused by a layer of sand over asphalt. The examples also show heavy traffic with numerous types of vehicles, motorcycles and pedestrians scattered all over the scene. However, our proposed EfficientPS model shows exceptional robustness in these immensely challenging scenes thereby demonstrating its suitability for autonomous driving applications.

\section{Conclusions}
\label{sec:conclusion}

In this paper, we presented our EfficientPS architecture for panoptic segmentation that achieves state-of-the-art performance while being computationally efficient. It incorporates our proposed panoptic backbone with a variant of Mask R-CNN augmented with depthwise separable convolutions as the instance head, a new semantic head that captures fine and contextual features efficiently, and our novel adaptive panoptic fusion module. We demonstrated that our panoptic backbone consisting of the modified EfficientNet encoder and our 2-way FPN achieves the right trade-off between performance and computational complexity. Our 2-way FPN achieves effective aggregation of semantically rich multi-scale features due to its bidirectional flow of information. Thus in combination with our encoder, it establishes a new strong panoptic backbone. We proposed a new semantic head that employs scale-specific feature aggregation to capture long-range context and characteristic features effectively, followed by correlating them to achieve better object boundary refinement capability. We also introduced our parameter-free panoptic fusion module that dynamically fuses logits from both heads based on their mask confidences and congruously integrates instance-specific ‘thing’ classes with ‘stuff’ classes to yield the panoptic segmentation output. 

Additionally, we introduced the KITTI panoptic segmentation dataset that contains panoptic groundtruth annotations for images from the challenging KITTI benchmark. We hope that our panoptic annotations complement the suite of other perception tasks in KITTI and encourage the research community to develop novel multi-task learning methods that include panoptic segmentation. We presented exhaustive benchmarking results on Cityscapes, Mapillary Vistas, KITTI and IDD datasets that demonstrate that our proposed EfficientPS sets the new state-of-the-art in panoptic segmentation while being faster and more parameter efficient than existing state-of-the-art architectures. In addition to being ranked first on the Cityscapes panoptic segmentation leaderboard, our model is ranked second on both the Cityscapes semantic segmentation and instance segmentation leaderboards. We also presented detailed ablation studies, qualitative analysis and visualizations that highlight the improvements that we make to various core modules of panoptic segmentation architectures. To the best of our knowledge, this work is the first to benchmark on all the four standard urban scene understanding datasets that support panoptic segmentation and exceed the state-of-the-art on each of them while simultaneously being the most efficient.

\section*{Acknowledgements}
This work was partly funded by the European Union's Horizon 2020 research and innovation program under grant agreement No 871449-OpenDR and a Google Cloud research grant.

\bibliographystyle{spbasic}      

\begin{small}
\bibliography{sections/references}
\end{small}

\end{document}